\documentclass[a4paper]{article} 

\usepackage[utf8]{inputenc}
\usepackage[english]{babel}
\usepackage{csquotes}
\usepackage{graphicx,float}
\usepackage{placeins}
\usepackage{natbib}
\usepackage{subcaption}
\usepackage{amsmath,amsbsy,amssymb,dsfont}
\usepackage{algorithm,algpseudocode}
\usepackage{amsthm}
\usepackage{etoolbox}
\usepackage{pgfplots}
\usepackage{diagbox}
\usepackage{enumitem}
\usepackage{tabu,makecell,multirow,booktabs}
\usepackage{xspace}
\usepackage{siunitx}
\usepackage{sectsty}
\usepackage[font=small]{caption}
\usepackage{hyperref}
\usepackage{cleveref}

\newcommand{\abstracttext}{We propose a novel methodology for general multi-class classification in arbitrary feature spaces, which results in a potentially well-calibrated classifier. Calibrated classifiers are important in many applications because, in addition to the prediction of mere class labels, they also yield a confidence level for each of their predictions. In essence, the training of our classifier proceeds in two steps. In a first step, the training data is represented in a latent space whose geometry is induced by a regular $(n-1)$-dimensional simplex, $n$ being the number of classes. We design this representation in such a way that it well reflects the feature space distances of the datapoints to their own- and foreign-class neighbors. In a second step, the latent space representation of the training data is extended to the whole feature space by fitting a regression model to the transformed data. With this latent-space representation, our calibrated classifier is readily defined. We rigorously establish its core theoretical properties and benchmark its prediction and calibration properties by means of various synthetic and real-world data sets from different application domains.}

\usepackage[font=small,labelfont=bf]{caption}
\usepackage{orcidlink}
\usepackage{hyperref}
\usepackage{cleveref}

\usepackage[top=25mm, bottom=25mm, left=25mm, right=25mm]{geometry}

\crefname{section}{Section}{Sections}
\Crefname{section}{Section}{Sections}
\crefname{subsection}{Section}{Sections}
\Crefname{subsection}{Section}{Sections}
\crefname{appendix}{Appendix}{Appendices}
\Crefname{appendix}{Appendix}{Appendices}
\crefname{figure}{Fig.}{Figs.}
\Crefname{figure}{Figure}{Figures}
\crefname{table}{Table}{Tables}
\Crefname{table}{Table}{Tables}
\crefname{equation}{}{}
\Crefname{equation}{}{}
\crefname{lem}{Lemma}{Lemmas}
\Crefname{lem}{Lemma}{Lemmas}
\crefname{algorithm}{Algorithm}{Algorithms}
\Crefname{algorithm}{Algorithm}{Algorithms}
\crefname{line}{line}{lines}
\Crefname{line}{line}{lines}

\hypersetup{colorlinks=true,allcolors=black}

\setcitestyle{square,numbers,comma}

\title{Calibrated simplex-mapping classification}
\author{Raoul~Heese\textsuperscript{1,2,*}\orcidlink{0000-0001-7479-3339}, Jochen~Schmid\textsuperscript{2}\orcidlink{0000-0001-6527-537X}, Micha{\l}~Walczak\textsuperscript{1,2}\orcidlink{0000-0001-5151-7462}, Michael~Bortz\textsuperscript{1,2}\orcidlink{0000-0001-8169-2907}}
\date{%
\small%
\textsuperscript{1}Fraunhofer Center for Machine Learning\\
\textsuperscript{2}Fraunhofer Institute for Industrial Mathematics ITWM, Fraunhofer-Platz 1, 67663 Kaiserslautern, Germany\\
\textsuperscript{*}{raoul.heese@itwm.fraunhofer.de}
}

\newcommand{\captiontitle}[2]{\caption{#1}}

\newcommand{\startdocument}{\relax}
\newcommand{\titlehere}{\maketitle}
\newcommand{\abstracthere}{\begin{abstract}
\abstracttext
\end{abstract}
}
\newcommand{\finalparthere}{%
\section*{Acknowledgments}
We would like to thank Janis Keuper and J\"urgen Franke for their helpful and constructive comments.
This work was developed in the Fraunhofer Cluster of Excellence \enquote{Cognitive Internet Technologies}. We also gratefully acknowledge additional funding from the Deutsche Forschungsgemeinschaft (DFG, German Research Foundation) within the Priority Programme \enquote{SPP 2331: Machine Learning in Chemical Engineering}.
\FloatBarrier
\begin{appendix}
\section{Segmentation of latent space and core properties of calibrated simplex-mapping classifiers} 
\label{sect:math-background}

In this appendix, we collect some supplementary results needed for a detailed and mathematically sound understanding of our CASIMAC methodology. In particular, we prove the core properties of our segmentation of the latent space and of our CASIMAC. As in the main text, $n$ always stands for an integer larger than or equal to $2$ and
\begin{align}
	\mathcal{Y} := \{1,\dots,n\} 
	\qquad \text{and} \qquad
	\mathcal{Z} := \R^{n-1}
\end{align}
denote the corresponding set of class labels and, respectively, the corresponding latent space. Also, $\norm{z}_2 := (z\cdot z)^{1/2}$ always stands for the standard $\ell^2$-norm of $z \in \mathcal{Z}$, which is induced by the standard scalar product $(z,w) \mapsto z\cdot w = \sum_{i=1}^{n-1} z_i w_i$ on $\mathcal{Z}$.

\subsection{Simplices} 

We begin by briefly discussing $(n-1)$-simplices in $\mathcal{Z}$~\citep{rockafellar1970} and, especially, regular $(n-1)$-simplices in $\mathcal{Z}$~\citep{coxeter1973}. Simplices are the generalization of triangles to arbitrary dimensions: for example, an $(n-1)$-simplex corresponds to a line segment for $n=2$, a triangle for $n=3$, and a tetrahedron for $n=3$. Specifically, a subset $\mathcal{S}$ of $\mathcal{Z}$ is called an \emph{$(n-1)$-dimensional simplex} or, for short, an \emph{$(n-1)$-simplex} in $\mathcal{Z}$ iff it is the convex hull of $n$ affinely independent points $p_1, \dots, p_n$ in $\mathcal{Z}$, that is, iff
\begin{align*}
	\mathcal{S} = \cnv \{p_1,\dots,p_n\} := \bigg\{ z \in \mathcal{Z}: z = \sum_{i \in \mathcal{Y}} \lambda_i p_i \text{ for some } \lambda_i \in [0,\infty) \text{ with } \sum_{i \in \mathcal{Y}} \lambda_i = 1 \bigg\} 
\end{align*} 
for some affinely independent points $p_1, \dots, p_n \in \mathcal{Z}$. As usual, affine independence of the points $p_1, \dots, p_n$ simply means that, for some (and hence every) $k \in \mathcal{Y}$, the set $\{p_i-p_k: i \in \mathcal{Y}\setminus\{k\}\}$ of all vectors connecting $p_k$ with the points $p_i$ is linearly independent. It is straightforward to verify that the affinely independent points $p_1, \dots, p_n$ with $\mathcal{S} = \cnv \{p_1,\dots,p_n\}$ are uniquely determined by the set $\mathcal{S}$, namely as the extreme points of $\mathcal{S}$. (See~\citep{tuy2016} (Proposition~1.17), for instance.) These uniquely determined extreme points $p_1, \dots, p_n$ of $\mathcal{S}$ are also called the \emph{vertices of $\mathcal{S}$}. If $\mathcal{S}$ is an $(n-1)$-simplex with vertices $p_1, \dots, p_n$, then the point
\begin{align}
	c := \frac{1}{n} \sum_{i\in\mathcal{Y}} p_i \in \cnv\{p_1,\dots,p_n\} = \mathcal{S}
\end{align}
is called the \emph{barycenter of $p_1, \dots, p_n$} and, by extension, the \emph{barycenter of $\mathcal{S}$}. Also, the simplex $\mathcal{S}$ is called \emph{regular} iff all of its $n(n-1)/2$ edges have the same (positive) length, that is, iff 
\begin{align}
	\norm{p_i-p_j}_2 = l \qquad (i,j \in \mathcal{Y} \text{ with } i \ne j)
\end{align}
for some $l \in (0,\infty)$, which is also called the edge length of $\mathcal{S}$. See page~121 of~\citep{coxeter1973}, for instance. In our proofs, we repeatedly use the following straightforward fact. 

\begin{lm} \label{lm:vertices-lin-independent}
	Suppose $p_1, \dots, p_n$ are the vertices of an $(n-1)$-simplex in $\mathcal{Z}$ with barycenter $0$. Then 
	\begin{align} \label{eq:vertices-lin-independent}
		\spn\{p_i: i \in \mathcal{Y}\setminus\{k\}\} = \mathcal{Z}
	\end{align}
	for every $k \in \mathcal{Y}$. In particular, for every $k \in \mathcal{Y}$, the $(n-1)$-element subset $\{p_i: i \in \mathcal{Y}\setminus\{k\}\}$ of the vertex set is linearly independent. 
\end{lm}

\begin{proof}
	We give the straightforward proof for the sake of completeness. Since, by the definition of simplex vertices, the set $\{p_i-p_k: i \in \mathcal{Y}\setminus\{k\}\}$ is a linearly independent set of $n-1$ vectors in the $(n-1)$-dimensional vector space $\mathcal{Z}$, we see that
	\begin{align} \label{eq:vertices-lin-independent-1}
		\mathcal{Z} = \spn\{p_i-p_k: i \in \mathcal{Y}\setminus\{k\}\}
	\end{align}
	for every $k \in \mathcal{Y}$. Since, moreover, the barycenter of the vertices $p_1,\dots, p_n$ is $0$, we further see that
	\begin{align} \label{eq:p_k-in-terms-of-other-p_i}
		p_k = - \sum_{i\in\mathcal{Y}\setminus\{k\}} p_i \in \spn\{p_i: i \in \mathcal{Y}\setminus\{k\}\}
	\end{align}
	and therefore
	\begin{align} \label{eq:vertices-lin-independent-2}
		\spn\{p_i-p_k: i \in \mathcal{Y}\setminus\{k\}\} \subset \spn\{p_i: i \in \mathcal{Y}\setminus\{k\}\} \subset \mathcal{Z}
	\end{align}
	for every $k \in \mathcal{Y}$. Combining~\eqref{eq:vertices-lin-independent-1} and~\eqref{eq:vertices-lin-independent-2}, we obtain the asserted equality~\eqref{eq:vertices-lin-independent}. In particular, this equality implies the asserted linear independence by a trivial dimensionality argument.
\end{proof}

We now construct an explicit example of the kind of simplices that underlie our segmentation of $\mathcal{Z}$ and our CASIMAC. In other words, we construct a regular $(n-1)$-simplex in $\mathcal{Z}$ with barycenter $0$ and with vertices $p_1, \dots, p_n$ at unit distance from $0$, that is, 
\begin{align} \label{eq:barycenter-0-and-unit-distance-from-0}
	\sum_{i\in\mathcal{Y}} p_i = 0 
	\qquad \text{and} \qquad
	\norm{p_i}_2 = 1 \qquad (i \in \mathcal{Y}). 
\end{align}
In the extreme special case $n=2$, it is clear that there is only one such simplex, namely the line segment between the vertices $-1$ and $1$ in $\mathcal{Z} = \R$, and we usually index these two  vertices increasingly (instead of decreasingly) then, that is, 
\begin{align} \label{eq:vertex-indexing-in-binary-case}
	p_1 = -1 \qquad \text{and} \qquad p_2 = 1.
\end{align}
In the following, we show that in higher dimensions there still is only one such simplex -- provided that we identify mutually congruent simplices. 

\begin{prop} \label{prop:sample-reg-simplex-in-Z}
	A regular $(n-1)$-simplex with vertices $p_1, \dots, p_n$ satisfying~\eqref{eq:barycenter-0-and-unit-distance-from-0} exists in $\mathcal{Z}$. 
	And, moreover, all other regular $(n-1)$-simplices in $\mathcal{Z}$ with~\eqref{eq:barycenter-0-and-unit-distance-from-0} are congruent to it. 
\end{prop}

\begin{proof}
	We begin by constructing just any regular $(n-1)$-simplex $\cnv\{q_1,\dots,q_n\}$ in $\mathcal{Z}$. In order to do so, we make the ansatz
	\begin{align} \label{eq:sample-reg-simplex-in-Z-1}
		q_i = e_i \qquad (i\in\{1,\dots,n-1\})
		\qquad \text{and} \qquad
		q_n = (\alpha_1,\dots,\alpha_{n-1}),
	\end{align}
	where $e_1,\dots,e_{n-1}$ denote the canonical unit vectors in $\R^{n-1}$ and the coordinates $\alpha_1,\allowbreak \dots,\allowbreak \alpha_{n-1} \in \R$ are to be determined such that the vertices $q_1,\dots,q_n$ all have the same distance from each other. Since, for $i, j \in \{1,\dots,n-1\}$, 
	\begin{gather} 
		\norm{q_i-q_j}_2^2 = \norm{e_i-e_j}_2^2 = 2 \qquad (i\ne j)
		\label{eq:sample-reg-simplex-in-Z-2}\\
		\norm{q_i-q_n}_2^2 = \norm{e_i-q_n}_2^2 = 1 - 2 e_i \cdot q_n + \norm{q_n}_2^2 = 1 - 2\alpha_i + \sum_{l=1}^{n-1} \alpha_l^2
		\label{eq:sample-reg-simplex-in-Z-3}
	\end{gather}
	by virtue of~\eqref{eq:sample-reg-simplex-in-Z-1}, we have to choose the coordinates
	\begin{align} \label{eq:sample-reg-simplex-in-Z-4}
		\alpha_i = \alpha := (\sum_{l=1}^{n-1} \alpha_l^2 - 1)/2
		\qquad (i\in\{1,\dots,n-1\})
	\end{align} 
	to be independent of $i$. Inserting~\eqref{eq:sample-reg-simplex-in-Z-4} into~\eqref{eq:sample-reg-simplex-in-Z-3}, we get a quadratic equation in $\alpha$ which has the solutions 
	\begin{align} \label{eq:sample-reg-simplex-in-Z-5}
		\alpha = \alpha_{\pm} := (1\pm \sqrt{n})/(n-1).
	\end{align}
	In view of these preliminary considerations, we define the points $q_1,\dots,q_n \in \mathcal{Z}$ by
	\begin{align} \label{eq:sample-reg-simplex-in-Z-6}
		q_i := e_i \qquad (i\in\{1,\dots,n-1\})
		\qquad \text{and} \qquad 
		q_n :=  \frac{1 + \sqrt{n}}{n-1} (1,\dots,1). 
	\end{align}
	It is then clear that $\spn\{q_i-q_n: i \in \{1,\dots,n-1\}\} = \mathcal{Z}$ and thus $q_1, \dots, q_n$ are the vertices of an $(n-1)$-simplex in $\mathcal{Z}$. It is also clear by~\eqref{eq:sample-reg-simplex-in-Z-2} and~\eqref{eq:sample-reg-simplex-in-Z-3} that these vertices all have the same distance from each other, namely $\sqrt{2}$. Consequently, $q_1,\dots,q_n$ are the vertices of a regular $(n-1)$-simplex in $\mathcal{Z}$. 
	In particular, by shifting them by their barycenter $c$ and by then  normalizing, we obtain the vertices
	\begin{align} \label{eq:sample-reg-simplex-in-Z-7}
		p_i := (q_i-c)/\nu \qquad (i \in \mathcal{Y})
	\end{align}
	of a regular $(n-1)$-simplex satisfying~\eqref{eq:barycenter-0-and-unit-distance-from-0}, as desired. Specifically,
	\begin{align} \label{eq:sample-reg-simplex-in-Z-8}
		c := \frac{1}{n} \sum_{i\in\mathcal{Y}} q_i = \frac{1+1/\sqrt{n}}{n-1} \, (1,\dots,1)
		\qquad \text{and} \qquad
		\nu := \norm{q_i-c}_2 = \sqrt{1-1/n}. 
	\end{align} 
	It remains to prove that any other regular $(n-1)$-simplex satisfying~\eqref{eq:barycenter-0-and-unit-distance-from-0} is congruent to the simplex constructed above. In order to do so, we have only to notice that any such other simplex has the same edge length as the simplex constructed above and that all regular simplices with equal edge length are congruent to each other.  
\end{proof}

\begin{lm} \label{lm:central-angle-of-reg-simplex}
	Suppose $p_1,\dots,p_n$ are the vertices of a regular $(n-1)$-simplex in $\mathcal{Z}$ with~\eqref{eq:barycenter-0-and-unit-distance-from-0}. Then 
	\begin{align} \label{eq:simplex-central-angle-of-reg-simplex}
		p_i \cdot p_j = 1 \qquad (i=j) 
		\qquad \text{and} \qquad
		p_i \cdot p_j = -1/(n-1) \qquad (i\ne j). 
	\end{align}
\end{lm}

\begin{proof}
	An immediate consequence of~\citep{parks2002} (proof of Theorem~1). 
\end{proof}

\subsection{Simplex-induced segmentation of the latent space}

In this section, we introduce the segmentation of the latent space $\mathcal{Z}$ underlying our classification method and establish the core properties of this segmentation. We choose a segmentation into $n$ convex cone segments $\mathcal{C}_1, \dots, \mathcal{C}_n$ and we define these cone segments in terms of the vertices of an $(n-1)$-simlex with barycenter $0$. As usual, a convex cone in $\mathcal{Z}$ is a convex subset of $\mathcal{Z}$ that is closed under addition and under scalar multiplication with positive scalars~\citep{tuy2016} (Proposition~1.7). 

\begin{lm} \label{lm:barycentric-char-of-C_k}
	Suppose $p_1,\dots,p_n$ are the vertices of an $(n-1)$-simplex in $\mathcal{Z}$ with~(\ref{eq:barycenter-0-and-unit-distance-from-0}.a) and, for every $k \in \mathcal{Y}$, let 
	\begin{align} \label{eq:def-of-C_k}
		\mathcal{C}_k := \bigg\{z \in \mathcal{Z}: z = \sum_{i\in\mathcal{Y}\setminus\{k\}} c_i \cdot (-p_i) \text{ for some } c_i \in [0,\infty) \bigg\}
	\end{align}
	be the convex cone generated by the mirrored vertices $-p_i$ for $i \in \mathcal{Y}\setminus\{k\}$. Suppose further that $z$ is a point in $\mathcal{Z}$ having the representation
	\begin{align} \label{eq:barycentric-representation-of-z}
		z = \sum_{i\in\mathcal{Y}} r_i \cdot (-p_i)
	\end{align}
	for some $r_1, \dots, r_n \in \R$. Then, for every $k \in \mathcal{Y}$, 
	one has the following equivalence: $z \in \mathcal{C}_k$ if and only if $r_k \le r_i$ for all $i \in \mathcal{Y}$.
\end{lm}

\begin{proof}
	Choose and fix an arbitrary $k \in \mathcal{Y}$. It then follows from~\eqref{eq:barycentric-representation-of-z} with the help of~\eqref{eq:p_k-in-terms-of-other-p_i} that
	\begin{align} \label{eq:barycentric-char-of-C_k-1}
		z = r_k \sum_{i\in\mathcal{Y}\setminus\{k\}} p_i + \sum_{i\in\mathcal{Y}\setminus\{k\}} r_i \cdot (-p_i) = \sum_{i\in\mathcal{Y}\setminus\{k\}} (r_i - r_k) \cdot (-p_i).
	\end{align}
	If $r_k \le r_i$ for all $i \in \mathcal{Y}$, then it immediately follows from~\eqref{eq:barycentric-char-of-C_k-1} and the definition~\eqref{eq:def-of-C_k} of $\mathcal{C}_k$ that $z \in \mathcal{C}_k$, as desired.
	If, conversely, $z \in \mathcal{C}_k$, then 
	\begin{align} \label{eq:barycentric-char-of-C_k-2}
		z = \sum_{i\in\mathcal{Y}\setminus\{k\}} c_i \cdot (-p_i)
	\end{align}
	for some $c_i \in [0,\infty)$ by the definition~\eqref{eq:def-of-C_k} of $\mathcal{C}_k$ and, therefore, it follows by virtue of~\eqref{eq:barycentric-char-of-C_k-1} and~\eqref{eq:barycentric-char-of-C_k-2} that
	\begin{align} \label{eq:barycentric-char-of-C_k-3}
		\sum_{i\in\mathcal{Y}\setminus\{k\}} (r_i - r_k) \cdot p_i = -z = \sum_{i\in\mathcal{Y}\setminus\{k\}} c_i \cdot p_i.
	\end{align}
	Since $\{p_i: i \in \mathcal{Y}\setminus\{k\}\}$ is linearly independent by Lemma~\ref{lm:vertices-lin-independent}, we conclude from~\eqref{eq:barycentric-char-of-C_k-3} that $r_i-r_k = c_i \ge 0$ for all $i \in \mathcal{Y}\setminus\{k\}$ and therefore $r_k \le r_i$ for all $i \in \mathcal{Y}$, as desired.
\end{proof}

\begin{lm} \label{lm:closure-and-interior-of-C_k}
	Suppose $p_1,\dots,p_n$ are the vertices of an $(n-1)$-simplex in $\mathcal{Z}$ with~(\ref{eq:barycenter-0-and-unit-distance-from-0}.a) and let $\mathcal{C}_k$ be the cone defined in~\eqref{eq:def-of-C_k}. Then the closure $\ol{\mathcal{C}}_k$ and the interior $\mathcal{C}_k^{\circ}$ of $\mathcal{C}_k$ are given, respectively, by
	\begin{align} \label{eq:closure-and-interior-of-C_k}
		\ol{\mathcal{C}}_k = \mathcal{C}_k 
		\quad \text{and} \quad 
		\mathcal{C}_k^{\circ} = \bigg\{z \in \mathcal{Z}: z = \sum_{i\in\mathcal{Y}\setminus\{k\}} c_i \cdot (-p_i) \text{ for some } c_i \in (0,\infty) \bigg\}.
	\end{align}
	In particular, $\mathcal{C}_k$ is a closed set in $\mathcal{Z}$. 
\end{lm}

\begin{proof}
	Choose and fix an arbitrary $k \in \mathcal{Y}$ and let $\phi_k: \R^{n-1} \to \mathcal{Z}$ be the linear map defined by $\phi_k(e_i) = -p_i$ for $i \in \{1, \dots, k-1\}$ and $\phi_k(e_i) = -p_{i+1}$ for $i \in \{k,\dots,n-1\}$, where $e_1, \dots, e_{n-1}$ denote the canonical unit vectors in $\R^{n-1}$. It is then obvious that the cone $\mathcal{C}_k$ is the image under $\phi_k$ of the non-negative orthant $P := [0,\infty)^{n-1}$ in $\R^{n-1}$. In short,
	\begin{align} \label{eq:closure-and-interior-of-C_k-1}
		\mathcal{C}_k = \phi_k(P).
	\end{align}
	It is also clear, by the linear independence of $\{\phi_k(e_i): i \in \{1,\dots,n-1\}\}$ (Lemma~\ref{lm:vertices-lin-independent}), that $\phi_k$ is a linear isomorphism of $\mathcal{Z} = \R^{n-1}$ and thus also a homeomorphism of $\mathcal{Z}$ (Theorem~VII.1.6 in of~\citep{amann2008}). Since obviously $\ol{P} = P$ and $P^{\circ} = (0,\infty)^{n-1}$, we see by~\eqref{eq:closure-and-interior-of-C_k-1} and~\citep{dugundji1966} (Theorem~III.11.3 and Theorem~III.11.4) that 
	\begin{align*}
		\ol{\mathcal{C}}_k = \ol{\phi_k(P)} = \phi_k(\ol{P}) = \phi_k(P) = \mathcal{C}_k,
		\qquad
		\mathcal{C}_k^{\circ} = (\phi_k(P))^{\circ} = \phi_k(P^{\circ}) = \phi_k((0,\infty)^{n-1}).
	\end{align*}
	Clearly, $\phi_k((0,\infty)^{n-1})$ is equal to the set on the right-hand side of~(\ref{eq:closure-and-interior-of-C_k}.b) and thus the proof is finished. 
\end{proof}

With the help of the preceding lemmas, we can prove that the cone segments cover the whole latent space and that they are essentially non-overlapping (up to boundary points). It should be noticed that for this result, we do not have to require the underlying $(n-1)$-simplex to be regular.

\begin{prop} \label{prop:segmentation-of-Z}
	Suppose $p_1,\dots,p_n$ are the vertices of an $(n-1)$-simplex in $\mathcal{Z}$ with~(\ref{eq:barycenter-0-and-unit-distance-from-0}.a) and let $\mathcal{C}_1, \dots, \mathcal{C}_n$ be the cones defined in~\eqref{eq:def-of-C_k}. Then, for every $k \in \mathcal{Y}$, the vertex $p_k$ lies on the central ray
	\begin{align} \label{eq:central-ray-of-C_k}
		\bigg\{z \in \mathcal{Z}: z = \sum_{i\in\mathcal{Y}\setminus\{k\}} c \cdot (-p_i) \text{ for some } c \in [0,\infty)\bigg\}
	\end{align}
	of the cone $\mathcal{C}_k$. Also, the cones $\mathcal{C}_1, \dots, \mathcal{C}_n$ cover the whole of $\mathcal{Z}$, and any two cones $\mathcal{C}_k$ and $\mathcal{C}_l$ for $k\ne l$ overlap only at their boundaries. In short,
	\begin{align} \label{eq:segmentation-of-Z-appendix}
		\mathcal{Z} = \bigcup_{k \in \mathcal{Y}} \mathcal{C}_k
		\qquad \text{and} \qquad
		\mathcal{C}_k^{\circ} \cap \mathcal{C}_l = \emptyset \qquad (k\ne l). 
	\end{align}
\end{prop}

\begin{proof}
	In view of~\eqref{eq:p_k-in-terms-of-other-p_i}, it is obvious that the vertex $p_k$ lies on the central ray~\eqref{eq:central-ray-of-C_k} of $\mathcal{C}_k$ for every $k \in \mathcal{Y}$.
	\par
	
	In order to prove~(\ref{eq:segmentation-of-Z-appendix}.a), we have only to show that every $z \in \mathcal{Z}$ is contained in some $\mathcal{C}_k$. Let $z \in \mathcal{Z}$ be an arbitrary point in $\mathcal{Z}$. It then follows by~\eqref{eq:vertices-lin-independent} that $z$ has a representation of the form~\eqref{eq:barycentric-representation-of-z} for some $r_1, \dots, r_n \in \R$. So, by Lemma~\ref{lm:barycentric-char-of-C_k}, we see that $z \in \mathcal{C}_k$ for every $k$ with $r_k = \min_{i\in\mathcal{Y}} r_i$.
	\par
	
	In order to prove~(\ref{eq:segmentation-of-Z-appendix}.b), we argue by contradiction. Let $k,l \in \mathcal{Y}$ be fixed with $k\ne l$ and assume, for the sake of argument, that there exists a point $z \in \mathcal{Z}$ with
	\begin{align} \label{eq:segmentation-of-Z-1}
		z \in \mathcal{C}_k^{\circ} \cap \mathcal{C}_l. 
	\end{align}
	In view of~\eqref{eq:vertices-lin-independent}, $z$ has a representation of the form~\eqref{eq:barycentric-representation-of-z} for some $r_1, \dots, r_n \in \R$. Since $z \in \mathcal{C}_k \cap \mathcal{C}_l$ by assumption~\eqref{eq:segmentation-of-Z-1}, we therefore see by Lemma~\ref{lm:barycentric-char-of-C_k} that
	\begin{align} \label{eq:segmentation-of-Z-2}
		r_k = \min_{i\in\mathcal{Y}} r_i = r_l.
	\end{align}
	In order to arrive at a contradiction, we consider suitable perturbations $z^{\eps}$ of $z$, namely
	\begin{align} \label{eq:segmentation-of-Z-3}
		z^{\eps} := \sum_{i\in\mathcal{Y}} r_i^{\eps} \cdot (-p_i)
		\qquad (\eps \in (0,\infty)),
	\end{align}
	where $r_i^{\eps} := r_i + \eps$ for $i \in \mathcal{Y}\setminus\{l\}$ and $r_l^{\eps} := r_l$. It is then clear by~\eqref{eq:segmentation-of-Z-2} that $r_l^{\eps} = r_l = r_k < r_k + \eps = r_k^{\eps}$ for every $\eps \in (0,\infty)$ 
	and therefore we see by Lemma~\ref{lm:barycentric-char-of-C_k} that
	\begin{align} \label{eq:segmentation-of-Z-5}
		z^{\eps} \notin \mathcal{C}_k 
		\qquad (\eps \in (0,\infty)).
	\end{align}
	Since, on the other hand, $z \in \mathcal{C}_k^{\circ}$ by assumption~\eqref{eq:segmentation-of-Z-1} and the perturbations $z^{\eps}$ obviously converge to $z$, we also see  that
	\begin{align} \label{eq:segmentation-of-Z-6}
		z^{\eps} \in \mathcal{C}_k 
		\qquad (\eps \in (0,\eps_0])
	\end{align} 
	for some sufficiently small $\eps_0 > 0$. Contradiction between~\eqref{eq:segmentation-of-Z-5} and~\eqref{eq:segmentation-of-Z-6}! So, our  assumption~\eqref{eq:segmentation-of-Z-1} cannot be true, as desired. 
\end{proof}

If we additionally require the $(n-1)$-simplex from the previous proposition to be regular, we can further prove that the corresponding cone segments are mutually congruent. 

\begin{prop} \label{prop:C_k-and-C_l-congruent}
	Suppose $p_1,\dots,p_n$ are the vertices of a regular $(n-1)$-simplex in $\mathcal{Z}$ with~\eqref{eq:barycenter-0-and-unit-distance-from-0} and let $\mathcal{C}_1, \dots, \mathcal{C}_n$ be the cone segments defined in~\eqref{eq:def-of-C_k}. Then any two of these cone segments are congruent to each other. 
\end{prop}

\begin{proof}
	We proceed in three steps and, for the entire proof, we fix $k,l \in \mathcal{Y}$ with $k \ne l$. 
	As a first step, we show that the mid-perpendicular hyperplane 
	\begin{align} \label{eq:def-M_kl}
		\mathcal{M}_{kl} := \{z \in \mathcal{Z}: z \cdot (p_k-p_l) = 0 \}
	\end{align}
	between the vertices $p_k$ and $p_l$ is equal to the subspace of $\mathcal{Z}$ spanned by all other vertices, that is,
	\begin{align} \label{eq:C_k-and-C_l-congruent-1}
		\mathcal{M}_{kl} = \spn\{p_i: i \in \mathcal{Y}\setminus\{k,l\}\}. 
	\end{align}
	Indeed, in view of Lemma~\ref{lm:central-angle-of-reg-simplex}, we have $p_i \cdot (p_k-p_l) = 0$ for all $i \in \mathcal{Y}\setminus\{k,l\}$ and therefore
	\begin{align} \label{eq:C_k-and-C_l-congruent-2}
		\{p_i: i \in \mathcal{Y}\setminus\{k,l\}\} \subset \mathcal{M}_{kl}.
	\end{align}
	Since $\{p_i: i \in \mathcal{Y}\setminus\{k,l\}\}$ is a linearly independent set of $n-2$ vectors by Lemma~\ref{lm:vertices-lin-independent} and $\mathcal{M}_{kl}$ is an $(n-2)$-dimensional subspace of $\mathcal{Z}$, the asserted equality~\eqref{eq:C_k-and-C_l-congruent-1} follows from~\eqref{eq:C_k-and-C_l-congruent-2} and a trivial dimensionality argument.
	\par
	
	As a second step, we show that the segment $\mathcal{C}_l$ is the mirror image of the segment $\mathcal{C}_k$ under the reflection $\rho_{kl}$ at the mid-perpendicular hyperplane $\mathcal{M}_{kl}$, 
	that is,
	\begin{align} \label{eq:C_k-and-C_l-congruent-3}
		\rho_{kl}(\mathcal{C}_k) = \mathcal{C}_l,
	\end{align}
	Indeed, the reflection at $\mathcal{M}_{kl}$ is the linear map $\rho_{kl}$ given by
	\begin{align} \label{eq:C_k-and-C_l-congruent-4}
		\rho_{kl}(z) = \pi_{kl}(z) - (z-\pi_{kl}(z)) = 2\pi_{kl}(z) - z 
		\qquad (z\in \mathcal{Z}),
	\end{align}
	where $\pi_{kl}$ is the orthogonal projection onto $\mathcal{M}_{kl}$. 
	It then follows 
	by the first step that $\pi_{kl}(p_i) = p_i$ for all $i \in \mathcal{Y}\setminus\{k,l\}$ and that $\pi_{kl}(p_k) = (p_k+p_l)/2 = \pi_{kl}(p_l)$ and therefore
	\begin{align} \label{eq:C_k-and-C_l-congruent-5}
		\rho_{kl}(p_i) = p_i \quad (i \in \mathcal{Y}\setminus\{k,l\})
		\quad \text{and} \quad
		\rho_{kl}(p_k) = p_l 
		\quad \text{and} \quad
		\rho_{kl}(p_l) = p_k. 
	\end{align}
	And from this, in turn, the asserted equality~\eqref{eq:C_k-and-C_l-congruent-3} immediately follows by the definition~\eqref{eq:def-of-C_k} of the cone segments.  
	\par
	
	As a third step, we finally establish the asserted congruence of the segments $\mathcal{C}_k$ and $\mathcal{C}_l$. 
	Indeed, this immediately follows by the second step, because reflections obviously are congruence transformations (isometries) of $\mathcal{Z}$. 
\end{proof}

With the preceding results at hand, we can establish a distance-based characterization of our cone segments which is central for an economical computation of our classifier's class label and class label probability predictions $\widehat{y}(x)$ and  $\widehat{p}(\,\cdot\,|x)$, respectively. See~\eqref{eq:casimac-computation-appendix} and~\eqref{eq:class-probability-approximant} below.

\begin{thm} \label{thm:norm-char-of-C_k}
	Suppose $p_1,\dots,p_n$ are the vertices of a regular $(n-1)$-simplex in $\mathcal{Z}$ with~\eqref{eq:barycenter-0-and-unit-distance-from-0} and let $\mathcal{C}_1, \dots, \mathcal{C}_n$ be the cone segments defined in~\eqref{eq:def-of-C_k}. Then these cone segments can be characterized in terms of the distances from the central vectors $p_1, \dots, p_n$ of the cones $\mathcal{C}_1, \dots, \mathcal{C}_n$. Specifically,
	\begin{align} \label{eq:norm-char-of-C_k}
		\mathcal{C}_k = \big\{ z \in \mathcal{Z}: \norm{z-p_k}_2 \le \norm{z-p_l}_2 \text{ for all } l \in \mathcal{Y} \big\}.
	\end{align}
	And analogously, 
	for the interiors of the cone segments one has the following characterization:
	\begin{align} \label{eq:norm-char-of-C_k^circ}
		\mathcal{C}_k^{\circ} = \big\{ z \in \mathcal{Z}: \norm{z-p_k}_2 < \norm{z-p_l}_2 \text{ for all } l \in \mathcal{Y}\setminus\{k\} \big\}.
	\end{align}
\end{thm}

\begin{proof}
	We proceed in four steps and, for the entire proof, we fix $k \in \mathcal{Y}$.
	As a first step, we observe that the set on the right-hand side of~\eqref{eq:norm-char-of-C_k} is nothing but the intersection of the  half-spaces
	\begin{align} \label{eq:def-H_kl}
		\mathcal{H}_{kl}(p_k) := \{z \in \mathcal{Z}: z\cdot (p_k-p_l) \ge 0\}
	\end{align}
	for all $l \in \mathcal{Y} \setminus \{k\}$, that is, the half-spaces confined by the mid-perpendicular hyperplanes $\mathcal{M}_{kl}$ from~\eqref{eq:def-M_kl} and stretching to the side of $p_k$ (instead of its mirror image $p_l$). In other words, as a first step we observe that 
	\begin{align} \label{eq:norm-char-of-C_k-1}
		\big\{ z \in \mathcal{Z}: \norm{z-p_k}_2 \le \norm{z-p_l}_2 \text{ for all } l \in \mathcal{Y} \big\} = \bigcap_{l \in \mathcal{Y} \setminus \{k\}} \mathcal{H}_{kl}(p_k).
	\end{align}
	Indeed, by~(\ref{eq:barycenter-0-and-unit-distance-from-0}.b) we immediately obtain
	\begin{align} \label{eq:norm-char-of-C_k-1.1}
		\norm{z-p_k}^2_2 = \norm{z}_2^2 - 2 z\cdot p_k + 1
		\qquad \text{and} \qquad
		\norm{z-p_l}^2_2 = \norm{z}_2^2 - 2 z\cdot p_l + 1
	\end{align}
	for every $l \in \mathcal{Y}$ and every $z \in \mathcal{Z}$. And from this, in turn, the asserted equality~\eqref{eq:norm-char-of-C_k-1} readily follows. 
	\par
	
	As a second step, we show that the segment $\mathcal{C}_k$ is contained in the right-hand side of~\eqref{eq:norm-char-of-C_k} or, equivalently (by the first step), that 
	\begin{align} \label{eq:norm-char-of-C_k-2}
		\mathcal{C}_k \subset \bigcap_{l \in \mathcal{Y} \setminus \{k\}} \mathcal{H}_{kl}(p_k). 
	\end{align}
	So, let $z \in \mathcal{C}_k$. We can then represent $z$ in the form~\eqref{eq:barycentric-char-of-C_k-2} with some $c_i \in [0,\infty)$ and we can thus conclude, using Lemma~\ref{lm:central-angle-of-reg-simplex}, that
	\begin{align*}
		z \cdot (p_k-p_l) 
		= c_l  (-p_l) \cdot (p_k-p_l) + \sum_{i \in \mathcal{Y}\setminus\{k,l\}} c_i  (-p_i) \cdot (p_k-p_l)  = c_l (1+1/(n-1)) \ge 0
	\end{align*}
	for every $l \in \mathcal{Y} \setminus \{k\}$. 
	And therefore, $z$ belongs to the set on the right-hand side of~\eqref{eq:norm-char-of-C_k-2}, as desired.
	\par
	
	As a third step, we show that conversely the right-hand side of~\eqref{eq:norm-char-of-C_k} is contained in the segment $\mathcal{C}_k$ or, equivalently (by the first step), that
	\begin{align} \label{eq:norm-char-of-C_k-3}
		\bigcap_{l \in \mathcal{Y} \setminus \{k\}} \mathcal{H}_{kl}(p_k)
		\subset \mathcal{C}_k.
	\end{align}
	So, let $z \in \bigcap_{l \in \mathcal{Y} \setminus \{k\}} \mathcal{H}_{kl}(p_k)$. We then have $z \cdot (p_k-p_l) \ge 0$ for every $l \in \mathcal{Y} \setminus \{k\}$ and therefore we see, using Lemma~\ref{lm:central-angle-of-reg-simplex}, that the perturbations
	\begin{align}
		z^{\eps} := z + \eps w := z + \eps \sum_{i \in \mathcal{Y}\setminus\{k\}} (p_k-p_i)
		\qquad (\eps \in (0,\infty))
	\end{align}
	of $z$ satisfy the strict inequality
	\begin{align} \label{eq:norm-char-of-C_k-3.1}
		z^{\eps} \cdot (p_k-p_l) = z \cdot (p_k-p_l) + \eps \norm{p_k-p_l}_2^2 + \eps (n-2)(1+1/(n-1)) > 0
	\end{align}
	for every $l \in \mathcal{Y} \setminus \{k\}$ and every $\eps \in (0,\infty)$. So, by~\eqref{eq:norm-char-of-C_k-1.1} with $z$ replaced by $z^{\eps}$, this strict inequality~\eqref{eq:norm-char-of-C_k-3.1} implies
	\begin{align} \label{eq:norm-char-of-C_k-3.2}
		\norm{z^{\eps}-p_k}_2^2 < \norm{z^{\eps}-p_l}_2^2
		\qquad (l \in \mathcal{Y} \setminus \{k\} \text{ and } \eps \in (0,\infty)).
	\end{align}
	And from~\eqref{eq:norm-char-of-C_k-3.2}, in turn, we can easily conclude that
	\begin{align} \label{eq:norm-char-of-C_k-3.3}
		z^{\eps} \in \mathcal{C}_k \qquad (\eps \in (0,\infty)). 
	\end{align}
	Indeed, by~(\ref{eq:segmentation-of-Z-appendix}.a), we see that for every $\eps \in (0,\infty)$ there exists an index $k' = k'_{\eps} \in \mathcal{Y}$ such that $z^{\eps} \in \mathcal{C}_{k'}$ and therefore, by the first and the second step with $k$ replaced by $k'$, 
	\begin{align}
		\norm{z^{\eps}-p_{k'}}_2 \le \norm{z^{\eps}-p_l}_2 
		\qquad (l \in \mathcal{Y} \setminus \{k'\}).
	\end{align}
	Clearly, this is compatible with~\eqref{eq:norm-char-of-C_k-3.2} only if $k' = k$. Consequently, we must have $k' = k$ so that $z^{\eps} \in \mathcal{C}_{k'} = \mathcal{C}_k$ and~\eqref{eq:norm-char-of-C_k-3.3} is established. Since the perturbations $z^{\eps}$ obviously converge to $z$, we see from~\eqref{eq:norm-char-of-C_k-3.3} and~(\ref{eq:closure-and-interior-of-C_k}.a) that $z \in \mathcal{C}_k$, as desired.
	\par
	
	As a fourth step, we finally establish the asserted equalities~\cref{eq:norm-char-of-C_k,eq:norm-char-of-C_k^circ}. 
	Indeed, \eqref{eq:norm-char-of-C_k} is an immediate consequence of~\cref{eq:norm-char-of-C_k-1,eq:norm-char-of-C_k-2,eq:norm-char-of-C_k-3}. And \eqref{eq:norm-char-of-C_k^circ}, in turn, is readily obtained by noting that the interior of the half-space $\mathcal{H}_{kl}(p_k)$ is given by
	\begin{align}
		(\mathcal{H}_{kl}(p_k))^{\circ} = \{z \in \mathcal{Z}: z\cdot (p_k-p_l) > 0\}
	\end{align}
	and by then taking the interior on both sides of the inclusions~\eqref{eq:norm-char-of-C_k-2} and~\eqref{eq:norm-char-of-C_k-3}. 
\end{proof}

In view of the normalization condition~(\ref{eq:barycenter-0-and-unit-distance-from-0}.b) on the vertices of our simplex, the norm characterization~\eqref{eq:norm-char-of-C_k} immediately translates to a scalar-product characterization of our cone segments. We will use it to prove the invariance of the cone segments under the compression and inflation maps from~\eqref{eq:def-of-compression-map} and~\eqref{eq:def-of-inflation-map} below. 

\begin{cor} \label{cor:scalar-product-char-of-C_k}
	Suppose $p_1,\dots,p_n$ are the vertices of a regular $(n-1)$-simplex in $\mathcal{Z}$ with~\eqref{eq:barycenter-0-and-unit-distance-from-0} and let $\mathcal{C}_1, \dots, \mathcal{C}_n$ be the cone segments defined in~\eqref{eq:def-of-C_k}. Then 
	\begin{align} \label{eq:scalar-product-char-of-C_k}
		\mathcal{C}_k = \big\{ z \in \mathcal{Z}: z \cdot p_k \ge z \cdot p_l  \text{ for all } l \in \mathcal{Y} \big\}.
	\end{align}
\end{cor}

\begin{proof}
	An immediate consequence of~\eqref{eq:norm-char-of-C_k} in conjunction with~\eqref{eq:norm-char-of-C_k-1.1}.
\end{proof}

\subsection{Core properties of calibrated simplex-mapping classifiers} 

In this section, we establish the core properties of our CASIMAC. We begin by proving that the underlying training data transformation $f$ maps every training datapoint $x_i$ to the interior $\mathcal{C}_{y(x_i)}^{\circ}$ of the cone segment corresponding to the true class label $y(x_i) = y_i$. 

\begin{prop} \label{prop:generic-properties-of-training-data-trf}
	Suppose $p_1,\dots,p_n$ are the vertices of a regular $(n-1)$-simplex in $\mathcal{Z}$ with~\eqref{eq:barycenter-0-and-unit-distance-from-0} and let $\mathcal{C}_1, \dots, \mathcal{C}_n$ be the cone segments defined in~\eqref{eq:def-of-C_k}. Suppose further that $\mathcal{D} = \{(x_i,y_i): i \in \{1,\dots,D\}\}$ is a finite subset of $\mathcal{X} \times \mathcal{Y}$ and that the map $f: \mathcal{D}_{\mathcal{X}} := \{x_1,\dots,x_D\} \to \mathcal{Z}$ is defined as in~\eqref{eq:training-data-trf-defining-formula}, \eqref{eq:attraction-and-repulsion-coefficients-definition}, and \eqref{eq:nearest-neighbors-definition} with a semimetric $d$ on $\mathcal{X}$ and hyperparameters $\alpha,\beta \in [0,\infty)$ and $k_{\alpha}, k_{\beta} \in \N$ satisfying~\eqref{eq:conditions-on-hyperparameters}. Then 
	\begin{align} \label{eq:generic-properties-of-training-data-trf}
		0 < A_{k_{\alpha},d}(x), R_{k_{\beta,d}}(x,y) < \infty 
	\end{align}
	for every $x \in \mathcal{D}_{\mathcal{X}}$ and $y \in \mathcal{Y} \setminus \{y(x)\}$. In particular, $f(x) \in \mathcal{C}_{y(x)}^{\circ}$ for every $x \in \mathcal{D}_{\mathcal{X}}$. 
\end{prop}

\begin{proof}
	Choose and fix $x \in \mathcal{D}_{\mathcal{X}}$. Also, for every $y \in \mathcal{Y}$, let $\mathcal{D}_{\mathcal{X},y} := \{x' \in \mathcal{D}_{\mathcal{X}}: y(x') = y\}$ be the set of training datapoints belonging to class $y$ and let $c := \min\{ |\mathcal{D}_{\mathcal{X},y}|: y \in \mathcal{Y}\}$ be the size of the smallest class in the training data. In order to prove~\eqref{eq:generic-properties-of-training-data-trf}, we also fix $y \in \mathcal{Y} \setminus \{y(x)\}$. It is then obvious that
	\begin{align}
		|\mathcal{D}_{\mathcal{X},y(x)} \setminus \{x\}| = |\mathcal{D}_{\mathcal{X},y(x)}| - 1 \ge c-1
		\qquad \text{and} \qquad
		|\mathcal{D}_{\mathcal{X},y} \setminus \{x\}| = |\mathcal{D}_{\mathcal{X},y}| \ge c.
	\end{align}
	Combining this with~(\ref{eq:conditions-on-hyperparameters}.b) and (\ref{eq:conditions-on-hyperparameters}.c), we further see that there exists a subset $X_{\alpha}' \subset \mathcal{D}_{\mathcal{X},y(x)} \setminus \{x\}$ with $|X_{\alpha}'| = k_{\alpha}$ and a subset $X_{\beta}' \subset \mathcal{D}_{\mathcal{X},y} \setminus \{x\}$ with $|X_{\beta}'| = k_{\beta}$ and therefore, by the definition~\eqref{eq:nearest-neighbors-definition} of nearest neighbors, 
	\begin{align} \label{eq:generic-properties-of-training-data-trf-1}
		\mathrm{NN}_{k_{\alpha},d}(x,\mathcal{D}_{\mathcal{X},y(x)}) < \infty
		\qquad \text{and} \qquad 
		\mathrm{NN}_{k_{\beta},d}(x,\mathcal{D}_{\mathcal{X},y}) < \infty.
	\end{align}
	Since moreover $d$ is a semimetric, we have $\sum_{x'\in X'} d(x,x') > 0$ for every non-empty subset $X'$ of $\mathcal{X} \setminus \{x\}$ and therefore the definition~\eqref{eq:nearest-neighbors-definition} of nearest neighbors further shows that
	\begin{align} \label{eq:generic-properties-of-training-data-trf-2}
		\mathrm{NN}_{k_{\alpha},d}(x,\mathcal{D}_{\mathcal{X},y(x)}) > 0
		\qquad \text{and} \qquad 
		\mathrm{NN}_{k_{\beta},d}(x,\mathcal{D}_{\mathcal{X},y}) > 0.
	\end{align}
	In view of~\cref{eq:attraction-and-repulsion-coefficients-definition,eq:generic-properties-of-training-data-trf-1,eq:generic-properties-of-training-data-trf-2}, the assertion~\eqref{eq:generic-properties-of-training-data-trf} is clear. And since  
	\begin{align*}
		f(x) = \sum_{y\in \mathcal{Y}\setminus\{y(x)\}} \Big( \alpha A_{k_{\alpha},d}(x) + \beta R_{k_{\beta},d}(x,y) \Big) \cdot (-p_y)
	\end{align*}
	by~(\ref{eq:barycenter-0-and-unit-distance-from-0}.a), we finally conclude from~(\ref{eq:conditions-on-hyperparameters}.a) and \eqref{eq:generic-properties-of-training-data-trf} in conjunction with~(\ref{eq:closure-and-interior-of-C_k}.b) that $f(x) \in \mathcal{C}_{y(x)}^{\circ}$, as desired.
\end{proof}

As an immediate consequence of the norm characterization~\eqref{eq:norm-char-of-C_k} of our cone segments, we obtain the  alternative representation~\eqref{eq:casimac-computation-appendix} of our classifier's class label predictions $\widehat{y}(x)$, which is much simpler computationally than the geometrically inspired definition~\eqref{eq:casimac-definition-appendix}. Additionally, we prove that for the training datapoints $x_i$, our classifier correctly predicts the true class label $y(x_i) = y_i$ provided that the regression model perfectly predicts its training datapoints $(x_i,f(x_i)) \in \mathcal{D}^f$. 

\begin{cor} \label{cor:casimac}
	Suppose the assumptions of the previous proposition are satisfied. Suppose further that $\widehat{f}: \mathcal{X} \to \mathcal{Z}$ is an arbitrary map and let $\widehat{y}: \mathcal{X} \to \mathcal{Y}$ be the map defined by
	\begin{align} \label{eq:casimac-definition-appendix}
		\widehat{y}(x) := \widehat{g}(\widehat{f}(x)) := \min\{y \in \mathcal{Y}: \widehat{f}(x) \in \mathcal{C}_y\} \qquad (x \in \mathcal{X}),
	\end{align}
	where $\widehat{g}(z) := \min\{y \in \mathcal{Y}: z \in \mathcal{C}_y\}$. Then
	\begin{align} \label{eq:casimac-computation-appendix}
		\widehat{y}(x) = \min\Big\{ y \in \mathcal{Y}: \big\|\widehat{f}(x) - p_y\big\|_2 = \min_{l\in\mathcal{Y}} \big\|\widehat{f}(x) - p_l\big\|_2 \Big\}
		\qquad (x \in \mathcal{X}).
	\end{align} 
	If, in addition, $\widehat{f}(x_i) = f(x_i)$ for all $i \in \{1,\dots,D\}$, then
	\begin{align} \label{eq:casimac-perfect-class-reconstruction-appendix}
		\widehat{y}(x_i) = y(x_i) = y_i \qquad (i \in \{1,\dots,D\}). 
	\end{align}
\end{cor}

\begin{proof}
	In view of the norm characterization~\eqref{eq:norm-char-of-C_k} of our cone segments, the assertion~\eqref{eq:casimac-computation-appendix} is obvious. 
	If $\widehat{f}(x_i) = f(x_i)$ for all $i$, then $\widehat{f}(x_i) = f(x_i) \in \mathcal{C}_{y(x_i)}^{\circ}$ 
	by Proposition~\ref{prop:generic-properties-of-training-data-trf}. Since $\mathcal{C}_{y(x_i)}^{\circ}$ by~(\ref{eq:segmentation-of-Z-appendix}.b) does not overlap with any other cone segment except for $\mathcal{C}_{y(x_i)}$ itself, we also obtain the assertion~\eqref{eq:casimac-perfect-class-reconstruction-appendix}.
\end{proof}

As a consequence of the essential disjointness~(\ref{eq:segmentation-of-Z-appendix}.b) of our cone segments, we obtain the convenient alternative representation~\eqref{eq:alternative-representation-of-class-probabilities} of our classifier's class label probability predictions $\widehat{p}(\,\cdot\,|x)$. 

\begin{cor} \label{cor:alternative-representation-of-class-probabilities}
	Suppose $p_1,\dots,p_n$ are the vertices of a regular $(n-1)$-simplex in $\mathcal{Z}$ with~\eqref{eq:barycenter-0-and-unit-distance-from-0} and let $\mathcal{C}_1, \dots, \mathcal{C}_n$ be the cone segments defined in~\eqref{eq:def-of-C_k}. Suppose further that $\widehat{q}(\,\cdot\,|x)$ for every $x \in \mathcal{X}$ is a probability density on $\mathcal{Z}$ and let
	\begin{align} \label{eq:def-of-class-probabilities}
		\widehat{p}(y|x) := \int_{\{\widehat{g} = y\}} \widehat{q}(z|x) \d z
		\qquad (x \in \mathcal{X} \text{ and } y \in \mathcal{Y}), 
	\end{align}
	where $\widehat{g}(z) := \min\{y \in \mathcal{Y}: z \in \mathcal{C}_y\}$. Then
	\begin{align} \label{eq:alternative-representation-of-class-probabilities}
		\widehat{p}(y|x) = \int_{\mathcal{C}_y} \widehat{q}(z|x) \d z = \int_{\mathcal{C}_y^{\circ}} \widehat{q}(z|x) \d z
		\qquad (x \in \mathcal{X} \text{ and } y \in \mathcal{Y}). 
	\end{align}
\end{cor}

\begin{proof}
	As a first step, we observe that 
	\begin{align} \label{eq:alternative-representation-of-class-probabilities-1}
		\mathcal{C}_y^{\circ} \subset \{z \in \mathcal{Z}: \widehat{g}(z) = y\} \subset \mathcal{C}_y
	\end{align}
	for every $y \in \mathcal{Y}$. Indeed, the second inclusion is just a trivial consequence of the definition of $\widehat{g}$ and the first inclusion directly follows from the fact that $\mathcal{C}_y^{\circ}$ by~(\ref{eq:segmentation-of-Z-appendix}.b) does not overlap with any cone segment except $\mathcal{C}_y$ itself.  
	As a second step, we show that  
	\begin{align} \label{eq:alternative-representation-of-class-probabilities-2}
		\partial \mathcal{C}_y = \ol{\mathcal{C}}_y \setminus \mathcal{C}_y^{\circ} = \mathcal{C}_y \setminus \mathcal{C}_y^{\circ}
		\subset \bigcup_{k \in \mathcal{Y}\setminus \{y\}} \mathcal{M}_{y k}
	\end{align}
	for every $y \in \mathcal{Y}$, where $\mathcal{M}_{y k}$ as before is the mid-perpendicular hyperplane between the vertices $p_y$ and $p_k$. In order to see this, let $y \in \mathcal{Y}$ be fixed and let $z \in \partial \mathcal{C}_y = \mathcal{C}_y \setminus \mathcal{C}_y^{\circ}$. We then have, on the one hand, that
	\begin{align} \label{eq:alternative-representation-of-class-probabilities-3}
		\norm{z-p_y}_2 = \min_{l\in \mathcal{Y}} \norm{z-p_l}_2 
	\end{align}
	by~\eqref{eq:norm-char-of-C_k} and, on the other hand, this minimum must be attained also at another index $k \ne y$, that is, there must be a $k \in \mathcal{Y} \setminus \{y\}$ with
	\begin{align} \label{eq:alternative-representation-of-class-probabilities-4}
		\norm{z-p_k}_2 = \min_{l\in \mathcal{Y}} \norm{z-p_l}_2.
	\end{align}
	(If the minimum was attained only for the index $y$, this would mean that $\norm{z-p_y}_2 < \norm{z-p_l}_2$ for all $l \ne y$ and therefore we would have  $z \in \mathcal{C}_y^{\circ}$ by~\eqref{eq:norm-char-of-C_k^circ}. Contradiction to our choice of $z$!) Combining~\eqref{eq:alternative-representation-of-class-probabilities-3} and~\eqref{eq:alternative-representation-of-class-probabilities-4}, we see that 
	\begin{align}
		\norm{z-p_y}_2 = \norm{z-p_k}_2
	\end{align}
	for some $k \in \mathcal{Y} \setminus \{y\}$. So, by~\eqref{eq:def-M_kl} and~\eqref{eq:norm-char-of-C_k-1.1}, we see that $z$ must lie on the mid-perpendicular hyperplane $\mathcal{M}_{yk}$. And thus~\eqref{eq:alternative-representation-of-class-probabilities-2} is proven. 
	Since hyperplanes are null sets \wrt~Lebesgue measure on $\mathcal{Z}$, we further conclude from~\eqref{eq:alternative-representation-of-class-probabilities-2} that the boundary $\partial \mathcal{C}_y$ of every cone segment is a null set as well. Combining this, in turn, with~\eqref{eq:alternative-representation-of-class-probabilities-1}, we immediately obtain the alternative representations of $\widehat{p}(y|x)$ from~\eqref{eq:alternative-representation-of-class-probabilities}.
\end{proof}

In the following, we turn to the computation of our class label probability predictions~\eqref{eq:def-of-class-probabilities}, using standard Monte Carlo approximation techniques. In principle, the following approximation result is true for completely arbitrary probability density functions $\widehat{q}(\,\cdot\,|x)$ on $\mathcal{Z}$, but it does require to draw samples from $\widehat{q}(\,\cdot\,|x)$. In the special case of normal-distribution densities $\widehat{q}(\,\cdot\,|x)$, drawing samples is standard, of course (Section~3.24 of~\citep{fishman1996}, for instance). 

\begin{prop} \label{prop:computation-of-class-probabilities}
	Suppose $p_1,\dots,p_n$ are the vertices of a regular $(n-1)$-simplex in $\mathcal{Z}$ with~\eqref{eq:barycenter-0-and-unit-distance-from-0} and let $\mathcal{C}_1, \dots, \mathcal{C}_n$ be the cone segments defined in~\eqref{eq:def-of-C_k}. Suppose further that $\widehat{q}(\,\cdot\,|x)$ for every $x \in \mathcal{X}$ is a probability density on $\mathcal{Z}$ and let
	\begin{align} \label{eq:class-probabilities}
		\widehat{p}(y|x) := \int_{\mathcal{C}_y} \widehat{q}(z|x) \d z = \int_{\mathcal{Z}} I_{\mathcal{C}_y}(z) \widehat{q}(z|x) \d z
		\qquad (x \in \mathcal{X} \text{ and } y \in \mathcal{Y}), 
	\end{align}
	where $I_{\mathcal{C}_y}$ is the indicator function of the cone segment $\mathcal{C}_y$ in analogy to \eqref{eqn:indicator}.
	In the special case where $n=2$ 
	and where $\widehat{q}(\,\cdot\,|x)$ is normally distributed with mean $\widehat{\mu}(x)$ and variance $\widehat{\sigma}(x)^2$, the class probability predictions $\widehat{p}(y|x)$ have the closed-form representation
	\begin{align} \label{eq:class-probability-closed-form}
		\widehat{p}(1|x) = \frac{1}{2}\Big(1 - \operatorname{erf}\big(\widehat{\mu}(x)/(\sqrt{2}\, \widehat{\sigma}(x))\big)\Big)
		= 1-\widehat{p}(2|x),
	\end{align}  
	where the convention~\eqref{eq:vertex-indexing-in-binary-case} was used. 
	In the general case, the class probability prediction $\widehat{p}(y|x)$, for every fixed $x \in \mathcal{X}$ and $y \in \mathcal{Y}$, can be approximated with probability $1$ by the sample means
	\begin{align} \label{eq:class-probability-approximant}
		\widehat{p}_N(y|x) := \frac{1}{N} \sum_{i=1}^N I_{\mathcal{C}_y}(z_i),
	\end{align}
	where $z_1, \dots, z_N$ are sampled independently from the probability density $\widehat{q}(\,\cdot\,|x)$ and where the binary values $I_{\mathcal{C}_y}(z_i)$ are determined by means of the norm characterization~\eqref{eq:norm-char-of-C_k} of the cone segment $\mathcal{C}_y$. Additionally, for every fixed sample size $N \in \N$ and every $\eps > 0$, the probability for $\widehat{p}_N(y|x)$ to miss $\widehat{p}(y|x)$ by more than $\eps$ is bounded above by $1/(4 \eps^2 N)$. 
\end{prop}

\begin{proof}
	In the mentioned special case, we have $\mathcal{C}_1 = (-\infty,0]$ and $\mathcal{C}_2 = [0,\infty)$ by our convention~\eqref{eq:vertex-indexing-in-binary-case} and, with this, the asserted closed-form expression immediately follows. 
	We therefore move on to discuss the general case now. We point out that all the arguments to come are fairly standard and we give them just for the reader's convenience. So, let $x \in \mathcal{X}$ and $y \in \mathcal{Y}$ be fixed, write $Q = Q_x$ for the probability measure on $\mathcal{Z}$ with density $\widehat{q}(\,\cdot\,|x)$, and let $z_1, \dots, z_N$ be independent random variables with distribution $Q$. Writing
	\begin{align}
		\widehat{p} := \widehat{p}(y|x)
		\qquad \text{and} \qquad
		h := h_y := I_{\mathcal{C}_y}
		\qquad \text{and} \qquad
		\widehat{p}_N := \widehat{p}_N(y|x) := \frac{1}{N} \sum_{i=1} h\circ z_i,
	\end{align}
	we can express the expectation value of $h \circ z_i$ \wrt~$Q$ as 
	\begin{align} \label{eq:computation-of-class-probabilities-1}
		\E(h\circ z_i) = \int_{\mathcal{Z}}h(z) \widehat{q}(z|x) \d z = \widehat{p}
	\end{align} 
	for every $i$ and, since $h^2 = h$, we see that the variance of $h \circ z_i$ \wrt~$Q$ is given by 
	\begin{align} \label{eq:computation-of-class-probabilities-2}
		\Var(h\circ z_i) = \E((h\circ z_i)^2) - (\E(h\circ z_i))^2 = \E(h\circ z_i) - (\E(h\circ z_i))^2 = \widehat{p}(1-\widehat{p})
	\end{align}
	for every $i$. Since the $z_i$ are independent and $h$ is bounded, the random variables $h\circ z_i$ are independent and (square) integrable and therefore the strong law of large numbers in conjunction with~\eqref{eq:computation-of-class-probabilities-1} shows that, as $N \to \infty$, the sample means $\widehat{p}_N$ converge to $\widehat{p}$ with $Q$-probability $1$ (or, put differently, $Q$-almost surely). 
	Also, Bienaym\'{e}'s identity in conjunction with~\eqref{eq:computation-of-class-probabilities-2} shows that 
	\begin{align}
		\Var(\widehat{p}_N) = \frac{1}{N^2} \sum_{i=1}^N \Var(h\circ z_i) = \frac{\widehat{p}(1-\widehat{p})}{N}
	\end{align}  
	for every fixed $N \in \N$. So, by Chebyshev's inequality, we get the upper bound
	\begin{align} \label{eq:computation-of-class-probabilities-3}
		Q(\{|\widehat{p}_N-\widehat{p}| > \eps\}) \le \frac{1}{\eps^2} \int |\widehat{p}_N - \widehat{p}|^2 \d Q = \frac{\Var(\widehat{p}_N)}{\eps^2} = \frac{\widehat{p}(1-\widehat{p})}{\eps^2 N}
	\end{align}
	on the $Q$-probability for $\widehat{p}_N(y|x)$ to miss $\widehat{p}(y|x)$ by more than $\eps$. 
	Since $\widehat{p}$ by definition lies between $0$ and $1$, the numerator on the right-hand side of~\eqref{eq:computation-of-class-probabilities-3} can be trivially estimated as $\widehat{p}(1-\widehat{p}) \le 1/4$. Inserting this estimate into~\eqref{eq:computation-of-class-probabilities-3}, we obtain 
	\begin{align} \label{eq:computation-of-class-probabilities-4}
		Q(\{|\widehat{p}_N-\widehat{p}| > \eps\}) \le \frac{1}{4 \eps^2 N}
	\end{align}
	for every $N \in \N$ and $\eps > 0$, which is precisely the asserted probabilistic bound on the approximation error. 
\end{proof}

In addition to the bound~\eqref{eq:computation-of-class-probabilities-4}, one could also apply the central limit theorem to get approximate $95 \%$-confidence intervals in terms of the sample means~\eqref{eq:class-probability-approximant} and the (biased) sample variances
\begin{align*}
	\widehat{\sigma}_N(y|x)^2 := \frac{1}{N} \sum_{i=1}^N \big( h\circ z_i - \widehat{p}_N(y|x) \big)^2
	= \widehat{p}_N(y|x) \big(1-\widehat{p}_N(y|x) \big)
\end{align*}   
for sufficiently large sample sizes $N$. See~\citep{madras2002} (Section~5.1), for instance. It should be noticed, however, that it is not so straightforward to determine what actually is a sufficiently large sample size $N$. 
A detailed discussion of this topic can be found~\citep{fishman1996} (Chapter~2).

\subsection{Compressing the latent space to a reference simplex}

In this section, we show how the latent space can be compressed to a reference simplex in a diffeomorphic and cone-segment-preserving manner. In order to do so, we need a lemma on barycentric coordinates.

\begin{lm} \label{lm:barycentric-coordinates}
	Suppose $p_1,\dots,p_n$ are the vertices of an $(n-1)$-simplex $\mathcal{S}$ in $\mathcal{Z}$. Then for every point $z \in \mathcal{S}$ there exist unique numbers $\lambda_i = \lambda_i(z) \in [0,\infty)$ for $i \in \mathcal{Y}$, the so-called barycentric coordinates of $z$ \wrt~$\mathcal{S}$, such that
	\begin{align} \label{eq:barycentric-coordinates}
		z = \sum_{i\in\mathcal{Y}} \lambda_i \cdot p_i 
		\qquad \text{and} \qquad
		\sum_{i\in\mathcal{Y}} \lambda_i = 1
	\end{align}
	Also, a point $z \in \mathcal{S}$ belongs to the interior $\mathcal{S}^{\circ}$ of $\mathcal{S}$ if and only if all its barycentric coordinates $\lambda_1(z), \allowbreak \dots, \allowbreak \lambda_n(z)$ \wrt~$\mathcal{S}$ are strictly positive. And finally, the barycentric-coordinate map $\mathcal{S} \ni z \mapsto (\lambda_1(z),\dots, \lambda_n(z))$ is infinitely differentiable. 
\end{lm}

\begin{proof}
	Consider the map $\phi: \R^{n-1} \to \mathcal{Z}$ defined by $\phi(\lambda) := \ul{\phi}(\lambda) + p_n$ for all $\lambda \in \R^{n-1}$, where $\ul{\phi}$ is the linear map with $\ul{\phi}(e_i) = p_i-p_n$ for all $i \in \{1,\dots,n-1\}$ and $e_1, \dots, e_{n-1}$ denote the canonical unit vectors in $\R^{n-1}$. It is then clear that 
	\begin{align} \label{eq:barycentric-coordinates-1}
		\phi(\lambda) = \sum_{i=1}^{n-1} \lambda_i \ul{\phi}(e_i) + p_n
		= \sum_{i=1}^{n-1} \lambda_i \cdot p_i + \bigg(1 - \sum_{i=1}^{n-1} \lambda_i\bigg) \cdot p_n
	\end{align}
	for every $\lambda = \sum_{i=1}^{n-1} \lambda_i e_i \in \R^{n-1}$ and from this, in turn, it immediately follows that $\mathcal{S} = \conv\{p_1, \dots, p_n\}$ is the image under $\phi$ of the lower-left corner $\Delta := \{\lambda \in [0,\infty)^{n-1}: \sum_{i=1}^{n-1} \lambda_i \le 1\}$ of the unit cube in $\R^{n-1}$. In short, 
	\begin{align} \label{eq:barycentric-coordinates-2}
		\mathcal{S} = \phi(\Delta). 
	\end{align} 
	It also immediately follows from the linear independence of $\{\ul{\phi}(e_i): i \in \{1,\dots,n-1\}\}$ that $\ul{\phi}$ is a linear isomorphism of $\mathcal{Z} = \R^{n-1}$ and thus also a diffeomorphism of $\mathcal{Z}$ (Theorem~VII.1.6 in conjunction with Example~VII.2.3 (a) and Exercise~VII.5.1 of~\citep{amann2008}). 
	Consequently, the translate $\phi$ of $\ul{\phi}$ is a diffeomorphism of $\mathcal{Z}$ as well. 
	With these preliminary observations, the assertions of the lemma immediately follow. Indeed, by~\eqref{eq:barycentric-coordinates-1} and the bijectivity of $\phi$, the barycentric coordinates of any given $z \in \mathcal{S}$ are uniquely determined by $z$, namely 
	\begin{align} \label{eq:barycentric-coordinates-3}
		(\lambda_1(z),\dots, \lambda_{n-1}(z)) = \phi^{-1}(z)
		\qquad \text{and} \qquad
		\lambda_n(z) = 1 - \sum_{i=1}^{n-1} \lambda_i(z)
	\end{align} 
	for every $z \in \mathcal{S}$. In view of~\eqref{eq:barycentric-coordinates-3} and the infinite differentiability of $\phi^{-1}$, in turn, the asserted infinite differentiability of $\mathcal{S} \ni z \mapsto (\lambda_1(z),\dots, \lambda_n(z))$ follows. And finally, since the interior of $\Delta$ is obviously given by $\Delta^{\circ} = \{\lambda \in (0,\infty)^{n-1}: \sum_{i=1}^{n-1} \lambda_i < 1\}$, we see by~\eqref{eq:barycentric-coordinates-2} and~\citep{dugundji1966} (Theorem~III.11.3) and~\eqref{eq:barycentric-coordinates-1} that
	\begin{align*}
		\mathcal{S}^{\circ} = (\phi(\Delta))^{\circ} = \phi(\Delta^{\circ}) 
		= \big\{z \in \mathcal{Z}: \eqref{eq:barycentric-coordinates} \text{ holds true with } \lambda_1, \dots, \lambda_n \in (0,\infty)\big\}
	\end{align*} 
	which, in turn, yields the asserted characterization of the interior points of $\mathcal{S}$. 
\end{proof}

\begin{prop} \label{prop:compression-of-Z}
	Suppose $p_1,\dots,p_n$ are the vertices of a regular $(n-1)$-simplex in $\mathcal{Z}$ with~\eqref{eq:barycenter-0-and-unit-distance-from-0} and let $\mathcal{C}_1, \dots, \mathcal{C}_n$ be the cone segments defined in~\eqref{eq:def-of-C_k}. Then the compression map $C: \mathcal{Z} \to \mathcal{S}^{\circ}$, defined by
	\begin{align} \label{eq:def-of-compression-map}
		C(z) := \sum_{i\in\mathcal{Y}} \mu_i(z) \cdot p_i
		\quad \text{with} \quad
		\mu_i(z) := \exp(\tau p_i \cdot z)/\Big(\sum_{j\in\mathcal{Y}}\exp(\tau p_j \cdot z)\Big)
	\end{align}
	and a fixed $\tau \in (0,\infty)$, is a diffeomorphism of $\mathcal{Z}$ onto $\mathcal{S}^{\circ}$. Its inverse is given by the inflation map $I: \mathcal{S}^{\circ} \to \mathcal{Z}$, defined by
	\begin{align} \label{eq:def-of-inflation-map}
		I(w) := \frac{n-1}{\tau n}\sum_{i\in\mathcal{Y}} \ln(\lambda_i(w)) \cdot p_i 
	\end{align}
	with $\lambda_1(w), \dots, \lambda_n(w)$ being the barycentric coordinates of $w$ \wrt~$\mathcal{S}$ and $\tau$ is the same positive number as in~\eqref{eq:def-of-compression-map}. Additionally, $C$ and $I$ leave the segments $\mathcal{C}_1, \dots, \mathcal{C}_n$ invariant. 
\end{prop}

\begin{proof}
	It is clear by the definitions~\eqref{eq:def-of-compression-map} and~\eqref{eq:def-of-inflation-map} and by Lemma~\ref{lm:barycentric-coordinates} that the compression $C$ is an  infinitely differentiable map from $\mathcal{Z}$ into $\mathcal{S}^{\circ}$ and that, conversely, the inflation $I$ is an infinitely differentiable map from $\mathcal{S}^{\circ}$ into $\mathcal{Z}$.
	We now show that $I$ is the inverse of $C$ or, equivalently, that
	\begin{align} \label{eq:compression-of-Z-1}
		C(I(w)) = w \qquad (w \in \mathcal{S}^{\circ})
		\qquad \text{and} \qquad
		I(C(z)) = z \qquad (z \in \mathcal{Z}). 
	\end{align}
	In order to verify~(\ref{eq:compression-of-Z-1}.a), we confirm with the help of Lemma~\ref{lm:central-angle-of-reg-simplex} that
	\begin{align*}
		\tau p_i \cdot I(w) = \ln \lambda_i(w) - \frac{1}{n} \sum_{j\in\mathcal{Y}} \ln \lambda_j(w)
		\qquad (w \in \mathcal{S}^{\circ})
	\end{align*}
	and from this, in turn, we arrive at~(\ref{eq:compression-of-Z-1}.a) in a straightforward manner.
	In order to verify~(\ref{eq:compression-of-Z-1}.b), we first notice  that the positive numbers $\mu_1(z), \dots, \mu_n(z)$ from~\eqref{eq:def-of-compression-map} are nothing but the barycentric coordinates $\lambda_1(C(z)), \dots, \lambda_n(C(z))$ of the point $C(z) \in \mathcal{S}^{\circ}$ \wrt~$\mathcal{S}$ and, thus, 
	\begin{align} \label{eq:compression-of-Z-2}
		I(C(z)) = \frac{n-1}{\tau n}\sum_{i\in\mathcal{Y}} \ln(\mu_i(z)) \cdot p_i = \frac{n-1}{n}\sum_{i\in\mathcal{Y}} (p_i \cdot z) \, p_i
		\qquad (z\in\mathcal{Z}),
	\end{align}
	where for the second equality we used (\ref{eq:barycenter-0-and-unit-distance-from-0}.a).
	With the help of Lemma~\ref{lm:central-angle-of-reg-simplex} and again~(\ref{eq:barycenter-0-and-unit-distance-from-0}.a), we further see that $\frac{n-1}{n}\sum_{i\in\mathcal{Y}} (p_i \cdot p_j) \, p_i = p_j$ for every $j \in \mathcal{Y}$ and, by linearly extending this identity to arbitrary $z = \sum_{j\in\mathcal{Y}} r_j p_j$ (Lemma~\ref{lm:vertices-lin-independent}), we conclude that the right-hand side of~\eqref{eq:compression-of-Z-2} equals $z$, as desired. So, \eqref{eq:compression-of-Z-1} is proved, that is, $I = C^{-1}$ is the inverse of $C$. In particular, by the infinite differentiability of $C$ and $I$ pointed out above, $C$ is a diffeomorphism from $\mathcal{Z}$ onto $\mathcal{S}^{\circ}$.
	It thus only remains to prove that $C$ and $I$ leave the segments $\mathcal{C}_1, \dots, \mathcal{C}_n$ invariant or, in other words, that
	\begin{align} \label{eq:compression-of-Z-3}
		C(\mathcal{C}_k) = \mathcal{C}_k \cap \mathcal{S}^{\circ} 
		\qquad \text{and} \qquad
		I(\mathcal{C}_k \cap \mathcal{S}^{\circ}) = \mathcal{C}_k
	\end{align}
	for every $k \in \mathcal{Y}$. In order to see this, we notice that 
	\begin{align*}
		C(z) = \sum_{i\in\mathcal{Y}} \mu_i(z) \cdot p_i = \sum_{i\in\mathcal{Y}} (-\mu_i(z)) \cdot (-p_i)
	\end{align*}
	and thus, by Lemma~\ref{lm:barycentric-char-of-C_k} and~\eqref{eq:def-of-compression-map} and Corollary~\ref{cor:scalar-product-char-of-C_k}, the following chain of equivalences holds true for every $k \in \mathcal{Y}$ and $z \in \mathcal{Z}$:
	\begin{align*} 
		C(z) \in \mathcal{C}_k 
		\quad \text{iff} \quad 
		-\mu_k(z) = \min_{l\in\mathcal{Y}} (-\mu_l(z)) 
		\quad \text{iff} \quad 
		p_k\cdot z = \max_{l\in\mathcal{Y}} p_l \cdot z 
		\quad \text{iff} \quad 
		z \in \mathcal{C}_k.
	\end{align*}
	In short, $C(z) \in \mathcal{C}_k$ if and only if $z \in \mathcal{C}_k$. And this, in turn, immediately implies the asserted invariance statements~\eqref{eq:compression-of-Z-3}. 
\end{proof}

\section{Implementation}
\label{sect:implementation}

A Python implementation of our proposed method is available online~\citep{casimac-github}. The implementation includes the following main functionality, which is outlined in \cref{alg:casimac}. A sketch of the functionality is presented in \cref{fig:implementation}.
\begin{itemize}
	\item \textproc{Train} (\cref{alg:casimac:train}): Calculate the simplex vertices $p_1, \dots, p_n$ and train the regression model based on a training data set $\mathcal{D}$, the user-defined parameters $\alpha$, $\beta$, $k_{\alpha}$, $k_{\beta}$, and $d$ for the training data transformation $f$ and the hyperparameters $h$ of the regression model. In this manner, obtain a point prediction $\widehat{f}(x)$ -- possibly along with a probabilistic prediction $\widehat{q}(\,\cdot\,|x)$ -- for every $x\in \mathcal{X}$. For the calculation of the simplex vertices $p_1, \dots, p_n$, we provide two methods. First, the explicit method via \cref{eq:sample-reg-simplex-in-Z-7} and second, an iterative method. For the latter, we choose $p_1$ to be the first canonical basis vector and determine the remaining elements of $p_2, \dots, p_n$ one after another based on the already specified elements such that (\ref{eq:barycenter-0-and-unit-distance-from-0}.b) and (\ref{eq:simplex-central-angle-of-reg-simplex}.b) are fulfilled. The resulting simplex vectors from the two methods differ only by a joint rotation so that there is no substancial effect on our method apart from minor numerical differences. All results in this paper are based on simplex vertices from the iterative method.
	\item \textproc{PredictClassLabel} (\cref{alg:casimac:predictlabel}): Predict the class label of (possibly unseen) feature space points $x \in \mathcal{X}$ based on the point prediction $\widehat{f}(x)$ of the trained regression model and on the simplex vertices $p_1, \dots, p_n$.
	\item \textproc{PredictClassLabelProbability} (\cref{alg:casimac:predictproba}): Predict the probability of class label $y \in \mathcal{Y}$ for unseen feature space points $x \in \mathcal{X}$ based on the probabilistic prediction $\widehat{q}(\,\cdot\,|x)$ of the trained regression model, 
	on the simplex vertices $p_1, \dots, p_n$, and on a fixed sample size $N$. We presume that $\widehat{q}(\,\cdot\,|x)$ is normally distributed with mean $\widehat{\mu}(x)$ and variance $\widehat{\sigma}(x)^2$.
	\item \textproc{Compress} (\cref{alg:casimac:transform}): Transform the latent space point $z \in \mathcal{Z}$ to reference-simplex counterpart $w = C(z) \in -\mathcal{S}^{\circ}$ based on the user-defined parameter $\tau$ and the simplex vertices $p_1, \dots, p_n$.
	\item \textproc{Inflate} (\cref{alg:casimac:invtransform}): Transform the reference-simplex point $w \in -\mathcal{S}^{\circ}$ to latent space counterpart $z = I(w) \in \mathcal{Z}$ based on the user-defined parameter $\tau$ and the simplex vertices $p_1, \dots, p_n$.
\end{itemize}

In particular, our implementation is intended to be used with scikit-sklearn~\citep{sklearn2011}. We therefore do not prescribe the specific form of the regression model, but instead allow the user to plug in an arbitrary scikit-sklearn estimator. 
\begin{algorithm}[!htb]
	\caption{Summary of the main functionality of the provided implementation~\citep{casimac-github}.} \label{alg:casimac}
	\begin{algorithmic}[1]
		\Function{Train}{$\mathcal{D}$, $\alpha$, $\beta$, $k_{\alpha}$, $k_{\beta}$, $d$, $h$} \label{alg:casimac:train}
		\State Calculate simplex vertices $p_1, \dots, p_n$ either explicitly according to~\cref{eq:sample-reg-simplex-in-Z-7} using~\cref{eq:sample-reg-simplex-in-Z-6,eq:sample-reg-simplex-in-Z-8} or iteratively as described in \cref{sect:implementation}.
		\State Calculate $f(x_i) = f_{\alpha,\beta,k_{\alpha},k_{\beta},d}(x_i|\mathcal{D})$ for $i \in \{1,\dots, D\}$ using~\cref{eq:training-data-trf-defining-formula,eq:attraction-and-repulsion-coefficients-definition,eq:nearest-neighbors-definition,eq:metric-d,eq:conditions-on-hyperparameters} 
		\State Calculate the transformed training data set $\mathcal{D}^f$ from \eqref{eq:transformed-training-data} 
		\State Train a regression model with hyperparameters $h$, by fitting it to $\mathcal{D}^f$ 
		\EndFunction
		
		\Function{PredictClassLabel}{$x$, $\widehat{f}$, $p_1, \dots, p_n$} \label{alg:casimac:predictlabel}
		\State Calculate the regression model's point prediction $\widehat{z} = \widehat{f}(x)$ for $x$
		\State Calculate the distances $\norm{\widehat{z}-p_1}_2, \dots, \norm{\widehat{z}-p_n}_2$ 
		\State Calculate the label $\widehat{y} = \widehat{g}(\widehat{z})$ based on the above distances and~\eqref{eq:casimac-alternative-definition}
		\State \Return $\widehat{y}$ 
		\EndFunction
		
		\Function{PredictClassLabelProbability}{$x$, $y$, $\widehat{\mu}$, $\widehat{\sigma}$, $p_1, \dots, p_n$, $N$} \label{alg:casimac:predictproba}
		\If {$n=2$}
		\State Calculate $\widehat{p} = \widehat{p}(y|x)$ according to~\eqref{eq:class-probability-closed-form}
		\Else
		\State Sample $z_1, \dots, z_N$ independently from $\widehat{q}(\,\cdot\,|x)$
		\State Calculate the distances $\norm{z_i-p_1}_2, \dots, \norm{z_i-p_n}_2$ for $i \in \{1,\dots,N\}$
		\State Calculate $I_{\mathcal{C}_y}(z_1),\dots, I_{\mathcal{C}_y}(z_N)$ based on the above distances and on~\eqref{eq:norm-char-of-C_k}
		\State Calculate $\widehat{p} = \widehat{p}_N(y|x)$ according to~\eqref{eq:class-probability-approximant} 
		\EndIf
		\State \Return $\widehat{p}$
		\EndFunction
		
		\Function{Compress}{$z$, $\tau$, $p_1, \dots, p_n$} \label{alg:casimac:transform}
		\State Calculate $w = C(z)$ according to~\eqref{eq:def-of-compression-map}
		\State \Return $w$	
		\EndFunction
		
		\Function{Inflate}{$w$, $\tau$, $p_1, \dots, p_n$} \label{alg:casimac:invtransform}
		\State Calculate $z = I(w)$ according to~\eqref{eq:def-of-inflation-map}
		\State \Return $z$	
		\EndFunction
	\end{algorithmic}
\end{algorithm}

\begin{figure}
	\begin{center}
		\includegraphics{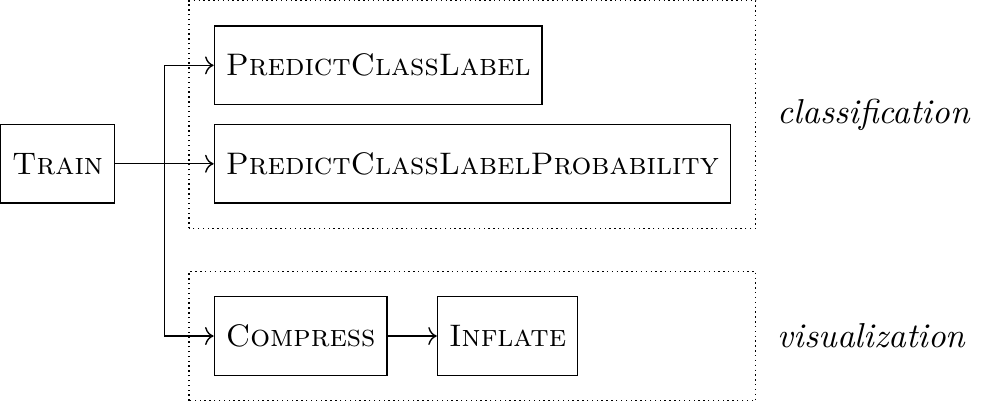}
		\caption{An overview of the functionality of the provided implementation~\citep{casimac-github} as outlined in \cref{alg:casimac}. After the initial training stage, there are two possible use cases, namely classification and visualization, as described in detail in \cref{sect:method}. The first use case refers to the prediction of class labels and of class probabilities for unseen feature space points, whereas the second use case allows to perform a transformation from the latent space to a reference simplex and back for visualization purposes.} \label{fig:implementation}
	\end{center}
\end{figure}

\section{Hyperparameters}
\label{sect:hyperparameters}

In \cref{sect:validation} we use cross-validation to select the best hyperparameters for the classifiers out of a pre-defined set of possible choices. Specifically, we vary the number of nearest neighbors $k \in \{ \num{5},\allowbreak \num{10},\allowbreak \num{15},\allowbreak \num{20} \}$ for kNN, the hidden layer sizes $L \in \{ (\num{5}),\allowbreak (\num{10}),\allowbreak (\num{5},\num{5}), (\num{5},\num{10}),\allowbreak (\num{10},\num{10}),\allowbreak (\num{5},\num{5},\num{5}),\allowbreak (\num{5},\num{5},\num{10}),\allowbreak (\num{5},\num{10},\num{10}),\allowbreak (\num{10},\num{10},\num{10}) \}$ (all with a rectified linear unit as their activation function) for MLP and the kernel parameters for both GPC and the GPR model used for CASIMAC. We choose a sum of a Mat\'{e}rn kernel and a white-noise kernel~\citep{rasmussen2006} for these kernels and tune the Mat\'{e}rn coefficient $\nu \in \{ \frac{3}{2}, \frac{5}{2}, \infty \}$. The remaining kernel parameters are optimized for each cross-validation setup as described in \cite{rasmussen2006}. Additionally, for CASIMAC, we tune both $\gamma$ from~\eqref{eq:gamma} as well as $k_{\alpha}$ and $k_{\beta}$ according to~\eqref{eq:conditions-on-hyperparameters}.
In \cref{tab:app:hyper-params}, we list the cross-validated hyperparameters which result in the best accuracy over all classification tasks for each data set from \cref{sect:real-world-data}. Analogously, for the synthetic data set from \cref{sect:synthetic-data} we get the best hyperparameters $\gamma=1 $, $k_\alpha = 10$, $k_\beta = 10$, and $\nu=\infty$ for CASIMAC and $\nu=\frac{3}{2}$ for GPC. And finally, for the \texttt{alcohol-3} data set from \cref{sect:visualization-example} $\gamma=\frac{1}{2}$, $k_\alpha = 1$, and $k_\beta = 5$ are fixed and we find $\nu=\infty$ for CASIMAC from the cross-validation.

\begin{table}
	\centering
	\caption{The best cross-validated hyperparameters for the real-world data sets from \cref{tab:datasets}.}\label{tab:app:hyper-params}
	\begin{tabular}{cccccc}\toprule
		& \texttt{alcohol} & \texttt{climate} & \texttt{hiv} & \texttt{pine} & \texttt{wifi} \\ \midrule
		\STAB{\rotatebox[origin=c]{90}{\hspace{.1cm}\textbf{CASIMAC}\hspace{.1cm}}}%
		& \makecell{$\gamma=0 $ \\ $k_\alpha = 5$ \\ $k_\beta = 5$ \\ $\nu=\infty$} &
		\makecell{$\gamma=0 $ \\ $k_\alpha = 5$ \\ $k_\beta = 5$ \\ $\nu=\infty$} &
		\makecell{$\gamma=0 $ \\ $k_\alpha = 20$ \\ $k_\beta = 20$ \\ $\nu=\frac{3}{2}$} &
		\makecell{$\gamma=1 $ \\ $k_\alpha = 1$ \\ $k_\beta =5$ \\ $\nu=\frac{3}{2}$} &
		\makecell{$\gamma=1/3$ \\ $k_\alpha = 20$ \\ $k_\beta = 20$ \\ $\nu=\infty$} \\ 
		\STAB{\rotatebox[origin=c]{90}{\hspace{.1cm}\textbf{GPC}\hspace{.1cm}}}%
		& $\nu=\frac{3}{2}$ &
		$\nu=\frac{3}{2}$ &
		$\nu=\infty$ &
		$\nu=\frac{3}{2}$ &
		$\nu=\frac{3}{2}$ \\ 
		\STAB{\rotatebox[origin=c]{90}{\hspace{.1cm}\textbf{kNN}\hspace{.1cm}}}%
		& $k=\num{5}$ &
		$k=\num{5}$ &
		$k=\num{5}$ &
		$k=\num{5}$ &
		$k=\num{5}$ \\ 
		\STAB{\rotatebox[origin=c]{90}{\hspace{.1cm}\textbf{MLP}\hspace{.1cm}}}%
		& \makecell{$L=$ \\ $(\num{10},\num{10},\num{10})$} &
		\makecell{$L=$ \\ $(\num{5}, \num{5})$} &
		\makecell{$L=$ \\ $(\num{5}, \num{10})$} &
		\makecell{$L=$ \\ $(\num{10},\num{10},\num{10})$} &
		\makecell{$L=$ \\ $(\num{5}, \num{5})$} \\ \bottomrule
	\end{tabular}
\end{table}
\end{appendix}
\FloatBarrier
\bibliography{literature}
} 

\pgfplotsset{compat=newest}
\usetikzlibrary{shapes.geometric}

\usepackage{titling}
\pretitle{\begin{center}\sffamily\LARGE}
\preauthor{\begin{center}
		\large\sffamily \lineskip 0.5em%
		\begin{tabular}[t]{c}}
\predate{\begin{center}\sffamily\large}

\usepackage{abstract}

\hypersetup{hidelinks}

\newcommand{\ie}{i.\,e.\xspace}
\newcommand{\eg}{e.\,g.\xspace}

\newcommand{\wrt}{w.r.t.\xspace}

\newcommand{\bnum}[1]{\textbf{\num{#1}}}

\newcommand{\norm}[1]{\left\| #1 \right\|}  
  
\renewcommand{\d}{\,\mathrm{d}} 
\newcommand{\e}{\mathrm{e}}

\newcommand{\N}{\mathbb{N}}

\newcommand{\R}{\mathbb{R}}

\newcommand{\spn}{\operatorname{span}}
\newcommand{\cnv}{\operatorname{conv}}

\DeclareMathOperator*{\argmax}{argmax}

\newcommand{\conv}{\operatorname{conv}}

\newcommand{\eps}{\varepsilon}
\renewcommand{\phi}{\varphi}
\newcommand{\ul}{\underline}
\newcommand{\ol}{\overline}

\newcommand{\E}{\operatorname{E}}
\newcommand{\Var}{\operatorname{Var}}

\newtheorem{thm}{Theorem}
\newtheorem{cor}[thm]{Corollary}
\newtheorem{prop}[thm]{Proposition}
\newtheorem{lm}[thm]{Lemma}

\theoremstyle{definition} 
\theoremstyle{definition} 
\theoremstyle{definition} 

\allsectionsfont{\normalfont\sffamily\bfseries}

\newlist{trainsteps}{enumerate}{1}
\setlist[trainsteps]{label={(t\arabic*)}, ref={(t\arabic*)}, align=left}
\newlist{predsteps}{enumerate}{1}
\setlist[predsteps]{label={(p\arabic*)}, ref={(p\arabic*)}, align=left}

\newenvironment{customlegend}[1][]{\begingroup\csname pgfplots@init@cleared@structures\endcsname\pgfplotsset{#1}}{\csname pgfplots@createlegend\endcsname \endgroup}\def\addlegendimage{\csname pgfplots@addlegendimage\endcsname}
\newcommand{\addlegendimageintext}[1]{\protect\raisebox{1pt}{\protect\tikz[]{\protect\begin{customlegend}[anchor=base,legend entries={\protect\empty},legend style={draw=none,inner sep=0pt,column sep=0pt,nodes={inner sep=0pt}}]\protect\addlegendimage{#1}\protect\end{customlegend}}}}

\newcommand{\STAB}[1]{\begin{tabular}{@{}c@{}}#1\end{tabular}}

\begin{document}\startdocument

\titlehere

\abstracthere

\section{Introduction} \label{sect:introduction}

In many classification tasks, it is not sufficient to merely predict the class label $\widehat{y}(x)$ for a given feature space point $x$. Instead, it is often important to also have good predictions $\widehat{p}(y|x)$ for the class label probabilities, because these probability predictions provide a measure for the confidence one can have in the individual class label predictions $\widehat{y}(x)$. Such additional confidence information is  important in many applications, for instance in clinical applications~\citep{challis2015}. Classifiers that come with such additional class probability predictions are called calibrated. Some classifiers from methods like logistic regression or Gaussian process classification (GPC) are intrinsically calibrated. Also, there are various methods to calibrate an intrinsically non-calibrated classifier or to improve the calibration quality of an ill-calibrated classifier~\citep{niculescu2005,platt2000,zadrozny2002,gebel2009}.

\subsection{Contribution} \label{sect:introduction:contribution}
In this paper, we propose a novel supervised learning method for multi-class classification that yields classifiers with a high potential to be intrinsically well-calibrated. It can be applied to general classification problems in an arbitrary metrizable feature space $\mathcal{X}$ of possibly non-numeric features and with an arbitrary number $n$ of classes with labels $y \in \mathcal{Y} := \{l_1,\dots, l_n\}$. Starting from a training data set
\begin{align} \label{eq:training-data-set}
	\mathcal{D} := \{(x_i,y_i): i \in \{1,\dots,D\} \} \subset \mathcal{X} \times \mathcal{Y}
\end{align}
of feature space points $x_1, \dots, x_D \in \mathcal{X}$ together with associated class labels $y_i = y(x_i) \in \mathcal{Y}$, the training of our classifier proceeds in two training steps:
\begin{enumerate}
	\item[1.] In a first step, the training datapoints $x_1, \dots, x_D$ are transformed by means of a suitable training data transformation $f: \{x_1, \dots, x_D\} \to \mathcal{Z}$ to a latent space $\mathcal{Z}$, which we partition into $n$ cone segments $\mathcal{C}_{l_1}, \dots, \mathcal{C}_{l_n}$ corresponding to the $n$ classes in $\mathcal{D}$ and defined in terms of a regular $(n-1)$-dimensional simplex in $\mathcal{Z}$. 
	\item[2.] In a second step, a regression model $\widehat{f}: \mathcal{X} \to \mathcal{Z}$ is trained based on the transformed training datapoints $(x_1,f(x_1)), \dots, (x_D,(f(x_D))$, which are obtained by means of the training data transformation $f$ from the first step. In this manner, the latent space representation of the training data is extended to the whole feature space.
\end{enumerate}
We design the training data transformation $f$ such that the latent space counterpart $f(x_i)$ of each datapoint $x_i$ is located in the corresponding cone segment $\mathcal{C}_{y_i} = \mathcal{C}_{y(x_i)}$ and such that the location of $f(x_i)$ in this segment reflects the distances of $x_i$ from its own-class and its foreign-class datapoint neighbors. Concerning the choice of the distance metric and the number of neighbors used in the definition of $f$, we are completely free, and the same is true for the choice of the regression model $\widehat{f}$. In particular, these quantities can be freely customized and tuned to the particular problem at hand. 
\par

As soon as the above training steps have been performed, our classifier $\widehat{y}: \mathcal{X} \to \mathcal{Y}$ is readily obtained. Indeed, its class label prediction $\widehat{y}(x)$ for a given feature space point $x \in \mathcal{X}$ is the label of the (first) cone segment $\mathcal{C}_y$ that contains $\widehat{f}(x)$, the regression model's point prediction for $x$. If in addition to these point predictions, the regression model also provides probabilistic predictions, then our classifier yields predictions $\widehat{p}(y|x)$ for the class label probabilities as well. Specifically, these class label probability predictions read
\begin{align}
	\widehat{p}(y|x) := \int_{\mathcal{C}_y} \widehat{q}(z|x) \d z 
	\qquad (y\in\mathcal{Y}),
\end{align}
where $\widehat{q}(\,\cdot\,|x)$ for a given $x \in \mathcal{X}$ is the regression model's prediction for the probabilistic distribution of latent space points. 
As a classifier coming with class label probability predictions, our classifier is calibrated. We refer to it as a \emph{calibrated simplex-mapping classifier} (CASIMAC) because the underlying latent-space mapping $f$ is defined in terms of the vertices of a 
simplex in $\mathcal{Z}$.  
\par
We point out that the concept of leveraging Bayesian probabilistic prediction power combined with latent space mappings has been studied before. Several recent publications propose to couple a deep neural network with Gaussian processes (GPs) for an improved uncertainty estimate of model predictions~\citep{calandra2016,wilson2016,alshedivat2017,bradshaw2017,daskalakis2020}. Alternative approaches explore the use of deep neural networks not as feature extraction methods but, for instance, to suitably estimate the mean functions of GPs~\citep{iwata2017} or to predict their covariance functions and hyperparameters~\citep{cremanns2017}. Yet, due to the high complexity of the deep neural network components in these models, the algorithms mentioned above are well-suited for large data sets with abundant training data available~\citep{liu2018}. In the present paper, by contrast, we propose a simple latent space representation of the original feature space as the core component of a well-calibrated classifier that also works on less complex data sets. In particular, our method has recently been successfully used for an industrial application~\citep{lifecycle2021}.
\par
In summary, our contribution consists of the following parts:
\begin{enumerate}
	\item We propose a novel supervised learning method for multi-class classification with a simplex-like latent space.
	\item We rigorously establish the theoretical background including detailed proofs.
	\item We find that the computational effort of making predictions with our proposed classifier is comparatively 
	low (in contrast to, \eg, GPC).
	\item We show how the latent space of our proposed classifier can be suitably visualized.
	\item We benchmark the prediction and calibration properties of our proposed classifier.
\end{enumerate}
Additionally, we discuss potential use cases and further research directions.

\subsection{Simple example}

\begin{figure}
	\begin{center}
		\includegraphics{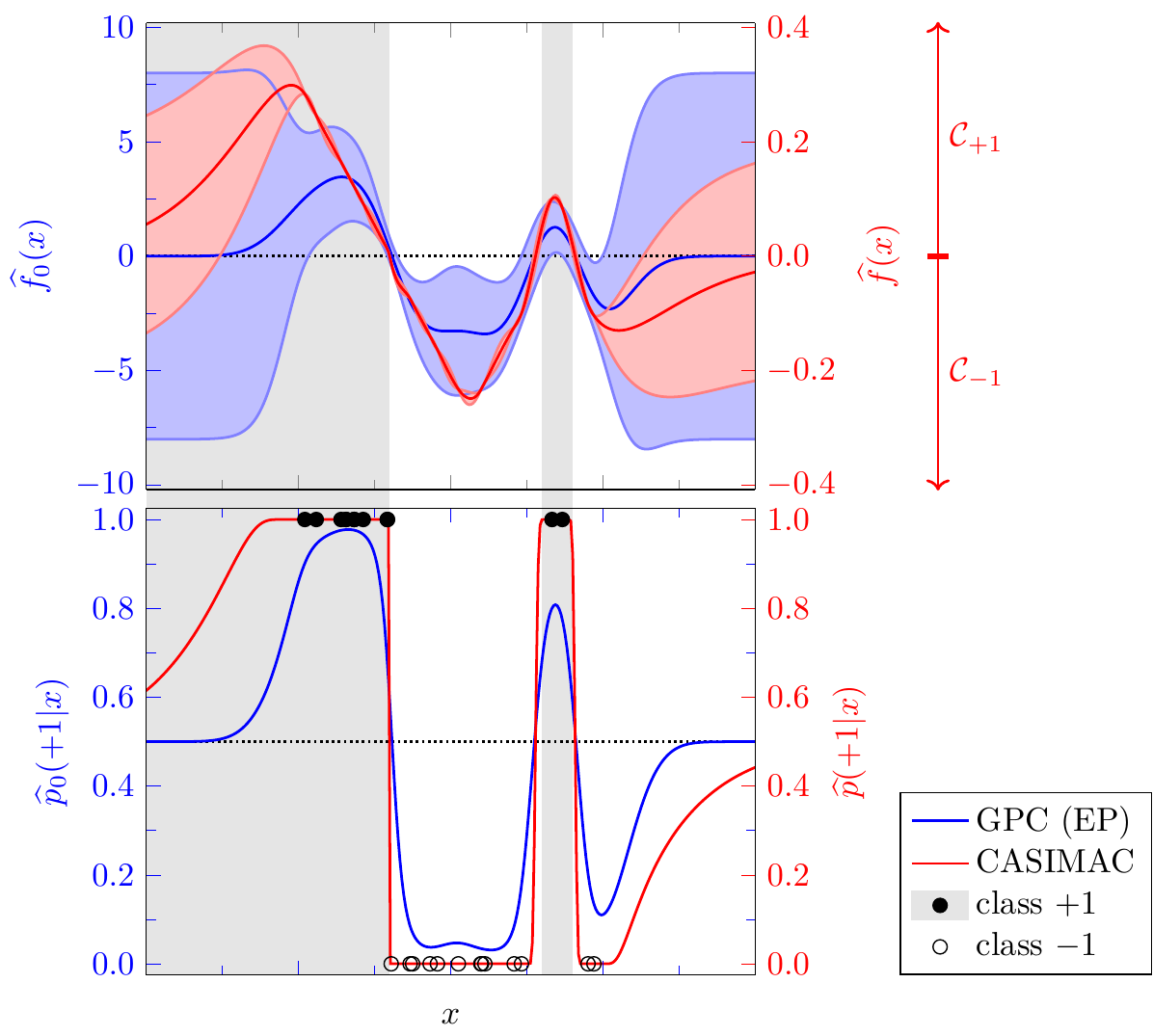}
		\captiontitle{An exemplary binary classification problem with one-dimensional feature space $\mathcal{X} = \R$ and class labels $y\in\{-1, +1\}$. We compare a GPC (using a radial-basis-function kernel and expectation propagation (EP) with the implementation from~\citep{gpy2014}) and our CASIMAC (based on a GPR with Mat\'{e}rn kernel as the underlying regression model), both trained on the same $D=25$ training datapoints. The top plot shows the expectation values $\widehat{f}_0(x)$, $\widehat{f}(x)$ and the standard deviations of the latent space distributions $\widehat{q}_0(\,\cdot\,|x)$, $\widehat{q}(\,\cdot\,|x)$ for both approaches (on two different scales), whereas the bottom plot contains the class probability predictions $\widehat{p}_0(+1|x)$ and $\widehat{p}(+1|x)$ for both approaches. The gray and white areas indicate the true regions of the class $+1$ and $-1$, respectively. All datapoints are sampled without noise from these regions. Finally, the dotted horizontal lines (\addlegendimageintext{densely dotted}) represent the decision boundaries. The corresponding cone segments for CASIMAC, $\mathcal{C}_{-1} = (-\infty,0]$ and $\mathcal{C}_{+1} = [0,\infty)$, are shown in the top plot.}{Exemplary binary classification problem.} \label{fig:gp-comparison}
	\end{center}
\end{figure}

In order to concretize 
the aforementioned assets of our method and paint a more intuitive picture, 
we briefly discuss a simple case with $n=2$ classes (\ie, a binary classification problem), which is shown in \cref{fig:gp-comparison}. The two class labels are chosen as $l_1 := -1$ and $l_2 := +1$, respectively. In that case, our latent space $\mathcal{Z} = \R$ is just the real line and the cone segments simply are $\mathcal{C}_{-1} = (-\infty,0]$ and $\mathcal{C}_{+1} = [0,\infty)$, the negative and the positive half-axis, respectively. Consequently, our definition of the class label predictions 
simplifies to
\begin{align}
	\widehat{y}(x) = -1 \quad \text{if } \widehat{f}(x) \le 0
	\qquad \text{and} \qquad 
	\widehat{y}(x) = +1 \quad \text{if } \widehat{f}(x) > 0,
\end{align}
while our definition of the class probability predictions simplifies to
\begin{align}
	\widehat{p}(-1|x) = \int_{-\infty}^0 \widehat{q}(z|x) \d z = 1-\widehat{p}(+1|x)
	\qquad \text{and} \qquad
	\widehat{p}(+1|x) = \int_0^{\infty} \widehat{q}(z|x) \d z.
\end{align}
As before, $\widehat{f}$ is a regression model that is fitted to the transformed training datapoints $(x_1,f(x_1)), \allowbreak \dots, \allowbreak (x_D,f(x_D))$ and $\widehat{q}(\,\cdot\,|x)$ is the associated  predictive a posteriori distribution for $x$ based on the same transformed training datapoints. 
\par

A simple choice for the regression model $\widehat{f}$ is a Gaussian process regressor (GPR). And a possible choice for the training data transformation $f$ on which $\widehat{f}$ depends is the simple distance-based map defined as follows:
\begin{align} \label{eq:training-data-trf-binary-special-case}
	f(x_i) := \text{signed distance of } x_i \text{ from its closest opposite-class datapoint neighbor},
\end{align}
where the sign in this formula is simply given by the class label $y_i = y(x_i) \in \mathcal{Y} = \{\pm 1\}$ of the datapoint $x_i$ and the distance is to be understood \wrt~the chosen metric on the feature space $\mathcal{X}$. If we train a CASIMAC with these choices for $f$ and $\widehat{f}$ on an exemplary training data set with $D = 25$ points in the one-dimensional feature space $\mathcal{X} = \R$, we obtain the class probability predictions $\widehat{p}(+1|x)$ depicted in red in \cref{fig:gp-comparison} (bottom plot). If we train a GPC on the same training data, we obtain the respective class probability predictions $\widehat{p}_0(+1|x)$ depicted in blue. As usual, 
\begin{align}
	\widehat{p}_0(+1|x) = \int_{-\infty}^{\infty} \sigma(z) \cdot\widehat{q}_0(z|x) \d z,
\end{align}
where $\sigma(z) := 1/(1+\e^{-z})$ represents the logistic sigmoid function and $\widehat{q}_0(\,\cdot\,|x)$ is the predictive a posteriori distribution of the GPC for the test point $x$~\citep{rasmussen2006,murphy2012}.
\par

We immediately see from the class probability plots in \cref{fig:gp-comparison} that our CASIMAC has a high confidence in its class label predictions in regions of densely sampled datapoints, as one would intuitively expect. In contrast, the GPC has considerably less confidence in its class label predictions at or near the datapoints despite its high prediction accuracy. In essence, this is because the training data transformation $f$ that underlies our classifier takes into account the actual distances of the datapoints and because the latent space probability density $\widehat{q}(\,\cdot\,|x)$ is considerably more concentrated around its expectation value $\widehat{f}(x)$ than is the case for its analog $\widehat{q}_0(\,\cdot\,|x)$. Another downside of the GPC -- further impairing its calibration quality -- is that its class probability predictions $\widehat{p}_0(+1|x)$ can be computed only approximately. In fact, already the computation of the non-normal distribution $\widehat{q}_0(\,\cdot\,|x)$, a $D$-dimensional integral, 
requires quite sophisticated approximations like the Laplace approximation~\citep{rasmussen2006}, expectation propagation~\citep{rasmussen2006}, variational inference~\citep{tran2015}, or the Markov chain Monte Carlo approximation~\citep{hensman2015}, to name a few. As opposed to this, no approximations are required for the computation of the CASIMAC class probabilities $\widehat{p}(+1|x)$ because there is a closed-form expression for them, due to the normality of the distributions $\widehat{q}(\,\cdot\,|x)$ assumed in our example from \cref{fig:gp-comparison}. 

\subsection{Structure of the paper}

In \cref{sect:method}, we formally introduce our CASIMAC method. We explain in detail the two training steps as well as how predictions of a trained classifier can be obtained. We also show how our latent space mappings can be leveraged for the convenient visualization of inter- or intra-class relationships in the data, especially in the case of a high-dimensional feature space and a moderate number of classes. In \cref{sect:validation}, we apply our method to various data sets and compare its performance and calibration qualities to several well-established benchmark classifiers. In particular, we demonstrate that our approach can be applied to many different application domains because our training data transformation reflects actual distances in the data set and because the respective distance metric as well as the regression model are freely customizable. \Cref{sect:conclusions} concludes the paper with a summary and an outlook on possible future research.
The appendix collects the mathematical and technical background underlying our CASIMAC method. In \cref{sect:math-background}, we rigorously prove the results needed for an in-depth and mathematically sound understanding of the method. In \cref{sect:implementation}, in turn, we summarize the main features of our implementation, and \cref{sect:hyperparameters} summarizes how the hyperparameters of the models in \cref{sect:validation} were chosen.

\section{Calibrated simplex-mapping classification} \label{sect:method}

In this section, we formally introduce our proposed calibrated simplex-mapping classifier, briefly referred to as CASIMAC. \Cref{sect:training-data-transformation,sect:regression-model} explain in detail the two training steps outlined in \cref{sect:introduction:contribution}. In \cref{sect:casimac}, we give the precise definition of our CASIMAC and, moreover, we explain how its 
predictions $\widehat{y}(x)$ and $\widehat{p}(y|x)$ for the class labels and the class label probabilities can be calculated in a computationally favorable manner. In \cref{sect:visualization} we finally explain how the latent space representation upon which our classification method relies can also be exploited 
for visualization purposes. 
Here and in the following, we consistently denote the training data set and the training datapoints as in~\eqref{eq:training-data-set} and write 
\begin{align}
	y(x_i) := y_i
\end{align}
for the true class labels of the training datapoints $x_1, \dots, x_D$. Also, $\mathcal{X}$ always denotes the feature space -- which is only assumed to be metrizable and, in particular, need not be embedded in any $\R^m$ --  and $\mathcal{Y}$ denotes the set of class labels. As usual, a set is called metrizable iff there exists at least one metric on it~\citep{dugundji1966}. 
We see below that we actually only need so-called semimetrics~\citep{wilson1931} and we exploit this even greater flexibility in our last application example. If $\mathcal{X}$ is embedded in some $\R^m$, then of course infinitely many metrics exist on $\mathcal{X}$, namely at least all the metrics induced by the $\ell^p$-norms $\norm{\cdot}_p$ on $\R^m$ for $p \in [1,\infty) \cup \{\infty\}$. In order to reduce cumbersome double indices, we assume from here on that the class labels are just $1, \dots, n$ (instead of the general labels $l_1, \dots, l_n$). In short, we assume that
\begin{align}
	\mathcal{Y} := \{1,\dots,n\}
\end{align}
without loss of generality.

\subsection{Training data transformation to a latent space}
\label{sect:training-data-transformation}

\begin{figure}
	\begin{center}
		\includegraphics{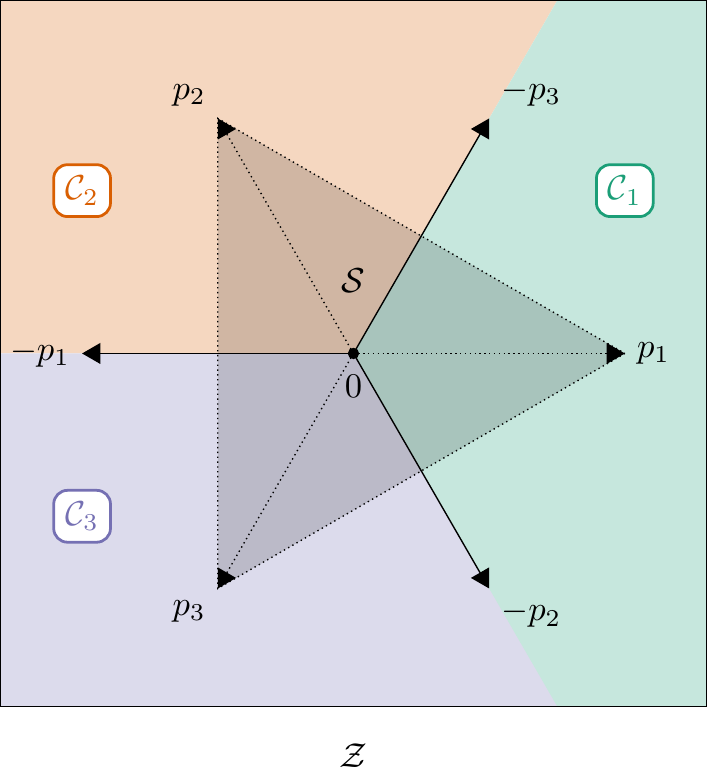}
		\captiontitle{Segmentation of the latent space $\mathcal{Z} = \R^2$ of a ternary classification problem ($n=3$) into three congruent cone segments $\mathcal{C}_1, \mathcal{C}_2, \mathcal{C}_3$, each associated with one of the three classes as indicated by the respective colors. The vertices $p_1,p_2,p_3$ of the simplex $\mathcal{S}$ (marked by the gray shading) with barycenter $0$ lie on the central ray of the respective segments. The borders of the segments are defined by the vertices $-p_1, -p_2, -p_3$ of the mirrored simplex $-\mathcal{S}$.}{Segmentation of the latent space.} \label{fig:simplex-segmentation}
	\end{center}
\end{figure}

In the first training step of our method, we transform the feature space training datapoints $x_1, \dots, x_D$ by means of a suitably designed training data transformation
\begin{align} \label{eq:training-data-trf}
	f: \mathcal{D}_{\mathcal{X}} \to \mathcal{Z} 
	\qquad (\mathcal{D}_{\mathcal{X}} := \{x_1, \dots, x_D\})
\end{align}
to a suitable latent space $\mathcal{Z}$. 
%
We choose this latent space to be $\mathcal{Z} := \R^{n-1}$, where  $n \ge 2$ as before is the number of classes observed in the training data $\mathcal{D}$. We decompose this space into $n$ conically shaped segments $\mathcal{C}_1, \dots, \mathcal{C}_n$, which are defined by the vertices $p_1, \dots, p_n$ of 
a regular $(n-1)$-dimensional simplex $\mathcal{S}$ in $\mathcal{Z}$ having barycenter $0$ and being at unit distance from their barycenter $0$, that is,
\begin{align} \label{eq:barycenter-0-and-unit-length-vertices}
	\sum_{i \in \mathcal{Y}} p_i = 0 \qquad \text{and} \qquad \norm{p_i}_2 = 1 \qquad (i \in \mathcal{Y}).
\end{align}
See \cref{sect:math-background} (Proposition~\ref{prop:sample-reg-simplex-in-Z}). Specifically, we define the segment $\mathcal{C}_k$ as the cone that is spanned by the mirrored vertices $-p_i$ with $i \ne k$, that is,
\begin{align} \label{eq:C_k-definition}
	\mathcal{C}_k := \bigg\{ z \in \mathcal{Z}: z = \sum_{i \in \mathcal{Y} \setminus \{k\}} c_i \cdot (-p_i) \text{ for some } c_i \in [0,\infty) \bigg\}.
\end{align}
In view of~(\ref{eq:barycenter-0-and-unit-length-vertices}.a), the vertex $p_k$ lies on the central ray of $\mathcal{C}_k$, 
which is why we also refer to $p_k$ as the central vector of $\mathcal{C}_k$. Also, 
the segments $\mathcal{C}_1, \dots, \mathcal{C}_n$ cover the whole latent space $\mathcal{Z}$ and any two of these segments overlap only at their boundaries. In short,
\begin{align} \label{eq:segmentation-of-Z}
	\bigcup_{k \in \mathcal{Y}} \mathcal{C}_k = \mathcal{Z}
	\qquad \text{and} \qquad 
	\mathcal{C}_k^{\circ} \cap \mathcal{C}_l = \emptyset \qquad (k \ne l),
\end{align}
where $\mathcal{C}_k^{\circ}$ is the interior of $\mathcal{C}_k$, that is, the set $\mathcal{C}_k$ without its boundary~\citep{dugundji1966}. It is given by
\begin{align} \label{eq:C_k-interior-representation}
\mathcal{C}_k^{\circ} = \bigg\{ z \in \mathcal{Z}: z = \sum_{i \in \mathcal{Y} \setminus \{k\}} c_i \cdot (-p_i) \text{ for some } c_i \in (0,\infty) \bigg\}.
\end{align}
And finally, the segments $\mathcal{C}_1, \dots, \mathcal{C}_n$ are pairwise congruent. All these statements are proven rigorously in \cref{sect:math-background} (Lemma~\ref{lm:closure-and-interior-of-C_k} and Propositions~\ref{prop:segmentation-of-Z} and~\ref{prop:C_k-and-C_l-congruent}). In the special case of just two or three classes, they can also be easily verified graphically. See \cref{fig:simplex-segmentation}, for instance. 
\par

With the help of the above segmentation of the latent space $\mathcal{Z}$, we can define the training data transformation~\eqref{eq:training-data-trf}. We design this mapping $f$ such that it maps each training datapoint $x \in \mathcal{D}_{\mathcal{X}}$ to the cone segment $\mathcal{C}_{y(x)}$ corresponding to its class label $y(x)$ in such a way that the location of $f(x)$ in this cone segment $\mathcal{C}_{y(x)}$ reflects the distances of $x$ from its own-class and from its foreign-class neighbors. We choose the location of $f(x)$ based on the following premises:
\begin{itemize}
\item[1.] $f(x)$ should be located the farther in the direction of $p_{y(x)}$ 
(and thus the farther inside the cone segment $\mathcal{C}_{y(x)}$), the closer $x$ is to its class-$y(x)$ datapoint neighbors
\item[2.] $f(x)$ should be located the farther in the direction of $-p_y$ 
(and thus the farther away from the cone segment $\mathcal{C}_{y}$), the farther $x$ is away from its class-$y$ datapoint neighbors for $y \in \mathcal{Y} \setminus \{y(x)\}$. 
\end{itemize}
Specifically, we define $f$ as follows:
\begin{align} \label{eq:training-data-trf-defining-formula}
	f(x) = f_{\alpha,\beta,k_{\alpha},k_{\beta},d}(x) := \alpha A_{k_{\alpha},d}(x) \cdot p_{y(x)} + \sum_{y\in \mathcal{Y}\setminus\{y(x)\}} \beta R_{k_{\beta},d}(x,y) \cdot (-p_y)
\end{align}
for every training datapoint $x \in \mathcal{D}_{\mathcal{X}}$. In particular, the coefficient $A_{k_{\alpha},d}(x)$ indicates how far $f(x)$ is pulled into the own-class segment $\mathcal{C}_{y(x)}$, while the coefficient  $R_{k_{\beta},d}(x,y)$ indicates how far $f(x)$ is pulled away from the foreign-class segment $\mathcal{C}_y$ for $y \in \mathcal{Y} \setminus \{y(x)\}$. We therefore refer to these coefficients as the attraction and the repulsion coefficients of $x$ and we define them, as indicated above, in terms of the mean distance of $x$ from its $k_{\alpha}$ closest datapoint neighbors of its own class $y(x)$ or, respectively, from its $k_{\beta}$ closest datapoint neighbors of the foreign class $y \in \mathcal{Y} \setminus \{y(x)\}$. That is, 
\begin{gather} \label{eq:attraction-and-repulsion-coefficients-definition}
	A_{k_{\alpha},d}(x) := \big( \mathrm{NN}_{k_{\alpha},d}(x,\mathcal{D}_{\mathcal{X},y(x)}) \big)^{-1} 
	\qquad \text{and} \qquad
	R_{k_{\beta},d}(x,y) := \mathrm{NN}_{k_{\beta},d}(x,\mathcal{D}_{\mathcal{X},y}),
\end{gather}
where $\mathcal{D}_{\mathcal{X},y} := \{x_i \in \mathcal{D}_{\mathcal{X}}: y_i = y(x_i) = y\}$ is the set of all training datapoints belonging to class $y$ and
\begin{align} \label{eq:nearest-neighbors-definition}
	\mathrm{NN}_{k,d}(x,X) := \min_{X' \subset X\setminus\{x\} \text{ with } |X'| = k} \frac{1}{k} \sum_{x'\in X'} d(x,x')
\end{align}
is the mean distance of $x$ from its $k$ nearest neighbors from the subset $X \subset \mathcal{X}$, the distance being measured in terms of some arbitrary semimetric~\citep{wilson1931}
\begin{align} \label{eq:metric-d}
	d: \mathcal{X} \times \mathcal{X} \to [0,\infty)
\end{align} 
on the feature space $\mathcal{X}$, that is, $d(x,x') = d(x',x)$ for all $x,x' \in \mathcal{X}$ (symmetry) and $d(x,x') = 0$ if and only if $x=x'$. At least one such semimetric exists on $\mathcal{X}$ because this space was assumed to be semimetrizable at the beginning of \cref{sect:method}. Also, $\alpha, \beta \in [0,\infty)$ and $k_{\alpha}, k_{\beta} \in \N$ are user-defined hyperparameters satisfying
\begin{align} \label{eq:conditions-on-hyperparameters}
	\alpha + \beta > 0 
	\qquad \text{and} \qquad
	k_{\alpha} \le c-1
	\qquad \text{and} \qquad
	k_{\beta} \le c,
\end{align}
where $c := \min\{ |\mathcal{D}_{\mathcal{X},y}|: y \in \mathcal{Y}\}$ is the cardinality of the smallest training data class. Conditions~(\ref{eq:conditions-on-hyperparameters}.b) and~(\ref{eq:conditions-on-hyperparameters}.c) guarantee that all the attraction and repulsion coefficients are strictly positive finite real numbers, that is,
\begin{align} \label{eq:attraction-and-repulsion-coefficients-positive-and-finite}
	0 < A_{k_{\alpha},d}(x), R_{k_{\beta},d}(x,y) < \infty 
\end{align}
for all $x \in \mathcal{D}_{\mathcal{X}}$ and all $y \in \mathcal{Y} \setminus \{y(x)\}$ (Proposition~\ref{prop:generic-properties-of-training-data-trf}). In most of our applications, $d$ is a proper metric, that is, a semimetric that also satisfies the triangle inequality. In our last application example, though, we make explicit use of semimetrics as well.   
\par

\begin{figure}
	\begin{center}
		\includegraphics{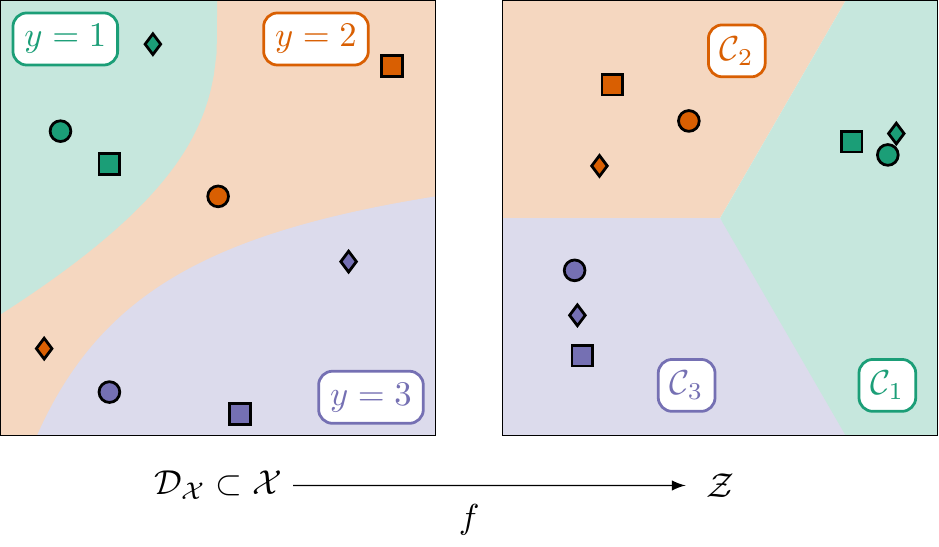}
		\captiontitle{Illustration of the training data transformation $f: \{x_1, \dots, x_D\} \to \mathcal{Z} = \R^2$ from \cref{eq:training-data-trf-defining-formula} for an exemplary ternary classification problem ($n=3$) with a two-dimensional feature space $\mathcal{X} \subset \R^2$. The colored regions on the left illustrate the true regions of the three classes with their respective labels, whereas the points denote the (noiselessly) sampled training datapoints $x_1, \dots, x_D$. We sample three points from each class (\ie, $D=9$) and use different symbols (\addlegendimageintext{only marks,mark=*,mark options={fill=white}}, \addlegendimageintext{only marks,mark=square*,mark options={fill=white}} and \addlegendimageintext{only marks,mark=diamond*,mark options={fill=white}}) to uniquely identify each point. As the distance metric $d$ underlying $f$, we choose the Euclidean distance. The other hyperparameters are chosen as $\alpha := 0$, $\beta := 1$ and $k_{\alpha}, k_{\beta} := 1$, respectively. The figure shows that the farther a training datapoint $x_i$ (for $i \in \{1,\dots,D\}$) is away from its nearest foreign-class neighbor, the clearer is the membership of $f(x_i)$ to the respective cone segment $\mathcal{C}_{y(x_i)}$.}{Illustration of the training data transformation.} \label{fig:training-data-trf}
	\end{center}
\end{figure}

Since the domain $\mathcal{D}_{\mathcal{X}}$ of $f$ and the attraction and repulsion coefficients occurring in the definition of $f$ obviously depend on the training data set $\mathcal{D}$, so does the training data transformation $f$ itself, and we sometimes make this dependence explicit by writing
\begin{align}
	f(x) = f(x|\mathcal{D}) = f_{\alpha,\beta,k_{\alpha},k_{\beta},d}(x|\mathcal{D}). 
\end{align}
It is straightforward to verify that the training data transformation $f$ from~\eqref{eq:training-data-trf-defining-formula}, with $\alpha := 0$, $\beta := 1$, $k_{\alpha}, k_{\beta} := 1$, reduces to the mapping from~\eqref{eq:training-data-trf-binary-special-case} in the special case of just $n = 2$ classes. 
It is also easy to verify from~\eqref{eq:training-data-trf-defining-formula}, using the barycenter condition~(\ref{eq:barycenter-0-and-unit-length-vertices}.a) along with~\eqref{eq:C_k-interior-representation}, (\ref{eq:conditions-on-hyperparameters}.a) and~\eqref{eq:attraction-and-repulsion-coefficients-positive-and-finite}, that
\begin{align} \label{eq:training-datapoint-transformed-to-interior-of-its-segment}
	f(x) = \sum_{y\in \mathcal{Y}\setminus\{y(x)\}} \Big( \alpha A_{k_{\alpha},d}(x) + \beta R_{k_{\beta},d}(x,y) \Big) \cdot (-p_y)
	\in \mathcal{C}_{y(x)}^{\circ}
\end{align}
for every $x \in \mathcal{D}_{\mathcal{X}}$ (Proposition~\ref{prop:generic-properties-of-training-data-trf}). In other words, $f(x)$ for every training datapoint $x \in \mathcal{D}_{\mathcal{X}}$ lies in the interior $\mathcal{C}_{y(x)}^{\circ}$ of the corresponding cone segment. 
Specifically, the membership of $f(x)$ to $\mathcal{C}_{y(x)}^{\circ}$ is the clearer, the closer $x$ is to its $k_{\alpha}$ nearest own-class neighbors and the farther $x$ is away from its $k_{\beta}$ nearest foreign-class neighbors.
\Cref{fig:training-data-trf} illustrates this behavior of the training data transformation $f$ for the simple case of a ternary classification problem ($n=3$). 
\par

We point out that this behavior of $f$ is generic in the sense that it is independent of the number of classes and independent of the specific choices of $\alpha, \beta, k_{\alpha}, k_{\beta}$ and $d$. In particular, we can freely choose and tune the semimetric $d$ as well as the hyperparameters $\alpha, \beta, k_{\alpha}, k_{\beta}$ within the bounds~\eqref{eq:conditions-on-hyperparameters} to the specific classification problem at hand~\citep{deza2016}. We make ample use of this customization flexibility for the exemplary classification problems in \cref{sect:validation}.

\subsection{Training of a regression model based on the transformed data}
\label{sect:regression-model}

In the second training step of our method, we train a regression model
\begin{align} \label{eq:regression-model}
	\widehat{f}: \mathcal{X} \to \mathcal{Z}
\end{align}
from the feature space $\mathcal{X}$ to the latent space $\mathcal{Z}$. As the training data set for this regression model, we take the transformed training data set
\begin{align} \label{eq:transformed-training-data}
	\mathcal{D}^f := \{(x_i,f(x_i)): i \in \{1,\dots,D\}\} \subset \mathcal{X} \times \mathcal{Z}
\end{align}
consisting of the feature space training datapoints $x_1, \dots, x_D \in \mathcal{D}_{\mathcal{X}}$ together with their latent space counterparts $f(x_1), \dots, f(x_D) \in \mathcal{Z}$ which are obtained by means of the training data transformation $f$ from the first training step as defined in \cref{eq:training-data-trf-defining-formula}. 
\par

\begin{figure}
	\begin{center}
		\includegraphics{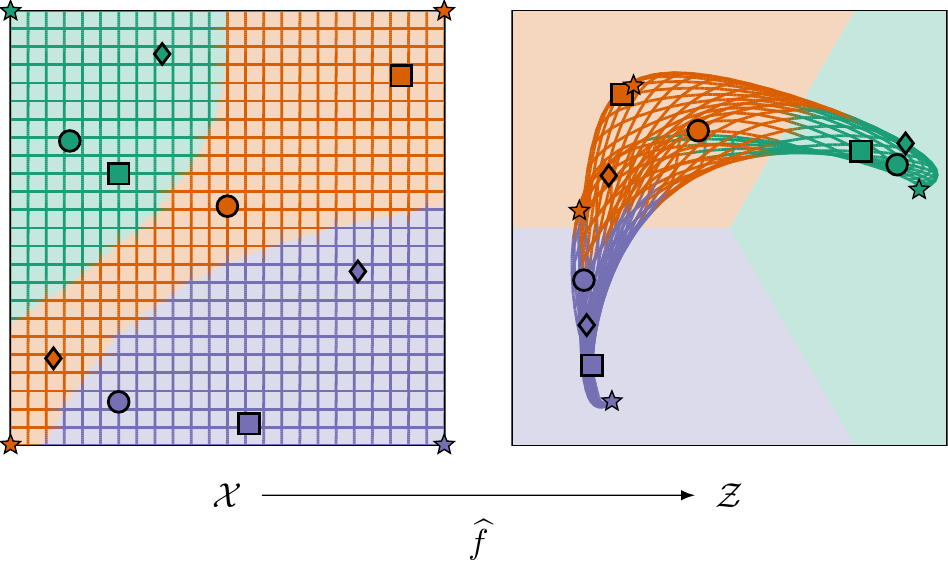}
		\captiontitle{Illustration of the regression model $\widehat{f}: \mathcal{X} \to \mathcal{Z} = \R^2$ for the exemplary ternary classification problem from \cref{fig:training-data-trf}. We choose a GPR model with a Mat\'{e}rn kernel, which is fitted to the transformed training datapoints $(x_1,f(x_1), \allowbreak, \dots, \allowbreak (x_D,f(x_D))$ that are obtained from the training data transformation $f$ according to \cref{fig:training-data-trf}. In order to illustrate the effect of $\widehat{f}$, we show how the points of the rectangular grid on the left get transformed by $\widehat{f}$. In particular, we show this transformation for the corner points marked by stars. The color of any point $x$ on the grid and, respectively, of any point $\widehat{f}(x)$ on the transformed grid indicates the true class of $x$.}{Illustration of the regression model.} \label{fig:regression-model}
	\end{center}
\end{figure}

In sharp contrast to $f$, the regression model $\widehat{f}$ is defined on the whole of $\mathcal{X}$ (instead of only the training datapoints $x_1, \dots, x_D$) and thus yields latent space predictions for arbitrary feature space points $x \in \mathcal{X}$, especially for hitherto unobserved and unclassified  feature space points. \Cref{fig:regression-model} illustrates the effect of the regression model for the special case of a ternary classification problem ($n=3$). 
\par

Since both $\mathcal{D}_{\mathcal{X}}$ and $f = f(\,\cdot\,|\mathcal{D})$ depend on the original training data set $\mathcal{D}$, so does the regression model $\widehat{f}$ and we sometimes make this dependence explicit by writing 
\begin{align}
	\widehat{f}(x) = \widehat{f}(x|\mathcal{D}).
\end{align}
We point out that -- similarly to the choice of the hyperparameters $\alpha, \beta, k_{\alpha}, k_{\beta}$ and $d$ of the training data transformation -- the specific choice of the regression model is completely arbitrary. 
\par

In particular, we can choose a suitable probabilistic regression model that is able to not only predict a single latent space point $\widehat{f}(x) = \widehat{f}(x|\mathcal{D})$ but also an entire probability density $\widehat{q}(\,\cdot\,|x) = \widehat{q}(\,\cdot\,|\mathcal{D}, x)$ for every $x \in \mathcal{X}$. Such a so-called predictive a posteriori distribution indicates the distribution of the latent space points for given $x$ and a given data set $\mathcal{D}$. 
In particular, the predictive a posteriori distribution $\widehat{q}(\,\cdot\,|x)$ indicates the level of confidence that the model has in its point prediction $\widehat{f}(x)$, namely it is all the more confident in the point prediction $\widehat{f}(x)$ the more $\widehat{q}(\,\cdot\,|x)$ is concentrated around $\widehat{f}(x)$. A typical example for regression models which provide both a point prediction $\widehat{f}(x)$ and a probability-density prediction $\widehat{q}(\,\cdot\,|x)$ for every $x \in \mathcal{X}$ are GPR models~\citep{rasmussen2006,murphy2012}. 
As is well-known, for these models the point predictions $\widehat{f}(x)$ are completely determined by the predictive a posteriori distribution $\widehat{q}(\,\cdot\,|x)$, namely as the maximum a posteriori prediction, that is,
\begin{align} \label{eq:MAP-prediction}
\widehat{f}(x) = \argmax_{z \in \mathcal{Z}} \widehat{q}(z|x).
\end{align}
We make ample use of GPR models in our benchmark examples from \cref{sect:validation}.

\subsection{Calibrated simplex-mapping classifier}
\label{sect:casimac}

After the two training steps explained above have been performed, we can define and put to use our CASIMAC. We define it as the classifier
\begin{align}
	\widehat{y}: \mathcal{X} \to \mathcal{Y}
\end{align}
that assigns to each feature space point $x \in \mathcal{X}$ the label $y$ of the first cone segment $\mathcal{C}_y$ that contains $\widehat{f}(x)$. That is,
\begin{align} \label{eq:casimac-definition}
	\widehat{y}(x) := \widehat{g}(\widehat{f}(x)) := \min \{y \in \mathcal{Y}: \widehat{f}(x) \in \mathcal{C}_y \},
\end{align}
where $\widehat{f}(x)$ is the regression model's latent space prediction for $x$ and where
\begin{align} \label{eq:class label-reconstruction-definition}
	\widehat{g}(z) := \min \{y \in \mathcal{Y}: z \in \mathcal{C}_y \}
\end{align}
is the label of the first cone segment containing $z$. Since by~(\ref{eq:segmentation-of-Z}.a) every latent space point $z \in \mathcal{Z}$ is contained in at least one of the segments $\mathcal{C}_1, \dots, \mathcal{C}_n$, the expression~\eqref{eq:casimac-definition} really yields a well-defined 
classifier. 
Since, on the other hand, the segments overlap at their respective borders, a given latent space point $z \in \mathcal{Z}$ is contained in several segments if it is located exactly on such a border. In this special case, we need to decide for one of the overlapping segments, to obtain a classifier that assigns a single label (instead of multiple labels) to each feature space point.
In our definition~\eqref{eq:casimac-definition}, we decide for the first of the segments containing $z = \widehat{f}(x)$ for the sake of simplicity, hence the minimum in that formula.
\par

\begin{figure}
	\begin{center}
		\includegraphics{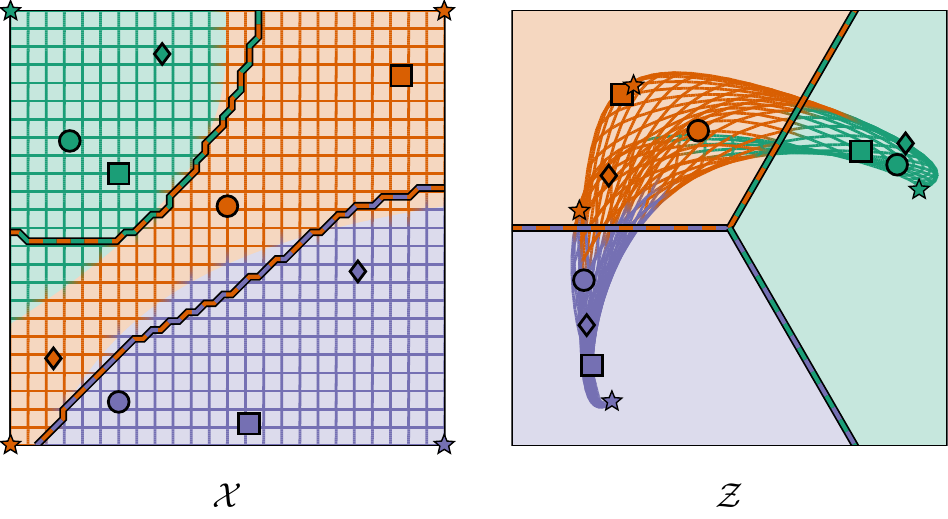}
		\captiontitle{Illustration of our proposed calibrated simplex-mapping classifier $\widehat{y}: \mathcal{X} \to \mathcal{Y}$ for the exemplary ternary classification problem from \cref{fig:training-data-trf} based on the GPR model $\widehat{f}$ from \cref{fig:regression-model}. According to our definition~\eqref{eq:casimac-definition}, the classifier determines for each feature space point $x \in \mathcal{X}$ (on the left) the cone segment containing the latent space counterpart $\widehat{f}(x) \in \mathcal{Z}$ (on the right) and then takes the label of this cone segment as the class label prediction $\widehat{y}(x) \in \mathcal{Y}$ for $x$. In other words, the cuts of the cone segment borders through the latent space $\mathcal{Z}$ determine the class membership of each data point $x$ according to its learned latent space position $\widehat{f}(x)$. We mark the predicted class boundaries (on the left) as well as the corresponding cone segment boundaries (on the right) by dashed lines. 
		As can be seen, the predicted class boundaries in $\mathcal{X}$ do not deviate much from the true class boundaries indicated by the different background and grid colors on the left. In other words, our classifier produces only a few misclassifications. This can also be seen from the fact that the color of most of the transformed grid points in $\mathcal{Z}$ on the right (indicating the true class) is identical to the underlying background color (indicating the predicted class).}{Illustration of our proposed calibrated simplex-mapping classifier.} \label{fig:casimac}
	\end{center}
\end{figure}

In view of the dependence of the regression model $\widehat{f} = \widehat{f}(\,\cdot\,|\mathcal{D})$ on the training data set $\mathcal{D}$, our CASIMAC depends on $\mathcal{D}$ as well and, to emphasize this, we sometimes write
\begin{align}
	\widehat{y}(x) = \widehat{y}(x|\mathcal{D}) = \widehat{g}(\widehat{f}(x|\mathcal{D})). 
\end{align}
According to the definition~\eqref{eq:casimac-definition}, in order to practically compute the class label prediction $\widehat{y}(x)$ for a given feature space point $x\in \mathcal{X}$, we have to determine all cone segments that contain $\widehat{f}(x)$. Yet, in view of the 
purely geometric definition~\eqref{eq:C_k-definition} of the cone segments, it is not so clear at first glance how this can be done 
in a computationally feasible and favorable manner -- especially in the case of many classes ($n \ge 5$) where the segments cannot be (directly) visualized anymore. 
We find, however, that the segments containing $\widehat{f}(x)$ can be determined solely in terms of the distances 
\begin{align} \label{eq:distances-to-central-vectors}
	\big\|\widehat{f}(x) - p_1\big\|_2, \dots, \big\|\widehat{f}(x) - p_n\big\|_2
\end{align}   
of $\widehat{f}(x)$ from the central vectors $p_1, \dots, p_n$ of the segments $\mathcal{C}_1, \dots, \mathcal{C}_n$. Specifically, we prove in \cref{sect:math-background} (Theorem~\ref{thm:norm-char-of-C_k}) that a latent space point $z \in \mathcal{Z}$ belongs to the segment $\mathcal{C}_y$ if and only if 
\begin{align}
	\norm{z-p_y}_2 = \min_{l\in\mathcal{Y}} \norm{z-p_l}_2,
\end{align}
that is, if and only if its distance from the other segments' central vectors $p_l$, $l \ne y$, is at least as large as its distance from the central vector $p_y$ of $\mathcal{C}_y$. Consequently, the geometrically inspired definition~\eqref{eq:casimac-definition} of our CASIMAC can be recast in the form
\begin{align} \label{eq:casimac-alternative-definition}
	\widehat{y}(x) = \min\Big\{ y \in \mathcal{Y}: \big\|\widehat{f}(x) - p_y\big\|_2 = \min_{l\in\mathcal{Y}} \big\|\widehat{f}(x) - p_l\big\|_2 \Big\}
\end{align}   
(Corollary~\ref{cor:casimac}). 
Computationally, \eqref{eq:casimac-alternative-definition} is much easier to evaluate than~\eqref{eq:casimac-definition} since $\widehat{y}(x)$ can be determined simply by calculating all distances~\eqref{eq:distances-to-central-vectors} and by then choosing the smallest one. 
In our CASIMAC implementation, we therefore use~\eqref{eq:casimac-alternative-definition} to perform predictions. 
We illustrate the method in \cref{fig:casimac}.
\par

If the regression model $\widehat{f}$ 
perfectly fits the training data~\eqref{eq:transformed-training-data} in the sense that 
\begin{align} \label{eq:perfect-fit}
	\widehat{f}(x_i) = f(x_i) \qquad (i \in \{1,\dots,D\}),
\end{align}
then it is easy to see from~\eqref{eq:casimac-definition}, using~(\ref{eq:segmentation-of-Z}.b) and~\eqref{eq:training-datapoint-transformed-to-interior-of-its-segment}, that
\begin{align}
	\widehat{y}(x_i) = y(x_i) = y_i \qquad (i \in \{1,\dots,D\}) 
\end{align}
(Corollary~\ref{cor:casimac}). 
In other words, if the regression model is perfectly interpolating in the sense of \eqref{eq:perfect-fit}, our CASIMAC perfectly reconstructs the true class labels of the training datapoints. A typical example of perfectly interpolating regression models are noise-free GPR models~\citep{rasmussen2006,murphy2012}.
\par

If the regression model apart from its point predictions also provides probabilistic predictions, we can estimate the class probabilities for arbitrary feature space points $x$. In that case, there is a latent probability density $\widehat{q}(\,\cdot\,|\mathcal{D},x)$ on $\mathcal{Z}$ for every $x \in \mathcal{X}$ which on the one hand determines the point prediction $\widehat{f}(x)$ in some way -- for instance, as the maximum a posteriori estimate~\eqref{eq:MAP-prediction} -- and which on the other hand also determines the uncertainty of this prediction. Since the regression model $\widehat{f}$ is our model for observations of latent-space points, the associated probabilities $\widehat{q}(z|\mathcal{D},x)$ 
can be considered as estimates for the probability 
of observing the latent-space point $z \in \mathcal{Z}$ for a given $x \in \mathcal{X}$. Since, moreover, $\widehat{y}(x) = y$ if and only if $\widehat{f}(x) \in \{z\in\mathcal{Z}: \widehat{g}(z) = y\}$ 
by the definition~\eqref{eq:casimac-definition} of our classifier, the expression
\begin{align} \label{eq:class-probability-estimate}
	\widehat{p}(y|\mathcal{D},x) := \int_{\{\widehat{g} = y\}} \widehat{q}(z|\mathcal{D},x) \d z
	= \int_{\mathcal{Z}} I_{\{\widehat{g} = y\}}(z) \cdot \widehat{q}(z|\mathcal{D},x) \d z
\end{align}
is an estimate 
for the probability of observing class $y$ for $x$. We therefore take this estimate as our classifier's prediction for the  probability of class $y$ given $x$. In the above equation, $I_{\{\widehat{g} = y\}}$ denotes the indicator function 
\begin{align} \label{eqn:indicator}
	I_{\{\widehat{g} = y\}}(z)
	= 
	\begin{cases}
		1, \quad \text{if } \widehat{g}(z) = y\\
		0, \quad \text{otherwise}
	\end{cases}
\end{align}
with respect to the set $\{\widehat{g} = y\} := \{z\in\mathcal{Z}: \widehat{g}(z) = y\}$.
In view of~(\ref{eq:segmentation-of-Z}.b) and~\eqref{eq:class label-reconstruction-definition}, we have $\mathcal{C}_y^{\circ} \subset \{\widehat{g} = y\} \subset \mathcal{C}_y$ for every $y \in \mathcal{Y}$ and thus
\begin{align} \label{eq:class-probability-estimate-alternative}
	\widehat{p}(y|\mathcal{D},x)
	= \int_{\mathcal{C}_y} \widehat{q}(z|\mathcal{D},x) \d z
	= \int_{\mathcal{C}_y^{\circ}} \widehat{q}(z|\mathcal{D},x) \d z
\end{align}
are alternative expressions for our classifier's class probability prediction (Corollary~\ref{cor:alternative-representation-of-class-probabilities}).\par

If the probability densities $\widehat{q}(\,\cdot\,|\mathcal{D},x)$ belong to a normal distribution and the number of classes is $n = 2$, then the class probability predictions~\eqref{eq:class-probability-estimate-alternative} can be computed by means of a simple analytical formula, namely~\eqref{eq:class-probability-closed-form} in \cref{sect:math-background} (Proposition~\ref{prop:computation-of-class-probabilities}). 
If, by contrast, the probability densities $\widehat{q}(\,\cdot\,|\mathcal{D},x)$ do not belong to a normal distribution or the number of classes is $n > 2$, then the multi-dimensional integrals in~\eqref{eq:class-probability-estimate-alternative} can still be computed approximately. In our CASIMAC implementation, we use Monte Carlo sampling for these approximate computations of the class probability predictions~\eqref{eq:class-probability-estimate-alternative} for $n >2$. See~\eqref{eq:class-probability-approximant} in \cref{sect:math-background} (Proposition~\ref{prop:computation-of-class-probabilities}). Standard examples of regression models $\widehat{f}$ with normally distributed probability densities $\widehat{q}(\,\cdot\,|\mathcal{D},x)$ are provided by GPR models. 
In particular, the class probabilities for the binary classification problem from \cref{fig:gp-comparison} were evaluated analytically. 

\subsection{Visualization}
\label{sect:visualization}

Apart from its central use in the definition of our classifier, the latent space representation given by $f$ and $\widehat{f}$ can also be beneficially used for visualization purposes. 
\par

In particular, the training data transformation $f$ can be exploited for visually detecting inter- or intra-class relationships in the data. Inspecting the latent space representation from \cref{fig:training-data-trf}, for instance, we find that the transformed datapoints of class $1$ and $3$ are farther apart from each other than they both are apart from the transformed datapoints of class $2$. 
We could have directly seen that, of course, by inspecting the untransformed  datapoints in feature space, which is only $2$-dimensional here. After all, class $2$ lies between class $1$ and $3$, as the left panel of \cref{fig:training-data-trf} reveals. Such a direct inspection of the data in feature space, however, is possible only for feature spaces with small dimensions. In case the feature space is high-dimensional or not even embedded in an $\R^m$, though, we can still use its latent space representation to uncover relations between the datapoints. All we need for that is a latent space of moderate dimension $n-1$, ideally $n \le 4$.
\par

Since our latent space $\mathcal{Z} = \R^{n-1}$ is an unbounded set by definition, the latent space counterparts $f(x_1), \dots, f(x_D)$ of the datapoints can be very far apart from each other. It can  therefore be useful to further transform the latent space -- and with it, the points $f(x_1), \dots, f(x_D)$ -- to a fixed bounded reference set. A natural way of achieving this is to compress the latent space $\mathcal{Z}$ to the interior of the reference simplex $\mathcal{S}$ underlying our CASIMAC. Indeed, there is a bijective compression map
\begin{align} \label{eq:compression-map}
	C: \mathcal{Z} \to \mathcal{S}^{\circ}
\end{align}
which is diffeomorphic (infinitely differentiable with infinitely differentiable inverse) and which leaves the cone segments $\mathcal{C}_1, \dots, \mathcal{C}_n$ invariant in the sense that, for every $z \in \mathcal{Z}$ and every $k \in \mathcal{Y}$, 
\begin{align} \label{eq:compression-respects-segmentation}
	C(z) \in \mathcal{C}_k 
	\qquad \text{if and only if} \qquad
	z \in \mathcal{C}_k.
\end{align}
See \cref{sect:math-background} (Proposition~\ref{prop:compression-of-Z}). We use this compression map for the visualization of one of our real-world data sets below.

\section{Benchmark} 
\label{sect:validation}

In this section, we apply our CASIMAC method to various data sets and compare its performance to several well-established benchmark classifiers. Specifically, we perform benchmarks using synthetic data in \cref{sect:synthetic-data} and real-world data in \cref{sect:real-world-data}. Furthermore, we exemplarily illustrate the latent space visualization in \Cref{sect:visualization-example}. In these first three subsections, the regression model underlying our classifier is a GPR. In \cref{sect:large-data}, by contrast, we use a neural network as our regression model, thereby demonstrating how a more complex regression model allows us to solve a classification task with more training data. 
For all calculations, we use the implementation of our method~\citep{casimac-github}, which is briefly outlined in \cref{sect:implementation}.

\subsection{Synthetic data}
\label{sect:synthetic-data}

\begin{figure}[!htb]
	\centering
	\includegraphics{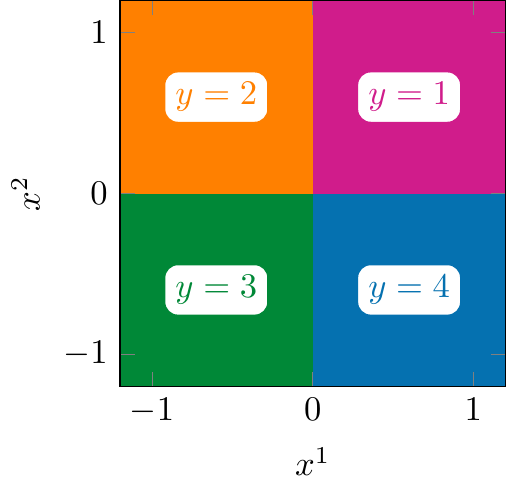}
	\captiontitle{Illustration of the ground truth \eqref{eq:artificial-ytrue} for the synthetic data class labels.}{Synthetic data visualization.} \label{fig:artificial}
\end{figure}

As a first benchmark, we test our method on a synthetically generated data set. Specifically, we consider the synthetic classification problem with $n=4$ classes and the $2$-dimensional feature space $\mathcal{X} := [-1,1]^2$, which is based on the explicit class membership rule
\begin{align} \label{eq:artificial-ytrue}
	y_{\mathrm{true}}(x) := \begin{cases} 1, \quad \text{if } x^1 \ge 0 \text{ and } x^2 \ge 0 \\ 2, \quad \text{if } x^1 < 0 \text{ and } x^2 \ge 0 \\ 3, \quad \text{if } x^1 < 0 \text{ and } x^2 < 0 \\ 4,  \quad \text{if } x^1 \ge 0 \text{ and } x^2 < 0 \end{cases}
\end{align}
for features $x := (x^1,x^2) \in \mathcal{X}$. The problem is illustrated in \cref{fig:artificial}. In order to obtain our training data set~\eqref{eq:training-data-set}, we uniformly sample $D = 40$ points from $\mathcal{X}$ and associate them with their true class labels according to~\eqref{eq:artificial-ytrue}. In particular, the synthetic data is free of noise. Analogously, we obtain the test data set
\begin{align} \label{eq:test-data-set}
	\mathcal{T} := \{(x_i,y_i): i \in \{D+1, \dots, D+T\}\}
\end{align}
by uniformly sampling $T = \num{10000}$ points from $\mathcal{X}$ and by associating them with their true class labels~\eqref{eq:artificial-ytrue}. Here and in the following, we always standardize the features based on the training data before feeding it to the considered classifiers.

As the distance metric $d$ in the training data transformation~\eqref{eq:training-data-trf-defining-formula} underlying our CASIMAC, we choose the Euclidean distance on $\mathcal{X} \subset \R^2$, that is,
\begin{align} \label{eq:l^2-distance}
	d(x,x') := \norm{x-x'}_2
\end{align}
for all $x, x' \in \mathcal{X}$. Also, we parameterize the weights $\alpha, \beta$ for the attraction and repulsion coefficients~\eqref{eq:attraction-and-repulsion-coefficients-definition} by a single parameter $\gamma \in [0,1]$, namely
\begin{align} \label{eq:gamma}
	\gamma := \beta := 1-\alpha,
\end{align}
so that $\alpha, \beta \in [0,1]$. And as the regression model~\eqref{eq:regression-model} underlying our CASIMAC, we choose a GPR with a combination of a Mat\'{e}rn and a white-noise kernel~\citep{rasmussen2006}. We compare our classifier to a (one-versus-rest) GPC with the same type of kernel. The hyperparameters of both classifiers are tuned by cross-validating the training data over a pre-defined set of different setups as outlined in \cref{sect:hyperparameters}. We made use of scikit-sklearn~\citep{sklearn2011} to realize the standard models.
\par

In total, we perform $\num{10}$ classification tasks, each with different test and training data sets. For each, we use the test data set to determine the accuracy (fraction of correctly predicted points) and the log-loss (that is, the logistic regression loss or cross-entropy loss). In addition to that, we calculate the proba-loss, which we define as the mean predicted probability error 
\begin{align} \label{eq:score-dp}
	\delta p := 1 - \frac{1}{T} \sum_{i=D+1}^{D+T} \widehat{p}(y = y_{\mathrm{true}}(x_i) | \mathcal{D}, x_i),
\end{align}
where $\widehat{p}(y|\mathcal{D},x)$ is our CASIMAC's or the GPC's  class probability prediction, respectively. Clearly,  $\delta p \in [0,1]$ with $0$ being the best possible outcome and $1$ the worst. We can compute this score only because we know the true class membership rule~\eqref{eq:artificial-ytrue} underlying the problem. 
\par

The results are shown in \cref{tab:results:artificial}. Clearly, our method has a better proba-loss and log-loss than the GPC, whereas the latter has a slightly better accuracy.
\par

\begin{table}[!htb]
	\centering
	\captiontitle{Test scores for the synthetic data set based on $T=\num{10000}$ uniformely sampled test datapoints. The proba-loss is defined in \eqref{eq:score-dp}. We show the means and the corresponding standard deviations (in brackets) over all $\num{10}$ classification tasks. The best mean results are highlighted in bold.}{Synthetic data test scores.}\label{tab:results:artificial}
	\begin{tabular}{lcc}\toprule
		\textbf{Score}      & \textbf{CASIMAC}      & \textbf{GPC}          \\ \midrule
		proba-loss          & \bnum{0.106 +- 0.014} & \num{0.552 +- 0.026}  \\ 
		log-loss            & \bnum{0.406 +- 0.332} & \num{0.825 +- 0.055}  \\ 
		accuracy            & \num{0.913 +- 0.017}  & \bnum{0.924 +- 0.018} \\ \bottomrule
	\end{tabular}
\end{table}

\begin{figure}[!htb]
	\begin{center}
		\begin{subfigure}[t]{.45\linewidth}
			\centering
			\includegraphics{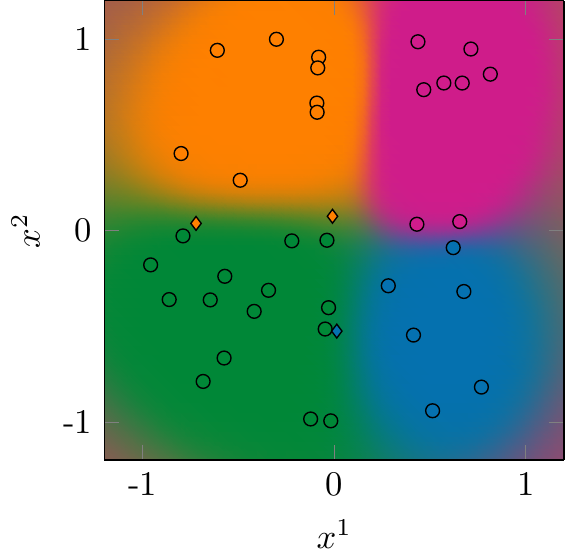}
			\caption{CASIMAC class probabilities.} \label{fig:artificial:results:gprc}
		\end{subfigure}
		\hspace{.25cm}
		\begin{subfigure}[t]{.45\linewidth}
			\centering
			\includegraphics{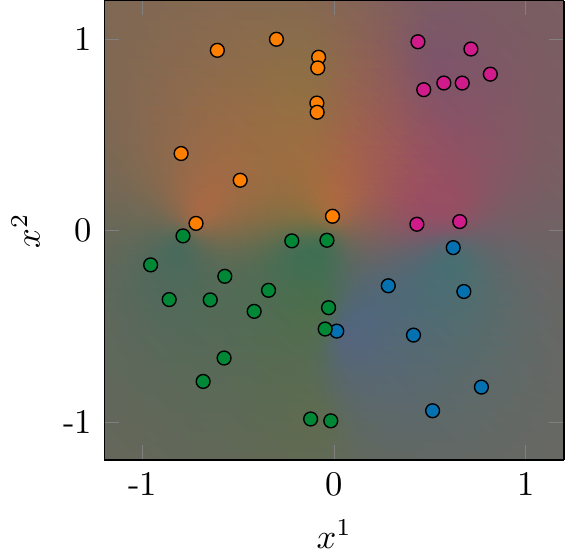}
			\caption{GPC class probabilities.} \label{fig:artificial:results:gpc}
		\end{subfigure}
	\end{center}
	\captiontitle{Class probability predictions of CASIMAC and of GPC for the synthetic data set. The color of each background point corresponds to the weighted average of the class colors from \cref{fig:artificial} with a weight corresponding to the predicted probability of the respective class at this point. Thus, clear colors as in \subref{fig:artificial:results:gprc} represent high probabilities for a single class, whereas washed-out colors as in \subref{fig:artificial:results:gpc} represent almost uniform probabilities. We also show the training data set (consisting of $D=\num{40}$ points) on which the classifiers have been trained. The color of the points corresponds to their true class. While most of the training datapoints are correctly classified (shown as \addlegendimageintext{only marks,mark=*,mark options={fill=white}}), our CASIMAC incorrectly predicts three training datapoints (shown as \addlegendimageintext{only marks,mark=diamond*,mark options={fill=white}}) close to the class borders.}{Synthetic data class probability predictions.} \label{fig:artificial:results}
\end{figure}

We also show an exemplary visualization of the predicted class probabilities in \cref{fig:artificial:results} for a single training data set. The background colors represent a weighted average of the class colors from \cref{fig:artificial} with a weight corresponding to the predicted probability of the respective class. So, for a perfectly calibrated classifier, the colors in \cref{fig:artificial:results} would be the same as the colors in \cref{fig:artificial}. Although our CASIMAC has a slightly lower training accuracy than the GPC ($\num{0.925}$ as opposed to $\num{1.000}$), it clearly exhibits brighter, more distinguishable colors. Consequently, the predicted probabilities are less uniform and correspond to a clearer decision for one of the four classes instead of an uncertain mixture. This observation corresponds to the lower proba-loss and log-loss results in \cref{tab:results:artificial}.

\subsection{Real-world data}
\label{sect:real-world-data}

Following the benchmark on synthetic data, we continue with a benchmark on five real-world data sets from different fields of application. Specifically, we consider the data sets from \cref{tab:datasets}. All of these data sets are publicly available online.

\begin{table}[!htb]
	\centering
	\captiontitle{Overview of the five real-world data sets and their basic characteristics: $n$ is the number of classes, $m$ is the number of features or, in other words, the dimension of the feature space $\mathcal{X} \subset \R^m$, $I$ is the total number of datapoints,  and $D$ is the number of training datapoints. The number $T$ of test datapoints is just $T := I - D$. We abbreviate the references as a)~\citep{uci}, b)~\citep{adak2019}, c)~\citep{lucas2013}, d)~\citep{thorsteinn2014}, e)~\citep{baumgardner2015}, f)~\citep{data-pine}, g)~\citep{niculescu2005}, and h)~\citep{rohra2916}. Also, we turned the originally multi-class data set $\texttt{pine}$ into a binary problem as described in~\citep{niculescu2005}.}{Data set overview.}\label{tab:datasets}
	\begin{tabular}{lccccc} \toprule
		\textbf{Name}    & $n$     & $m$       & $I$         & $D$        & \textbf{Refs.} \\ \midrule
		\texttt{alcohol} & \num{5} & \num{10}  & \num{125}   & \num{50}   & a), b)         \\ 
		\texttt{climate} & \num{2} & \num{18}  & \num{540}   & \num{50}   & a), c)         \\ 
		\texttt{hiv}     & \num{2} & \num{160} & \num{6590}  & \num{125}  & a), d)         \\ 
		\texttt{pine}    & \num{2} & \num{200} & \num{21025} & \num{2000} & e), f), g)     \\ 
		\texttt{wifi}    & \num{4} & \num{7}   & \num{2000}  & \num{500}  & a), h)         \\ \bottomrule
	\end{tabular}
\end{table}

As the distance metric $d$ in the training data transformation~\eqref{eq:training-data-trf-defining-formula} underlying our CASIMAC, we choose the Euclidean distance~\eqref{eq:l^2-distance} on $\mathcal{X} \subset \R^m$ for the data sets \texttt{alcohol}, \texttt{climate}, \texttt{hiv}, and \texttt{pine}, whereas for the remaining data set \texttt{wifi} we choose the taxicab distance on $\mathcal{X} \subset \R^m$ defined by
\begin{align} \label{eq:l^1-distance}
	d(x,x') := \norm{x-x'}_1,
\end{align}
because it leads to a better performance there. Additionally, we parameterize the weights $\alpha, \beta$ of the attraction and repulsion coefficients~\eqref{eq:attraction-and-repulsion-coefficients-definition} as in~\eqref{eq:gamma} for all our data sets. And as the regression model~\eqref{eq:regression-model} underlying our CASIMAC, we again choose a GPR with a combination of a Mat\'{e}rn and a white-noise kernel for all our data sets. 
We compare our CASIMAC on each data set to three other classifiers, namely to a GPC as before and, in addition to that, to an artificial neural network with fully-connected layers (MLP) and to a $k$-nearest neighbor classifier (kNN). This choice of classifiers was dictated by their reported good calibration properties~\citep{niculescu2005}. Again, we tune the hyperparameters of the classifiers by cross-validation as outlined in \cref{sect:hyperparameters}. As in the previous section, we made use of scikit-sklearn~\citep{sklearn2011} to realize the standard models.
\par

The test-training split of the data is performed by means of a stratified random sampling with respect to the class labels. Average test scores over $\num{10}$ classification tasks with different training data sets are reported in \cref{tab:results:test}. Note that for multi-class data sets, the f1 score is calculated as the weighted arithmetic mean over harmonic means~\citep{opitz2019macro}, where the weight is determined by the number of true instances for each class. Analogously, we calculate the precision score (ratio of true positives to the sum of true positives and false positives) and recall score (ratio of true positives to the sum of true positives and false negatives) as weighted averages over all classes.
\par

\begin{table}[!htb]
	\centering
	\captiontitle{Test scores for the benchmarked classifiers on the five real-world data sets from \cref{tab:datasets}. We show the means and the corresponding standard deviations (in brackets) over all $\num{10}$ classification tasks. The best mean results are highlighted in bold.}{Real-world data test scores.}\label{tab:results:test}
	\begin{tabular}{lcccc}\toprule
		\textbf{Score}  & \textbf{CASIMAC}      & \textbf{GPC}          & \textbf{kNN}          & \textbf{MLP}          \\ \midrule
		\multicolumn{5}{c}{\texttt{alcohol}} \\ \midrule
		accuracy   	    & \bnum{0.907 +- 0.062} & \num{0.897 +- 0.044}  & \num{0.545 +- 0.088}  & \num{0.848 +- 0.047}  \\ 
		f1         	    & \bnum{0.901 +- 0.067} & \num{0.894 +- 0.048}  & \num{0.526 +- 0.093}  & \num{0.845 +- 0.053}  \\ 
		precision  	    & \bnum{0.928 +- 0.044} & \num{0.922 +- 0.030}  & \num{0.556 +- 0.106}  & \num{0.872 +- 0.043}  \\ 
		recall          & \bnum{0.907 +- 0.062} & \num{0.897 +- 0.044}  & \num{0.545 +- 0.088}  & \num{0.848 +- 0.047}  \\ 
		log-loss        & \bnum{0.394 +- 0.338} & \num{0.932 +- 0.050}  & \num{1.752 +- 1.442}  & \num{0.755 +- 0.412}  \\ \midrule
		\multicolumn{5}{c}{\texttt{climate}} \\ \midrule 
		accuracy        & \num{0.914 +- 0.002}  & \bnum{0.915 +- 0.004} & \bnum{0.915 +- 0.001} & \num{0.906 +- 0.012}  \\ 
		f1              & \num{0.877 +- 0.007}  & \num{0.879 +- 0.008}  & \num{0.875 +- 0.002}  & \bnum{0.883 +- 0.015} \\ 
		precision       & \num{0.854 +- 0.030}  & \num{0.861 +- 0.030}  & \num{0.854 +- 0.029}  & \bnum{0.874 +- 0.032} \\ 
		recall          & \num{0.914 +- 0.002}  & \bnum{0.915 +- 0.004} & \bnum{0.915 +- 0.001} & \num{0.906 +- 0.012}  \\ 
		log-loss        & \bnum{0.249 +- 0.017} & \num{0.281 +- 0.009}  & \num{1.060 +- 0.337}  & \num{0.341 +- 0.048}  \\ \midrule
		\multicolumn{5}{c}{\texttt{hiv}} \\ \midrule
		accuracy        & \bnum{0.902 +- 0.004} & \num{0.900 +- 0.005}  & \num{0.868 +- 0.007}  & \num{0.880 +- 0.006}  \\ 
		f1              & \num{0.898 +- 0.005}  & \bnum{0.898 +- 0.005} & \num{0.862 +- 0.007}  & \num{0.880 +- 0.006}  \\ 
		precision       & \bnum{0.899 +- 0.004} & \num{0.898 +- 0.005}  & \num{0.861 +- 0.007}  & \num{0.880 +- 0.006}  \\ 
		recall          & \bnum{0.902 +- 0.004} & \num{0.900 +- 0.005}  & \num{0.868 +- 0.007}  & \num{0.880 +- 0.006}  \\ 
		log-loss        & \bnum{0.263 +- 0.017}  & \num{0.290 +- 0.007}  & \num{0.405 +- 0.031}  & \num{0.404 +- 0.018}  \\ \midrule
		\multicolumn{5}{c}{\texttt{pine}} \\ \midrule 
		accuracy        & \bnum{0.937 +- 0.001} & \num{0.934 +- 0.002}  & \num{0.916 +- 0.002}  & \num{0.923 +- 0.005}  \\ 
		f1              & \bnum{0.934 +- 0.001} & \num{0.931 +- 0.002}  & \num{0.914 +- 0.003}  & \num{0.922 +- 0.005}  \\ 
		precision       & \bnum{0.933 +- 0.002} & \num{0.930 +- 0.002}  & \num{0.913 +- 0.003}  & \num{0.922 +- 0.005}  \\ 
		recall          & \bnum{0.937 +- 0.001} & \num{0.934 +- 0.002}  & \num{0.916 +- 0.002}  & \num{0.923 +- 0.005}  \\ 
		log-loss        & \bnum{0.169 +- 0.007} & \num{0.182 +- 0.004}  & \num{0.396 +- 0.117}  & \num{0.212 +- 0.043}  \\ \midrule
		\multicolumn{5}{c}{\texttt{wifi}} \\ \midrule 
		accuracy        & \bnum{0.977 +- 0.002} & \num{0.976 +- 0.002}  & \num{0.968 +- 0.003}  & \num{0.969 +- 0.004}  \\ 
		f1              & \bnum{0.977 +- 0.002} & \num{0.976 +- 0.002}  & \num{0.969 +- 0.003}  & \num{0.969 +- 0.004}  \\ 
		precision       & \bnum{0.978 +- 0.002} & \num{0.977 +- 0.002}  & \num{0.969 +- 0.003}  & \num{0.969 +- 0.004}  \\ 
		recall          & \bnum{0.977 +- 0.002} & \num{0.976 +- 0.002}  & \num{0.968 +- 0.003}  & \num{0.969 +- 0.004}  \\ 
		log-loss        & \num{0.130 +- 0.048}  & \num{0.362 +- 0.048}  & \num{0.195 +- 0.059}  & \bnum{0.093 +- 0.014} \\ \bottomrule
	\end{tabular}
\end{table}

We find from \cref{tab:results:test} that our CASIMAC exhibits a comparatively good overall score on all data sets considered. In particular, its log-loss is better in all cases than that of the GPC. It is also better than the log-loss of all other classifiers except for the \texttt{wifi} data set on which the MLP is superior. Similarly, the accuracy of our approach is also better than that of the other candidates except for the \texttt{climate} data set on which GPC and kNN perform slightly better. Taking the uncertainties of the results into account, it turns out that in most cases the scores fall within the range of a single standard deviation of each other. In particular the log-loss, however, shows the most discrepancies between the classifiers.
\par

\begin{figure}[!htb]
	\centering
	\includegraphics{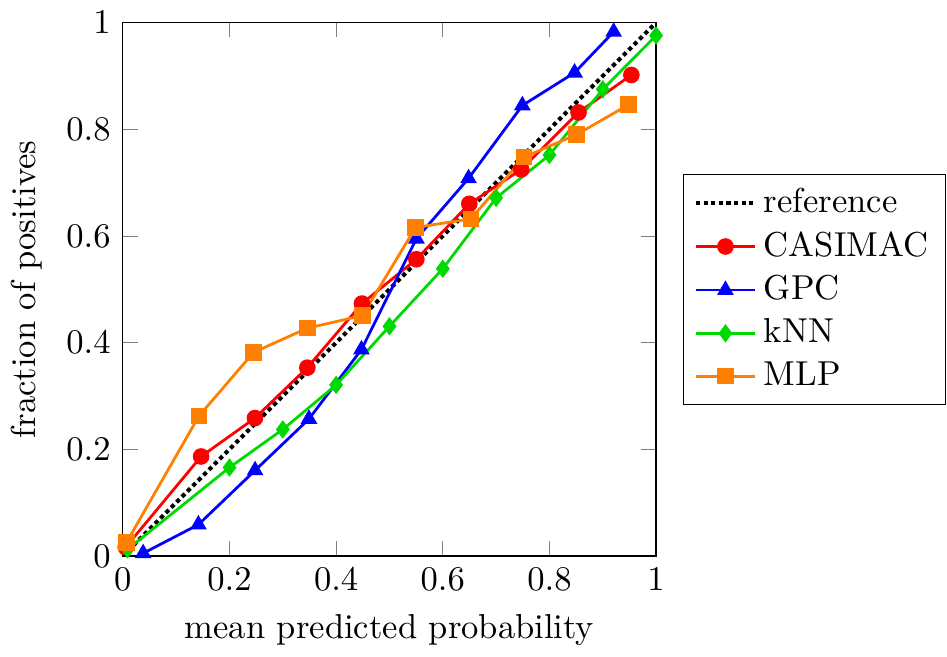}
	\captiontitle{Calibration curves for the benchmarked classifiers on the \texttt{pine} data set. The closer the curves are to the diagonal reference line, the better the calibration of the respective classifier. In particular, our method exhibits the best calibration properties.}{Calibration curves.} \label{fig:calibration-curve}
\end{figure}

For binary classification problems, it is also interesting to analyze the calibration curves (or reliability diagrams) in addition to the scores~\citep{degroot1983, niculescu2005}. For this purpose, we predict the probability of class $1$ for all test samples and discretize the results into ten bins. For each bin, we plot the true fraction of class $1$ against the arithmetic mean of the predicted probabilities. For a perfectly calibrated classifier, such a curve corresponds to a diagonal line and deviations from this line can therefore be understood as miscalibrations. As an example, we consider the \texttt{pine} data set and show a typical calibration curve \cref{fig:calibration-curve}.
\par

Clearly, the curve of our CASIMAC is closer to the diagonal reference line than the curves of the other classifiers. In particular, the GPC curve takes on a sigmoidal shape with major deviations from the diagonal at the beginning and the end. In order to obtain a quantitative measure for the observed miscalibrations, we calculate the corresponding area-deviation, that is, the area between each curve and the diagonal reference line. In case of an optimal calibration, this area vanishes, otherwise it is positive. The results are listed in \cref{tab:results:calibration-curve-area} and allow us to rank the classifiers by calibration quality: CASIMAC gives by far the best calibration result, followed by kNN and MLP, and with GPC at the very end.
\par

\begin{table}[!htb]
	\centering
	\captiontitle{Calibration score measured in terms of the area-deviation (area between each curve and the diagonal reference line) for the calibration curves from \cref{fig:calibration-curve}, which refer to the \texttt{pine} data set. The best result is highlighted in bold.}{Calibration score.}\label{tab:results:calibration-curve-area}
	\begin{tabular}{lcccc}\toprule
		\textbf{Score}  & \textbf{CASIMAC} & \textbf{GPC}     & \textbf{kNN}     & \textbf{MLP}    \\ \midrule
		area-deviation  & \bnum{0.019} & \num{0.062}  & \num{0.044}  & \num{0.058} \\ \bottomrule
	\end{tabular}
\end{table}

Summarized, our benchmark on different real-world data sets shows that our proposed CASIMAC can compete with other well-established classifiers and exhibits comparably good calibration properties.

\subsection{Visualization}
\label{sect:visualization-example}

In order to illustrate the use of our compressed latent space representation for visualization purposes (\cref{sect:visualization}), we consider a simplified version of the data set \texttt{alcohol} from \cref{tab:datasets} and we  refer to this simplified version as \texttt{alcohol-3}. It is obtained from \texttt{alcohol} by merging the last three classes into one and, thus, it has only three instead of five classes (that is, $n=3$ and $m=10$).
\par

We train an exemplary CASIMAC using $D = 25$ points from \texttt{alcohol-3} as training datapoints, while using the remaining $T = 100$ points as test datapoints. As the distance metric $d$ underlying the training data transformation $f$ of our CASIMAC, we choose the Euclidean distance, and for the hyperparameters $\alpha, \beta, k_{\alpha}, k_{\beta}$ of $f$ we choose 
\begin{align*}
	\alpha := 1/2 =: \beta \qquad \text{and} \qquad k_{\alpha} := 1 \qquad \text{and} \qquad k_{\beta} := 5.
\end{align*}
As the regression model $\widehat{f}$ underlying our CASIMAC, in turn, we choose the same GPR as in our previous benchmarks, see \cref{sect:hyperparameters}. 
\par

\begin{figure}[!htb]
	\begin{center}
		\begin{subfigure}[t]{\linewidth}
			\centering\includegraphics{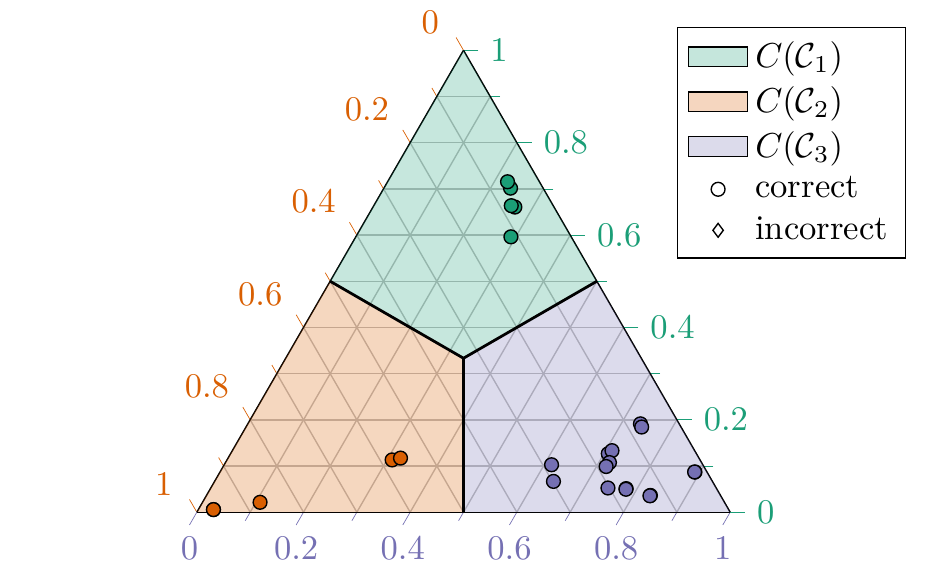}
			\caption{Compressed latent space representation of the training data.} \label{fig:visual:train}
		\end{subfigure}
		\vspace{.1cm}\\
		\begin{subfigure}[t]{\linewidth}
			\centering\includegraphics{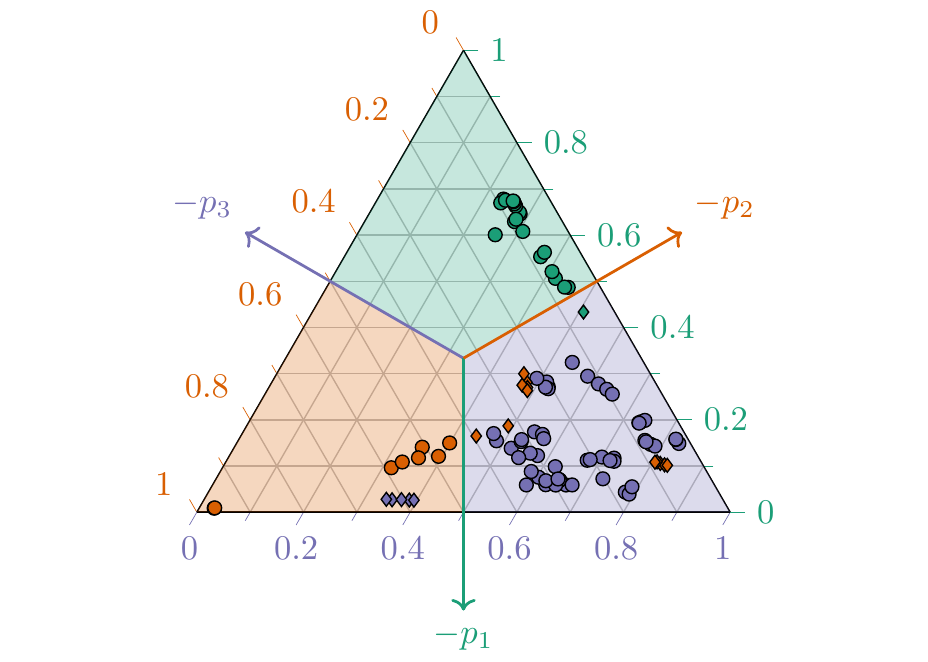}
			\caption{Compressed latent space representation of the test data.} \label{fig:visual:test}
		\end{subfigure}
	\end{center}
	\captiontitle{Images of \subref{fig:visual:train} the training datapoints under the transformation $C \circ f$ and of \subref{fig:visual:test} the test datapoints under the transformation $C \circ \widehat{f}$. See~\eqref{eq:transformed-training-datapoints} and \eqref{eq:transformed-test-datapoints}, respectively. The color of each point represents its true class, while the form of the markers indicates whether a point is correctly classified (\addlegendimageintext{only marks,mark=*,mark options={fill=white}}) or not (\addlegendimageintext{only marks,mark=diamond*,mark options={fill=white}}) by our CASIMAC. All training datapoints are correctly classified by our construction of $f$ and $C$, see~\eqref{eq:training-datapoint-transformed-to-interior-of-its-segment} and~\eqref{eq:compression-respects-segmentation}. The scales on the three simplex edges indicate the barycentric coordinates of the simplex points (Lemma~\ref{lm:barycentric-coordinates}).
	}{Compressed latent space representation.} \label{fig:visual}
\end{figure}

In \cref{fig:visual}, we show the reference simplex $\mathcal{S} \subset \mathcal{Z}$ with the segmentation induced by the segmentation~\eqref{eq:segmentation-of-Z} of $\mathcal{Z}$. So, the differently colored segments in \cref{fig:visual} are nothing but the sets $\mathcal{S}^{\circ} \cap \mathcal{C}_k$ and by~\eqref{eq:compression-respects-segmentation}, in turn, these simplex segments are precisely the images of the cone segments $\mathcal{C}_k$ under the compression map: $\mathcal{S}^{\circ} \cap \mathcal{C}_k = C(\mathcal{C}_k)$ for $k \in \mathcal{Y} = \{1,2,3\}$.
\par

Specifically, \cref{fig:visual:train} displays the compressed latent space representation of the training datapoints $x_1, \dots, x_D$, that is, the points
\begin{align} \label{eq:transformed-training-datapoints}
	w_i := C(f(x_i)) \qquad (i \in \{1,\dots,D\})
\end{align}
where $C$ is the compression map from~\eqref{eq:compression-map}. In view of~\eqref{eq:training-datapoint-transformed-to-interior-of-its-segment} and~\eqref{eq:compression-respects-segmentation}, any one of these transformed datapoints $w_1, \dots, w_D$ belongs to the simplex segment corresponding to its true class label. Also, the distances between points $w_i$ of neighboring classes can indicate inter-class relationships in the original feature space data set. We see from \cref{fig:visual:train}, for instance, that the points $w_i$ belonging to class 1 or 2, respectively, have a larger distance from each other than from the points of class 3. And this leads us 
to the conclusion 
that also the original classes 1 and 2 are more clearly separated than are the classes 1 and 3 and, respectively, the classes 2 and 3. 
\par

In turn, \cref{fig:visual:test} displays the compressed latent space representation of the test datapoints $x_{D+1}, \allowbreak \dots, \allowbreak x_{D+T}$, that is, the points
\begin{align} \label{eq:transformed-test-datapoints}
	w_i := C(\widehat{f}(x_i)) \qquad (i \in \{D+1,\dots,D+T\}). 
\end{align}
Since the regression model is fitted to the training datapoints $(x_1,f(x_1)), \allowbreak \dots, \allowbreak (x_D,f(x_D))$ only, a test point $w_i$ from~\eqref{eq:transformed-test-datapoints} can lie in the simplex segment corresponding to its true class label but it can also lie in a simplex segment corresponding to a false class label. In the latter case, our CASIMAC leads to an incorrect class label prediction and the degree of misclassification can be gathered from the degree of misplacement of $w_i$. As we can see from \cref{fig:visual:test}, there are misclassifications between members of class $1$ and $3$ and especially between members of class $2$ and $3$, but not between points of the classes $1$ and $2$, as one would expect from our previous observation about inter-class distances in the training data set. We list the detailed classification results of the test data set in form of a (transposed) confusion matrix in \cref{tab:visual}. As expected, most misclassifications happen between the classes 2 and 3.

\begin{table}[!htb]
	\centering
	\captiontitle{Confusion matrix (error matrix): the entry in row $k$ and column $l$ is the number of test datapoints which belong to class $k$ and for which our CASIMAC predicts the class label $l$. Correct classifications (on the diagonal) are highlighted in bold. There are no misclassifications between members of the classes $1$ and $2$, as can be expected from \cref{fig:visual:train}.}{Confusion matrix}\label{tab:visual}
	\begin{tabular}{lcccc}\toprule
		\textbf{True class}    & $\widehat{y}=1$ &  $\widehat{y}=2$ &  $\widehat{y}=3$  \\ \midrule
		$1$ (\num{20} members) & \bnum{19}   & \num{0}      & \num{1}       \\ 
		$2$ (\num{20} members) & \num{0}     & \bnum{8}     & \num{12}      \\ 
		$3$ (\num{60} members) & \num{0}     & \num{5}      & \bnum{55}     \\ \bottomrule
	\end{tabular}
\end{table}

\subsection{Towards deep learning}
\label{sect:large-data}

Finally, as a proof of concept for a classification task with a larger training data set, we consider the \texttt{fashion-mnist} data set. It consists of $28\times 28$ pixel images of fashion articles in 8-bit grayscale format (\ie, $\mathcal{X} := \{0,\dots,255\}^{784}$) as described in~\citep{xiao2017}. In total, there are $D = \num{60000}$ training images and $T = \num{10000}$ test images, which are assigned to $n=10$ classes. We perform a min-max normalization of the data  before we feed it to our classifier, so that the individual features lie within the range $[0, 1]$.
\par

Concerning the training data transformation $f$ underlying our CASIMAC, we use two approaches. In the first approach, we make the same ansatz~\eqref{eq:training-data-trf-defining-formula} for the training data transformation $f$ as in all previous examples, 
that is, 
\begin{align} \label{eq:f-naive}
	f = f_{\alpha,\beta,k_{\alpha},k_{\beta},d}.
\end{align} 
As the distance metric $d$, we choose the Euclidean distance, and the hyperparameters $\alpha,\beta,k_{\alpha},k_{\beta}$ we choose to be
\begin{align} \label{eq:hyperparameters-f-naive}
	\alpha := 0, \qquad \beta := 1, \qquad k_{\alpha} := 1, \qquad k_{\beta} := 1.
\end{align}
In the second approach, we make a mixture ansatz for the training data transformation, namely, we take $f$ to be the average of three training data transformations of the form~\eqref{eq:training-data-trf-defining-formula}. In short,  
\begin{align} \label{eq:f-informed}
	f = (f_1 + f_2 + f_3)/3 
	\qquad \text{with} \qquad
	f_i = f_{\alpha,\beta,k_{\alpha},k_{\beta},d_i}
\end{align}
and for all three components $f_1,f_2,f_3$ we choose the hyperparameters $\alpha,\beta,k_{\alpha},k_{\beta}$ as in~\eqref{eq:hyperparameters-f-naive}. As the distance metric $d_1$ for $f_1$, we again choose the Euclidean distance, but the distance metrics $d_2$ and $d_3$ for $f_2$, $f_3$ we choose in a problem-specific way, namely as the similarity metrics defined by
\begin{align*}
	d_2(x,x') := 1-s_5(x,x')
	\qquad \text{and} \qquad 
	d_3(x,x') := 1-s_{13}(x,x').
\end{align*} 
In these definitions, $s_w(x,x') \in [-1,1]$ is the structural similarity index for two images $x$, $x'$ with sliding window size $w$~\citep{wang2004,wang2009} and we use the implementation from~\citep{scikit-image}. In particular, $d_2$ and $d_3$ are valid semimetrics because $s_w$ is symmetric and because $s_w(x,x') = 1$ if and only if $x = x'$, as pointed out on page 106 of~\citep{wang2009}.
\par
Since our first approach~\eqref{eq:f-naive} with its purely Euclidean distance metric does not take into account the structural properties of our image data, it can be considered naive. In contrast, our second approach~\eqref{eq:f-informed} is informed because it brings to bear the fact that the data consists of images, between which a structural relationship can be established. A general overview of such informed machine learning techniques can be found, for instance, in~\citep{informed2019}.  
\par

As the regression model $\widehat{f}$ underlying our CASIMAC, we take a fully connected neural network, both in the naive and in the informed approach. The network contains a first hidden layer with $\num{100}$ neurons and a sigmoid activation function and a second hidden layer with $\num{18}$ neurons and a linear activation function. The output of the second layer is interpreted as the mean and standard deviation of a normal distribution. We optimize the log-likelihood of this distribution with an Adam approach to determine the best model parameters. In total, there are $\num{80318}$ trainable parameters. This model is realized with the help of Tensorflow-probability~\citep{dillon2017}. Since we predict a distribution $\widehat{q}(\,\cdot\,|x)$ for each input point $x$, we can calculate class label probabilities according to~\eqref{eq:class-probability-estimate}.
\par

In Table \ref{tab:results:fashion-mnist}, we summarize the classification results for our CASIMAC based on the naive training data transformation~\eqref{eq:f-naive} and our CASIMAC based on the informed training data transformation~\eqref{eq:f-informed}. Specifically, we show the top-$\num{1}$ to top-$\num{5}$ accuracy scores, which are based on the probability prediction of the classifier. It turns out that our informed approach is slightly better or equal to the naive approach for all accuracies. A list of benchmarked accuracies for other classifiers can be found in~\citep{xiao2017}, for instance.
\par

\begin{table}[!htb]
	\centering
	\captiontitle{Top-$\num{1}$ to top-$\num{5}$ accuracy of our naive and of our informed CASIMAC on the \texttt{fashion-mnist} data set. In  the naive approach we use a purely Euclidean distance metric between the images, whereas the informed approach also takes the structrual image similarity into account. The best scores are highlighted in bold.}{Accuracy for the \texttt{fashion-mnist} data set.}\label{tab:results:fashion-mnist}
	\begin{tabular}{lcc}\toprule
		\textbf{Accuracy} & \textbf{naive} & \textbf{informed} \\ \midrule
		top-$\num{1}$     & \num{0.874}    & \bnum{0.880}      \\ 
		top-$\num{2}$     & \bnum{0.961}   & \bnum{0.961}      \\ 
		top-$\num{3}$     & \bnum{0.984}   & \bnum{0.984}      \\ 
		top-$\num{4}$     & \num{0.990}    & \bnum{0.993}      \\ 
		top-$\num{5}$     & \num{0.993}    & \bnum{0.996}      \\ \bottomrule
	\end{tabular}
\end{table}

In contrast to the approach from~\citep{wilson2016}, we use the neural network to directly perform the latent space mapping instead of linking the network to a series of GPs. Additionally, our network directly predicts the estimated mean and variance of the latent space mapping. It would, however, be a promising approach to further improve our results by incorporating  a more advanced form of feature extraction like the one from~\citep{wilson2016}.

\section{Conclusions and outlook}
\label{sect:conclusions}

In this paper, we have introduced a novel classifier called CASIMAC for multi-class classification in arbitrary semimetrizable feature spaces. It is based on the idea of transforming the classification problem into a regression problem. We achieve this by mapping the training data onto a latent space with a simplex-like geometry and subsequently fitting a regression model to the transformed training data. With the help of this regression model, the predictions of our classifier for the class labels and for the class label probabilities can be obtained in a conceptually and computationally simple manner. We have described in detail how our proposed method works and have demonstrated that it can be successfully applied to real-world data sets from various application domains. In particular, we see three major benefits of our approach.\par

First, it is generic and flexible in the sense that the choice of the particular distance semimetric for our training data transformation and the choice of the regression model underlying our classifier are completely arbitrary. In particular, this capability allows for non-numeric features. Moreover, it enables the integration of additional expert knowledge in the chosen distance metric~\citep{informed2019}. For instance, to classify molecules a distance measure reflecting stoichiometry and configuration variations could be applied~\citep{rupp2012}. Another possible strategy would be to infer the distance metric 
from the data itself, possibly based on certain informed assumptions~\citep{bellet2013}. Similarly, expert knowledge could be brought to bear in the training of the regression model, for example, in the form of shape constraints~\citep{kurnatowski2021,schmid2021,link2022}.
\par

Second, the intuitive latent space representation with its simple geometric concept has a direct interpretation. In particular, this can be exploited to visually detect inter-class relationships and is especially useful for classification problems with a large feature space dimension and a small number of classes. 
\par

Third, as our benchmarks have shown, our method leads to classifiers with comparably good prediction and calibration qualities. To determine class probabilities, we only require a regression model with probabilistic predictions. Also, the effort of computing these class probability predictions is quite low, especially compared to the computational effort necessary for GPC. In particular, no complicated approximations are required in our approach and, for a binary classification problem and a regression model with a normally distributed probabilistic prediction, there even exists a closed-form expression for the class probability predictions.
\par

A challenge of our method is that its training requires the calculation of nearest datapoint neighbors, 
which is computationally expensive for larger data sets~\citep{dhanabal2011}. It would be a natural starting point for further studies to investigate how this computational limitation in the training of our classifiers can be overcome. 
A related challenge of our method is that the tuning of hyperparameters can be costly. Instead of using a cross-validated grid search like we did in this paper, it could be advantageous to consider more elaborate strategies, for example, to infer the hyperparameters from the statistical properties of the training data. We leave this topic as an open question for future research.

\section*{Acknowledgments}
We would like to thank Janis Keuper and J\"urgen Franke for their helpful and constructive comments.
This work was developed in the Fraunhofer Cluster of Excellence \enquote{Cognitive Internet Technologies}. We also gratefully acknowledge additional funding from the Deutsche Forschungsgemeinschaft (DFG, German Research Foundation) within the Priority Programme \enquote{SPP 2331: Machine Learning in Chemical Engineering}.
\FloatBarrier
\begin{appendix}
\section{Segmentation of latent space and core properties of calibrated simplex-mapping classifiers} 
\label{sect:math-background}

In this appendix, we collect some supplementary results needed for a detailed and mathematically sound understanding of our CASIMAC methodology. In particular, we prove the core properties of our segmentation of the latent space and of our CASIMAC. As in the main text, $n$ always stands for an integer larger than or equal to $2$ and
\begin{align}
	\mathcal{Y} := \{1,\dots,n\} 
	\qquad \text{and} \qquad
	\mathcal{Z} := \R^{n-1}
\end{align}
denote the corresponding set of class labels and, respectively, the corresponding latent space. Also, $\norm{z}_2 := (z\cdot z)^{1/2}$ always stands for the standard $\ell^2$-norm of $z \in \mathcal{Z}$, which is induced by the standard scalar product $(z,w) \mapsto z\cdot w = \sum_{i=1}^{n-1} z_i w_i$ on $\mathcal{Z}$.

\subsection{Simplices} 

We begin by briefly discussing $(n-1)$-simplices in $\mathcal{Z}$~\citep{rockafellar1970} and, especially, regular $(n-1)$-simplices in $\mathcal{Z}$~\citep{coxeter1973}. Simplices are the generalization of triangles to arbitrary dimensions: for example, an $(n-1)$-simplex corresponds to a line segment for $n=2$, a triangle for $n=3$, and a tetrahedron for $n=3$. Specifically, a subset $\mathcal{S}$ of $\mathcal{Z}$ is called an \emph{$(n-1)$-dimensional simplex} or, for short, an \emph{$(n-1)$-simplex} in $\mathcal{Z}$ iff it is the convex hull of $n$ affinely independent points $p_1, \dots, p_n$ in $\mathcal{Z}$, that is, iff
\begin{align*}
	\mathcal{S} = \cnv \{p_1,\dots,p_n\} := \bigg\{ z \in \mathcal{Z}: z = \sum_{i \in \mathcal{Y}} \lambda_i p_i \text{ for some } \lambda_i \in [0,\infty) \text{ with } \sum_{i \in \mathcal{Y}} \lambda_i = 1 \bigg\} 
\end{align*} 
for some affinely independent points $p_1, \dots, p_n \in \mathcal{Z}$. As usual, affine independence of the points $p_1, \dots, p_n$ simply means that, for some (and hence every) $k \in \mathcal{Y}$, the set $\{p_i-p_k: i \in \mathcal{Y}\setminus\{k\}\}$ of all vectors connecting $p_k$ with the points $p_i$ is linearly independent. It is straightforward to verify that the affinely independent points $p_1, \dots, p_n$ with $\mathcal{S} = \cnv \{p_1,\dots,p_n\}$ are uniquely determined by the set $\mathcal{S}$, namely as the extreme points of $\mathcal{S}$. (See~\citep{tuy2016} (Proposition~1.17), for instance.) These uniquely determined extreme points $p_1, \dots, p_n$ of $\mathcal{S}$ are also called the \emph{vertices of $\mathcal{S}$}. If $\mathcal{S}$ is an $(n-1)$-simplex with vertices $p_1, \dots, p_n$, then the point
\begin{align}
	c := \frac{1}{n} \sum_{i\in\mathcal{Y}} p_i \in \cnv\{p_1,\dots,p_n\} = \mathcal{S}
\end{align}
is called the \emph{barycenter of $p_1, \dots, p_n$} and, by extension, the \emph{barycenter of $\mathcal{S}$}. Also, the simplex $\mathcal{S}$ is called \emph{regular} iff all of its $n(n-1)/2$ edges have the same (positive) length, that is, iff 
\begin{align}
	\norm{p_i-p_j}_2 = l \qquad (i,j \in \mathcal{Y} \text{ with } i \ne j)
\end{align}
for some $l \in (0,\infty)$, which is also called the edge length of $\mathcal{S}$. See page~121 of~\citep{coxeter1973}, for instance. In our proofs, we repeatedly use the following straightforward fact. 

\begin{lm} \label{lm:vertices-lin-independent}
	Suppose $p_1, \dots, p_n$ are the vertices of an $(n-1)$-simplex in $\mathcal{Z}$ with barycenter $0$. Then 
	\begin{align} \label{eq:vertices-lin-independent}
		\spn\{p_i: i \in \mathcal{Y}\setminus\{k\}\} = \mathcal{Z}
	\end{align}
	for every $k \in \mathcal{Y}$. In particular, for every $k \in \mathcal{Y}$, the $(n-1)$-element subset $\{p_i: i \in \mathcal{Y}\setminus\{k\}\}$ of the vertex set is linearly independent. 
\end{lm}

\begin{proof}
	We give the straightforward proof for the sake of completeness. Since, by the definition of simplex vertices, the set $\{p_i-p_k: i \in \mathcal{Y}\setminus\{k\}\}$ is a linearly independent set of $n-1$ vectors in the $(n-1)$-dimensional vector space $\mathcal{Z}$, we see that
	\begin{align} \label{eq:vertices-lin-independent-1}
		\mathcal{Z} = \spn\{p_i-p_k: i \in \mathcal{Y}\setminus\{k\}\}
	\end{align}
	for every $k \in \mathcal{Y}$. Since, moreover, the barycenter of the vertices $p_1,\dots, p_n$ is $0$, we further see that
	\begin{align} \label{eq:p_k-in-terms-of-other-p_i}
		p_k = - \sum_{i\in\mathcal{Y}\setminus\{k\}} p_i \in \spn\{p_i: i \in \mathcal{Y}\setminus\{k\}\}
	\end{align}
	and therefore
	\begin{align} \label{eq:vertices-lin-independent-2}
		\spn\{p_i-p_k: i \in \mathcal{Y}\setminus\{k\}\} \subset \spn\{p_i: i \in \mathcal{Y}\setminus\{k\}\} \subset \mathcal{Z}
	\end{align}
	for every $k \in \mathcal{Y}$. Combining~\eqref{eq:vertices-lin-independent-1} and~\eqref{eq:vertices-lin-independent-2}, we obtain the asserted equality~\eqref{eq:vertices-lin-independent}. In particular, this equality implies the asserted linear independence by a trivial dimensionality argument.
\end{proof}

We now construct an explicit example of the kind of simplices that underlie our segmentation of $\mathcal{Z}$ and our CASIMAC. In other words, we construct a regular $(n-1)$-simplex in $\mathcal{Z}$ with barycenter $0$ and with vertices $p_1, \dots, p_n$ at unit distance from $0$, that is, 
\begin{align} \label{eq:barycenter-0-and-unit-distance-from-0}
	\sum_{i\in\mathcal{Y}} p_i = 0 
	\qquad \text{and} \qquad
	\norm{p_i}_2 = 1 \qquad (i \in \mathcal{Y}). 
\end{align}
In the extreme special case $n=2$, it is clear that there is only one such simplex, namely the line segment between the vertices $-1$ and $1$ in $\mathcal{Z} = \R$, and we usually index these two  vertices increasingly (instead of decreasingly) then, that is, 
\begin{align} \label{eq:vertex-indexing-in-binary-case}
	p_1 = -1 \qquad \text{and} \qquad p_2 = 1.
\end{align}
In the following, we show that in higher dimensions there still is only one such simplex -- provided that we identify mutually congruent simplices. 

\begin{prop} \label{prop:sample-reg-simplex-in-Z}
	A regular $(n-1)$-simplex with vertices $p_1, \dots, p_n$ satisfying~\eqref{eq:barycenter-0-and-unit-distance-from-0} exists in $\mathcal{Z}$. 
	And, moreover, all other regular $(n-1)$-simplices in $\mathcal{Z}$ with~\eqref{eq:barycenter-0-and-unit-distance-from-0} are congruent to it. 
\end{prop}

\begin{proof}
	We begin by constructing just any regular $(n-1)$-simplex $\cnv\{q_1,\dots,q_n\}$ in $\mathcal{Z}$. In order to do so, we make the ansatz
	\begin{align} \label{eq:sample-reg-simplex-in-Z-1}
		q_i = e_i \qquad (i\in\{1,\dots,n-1\})
		\qquad \text{and} \qquad
		q_n = (\alpha_1,\dots,\alpha_{n-1}),
	\end{align}
	where $e_1,\dots,e_{n-1}$ denote the canonical unit vectors in $\R^{n-1}$ and the coordinates $\alpha_1,\allowbreak \dots,\allowbreak \alpha_{n-1} \in \R$ are to be determined such that the vertices $q_1,\dots,q_n$ all have the same distance from each other. Since, for $i, j \in \{1,\dots,n-1\}$, 
	\begin{gather} 
		\norm{q_i-q_j}_2^2 = \norm{e_i-e_j}_2^2 = 2 \qquad (i\ne j)
		\label{eq:sample-reg-simplex-in-Z-2}\\
		\norm{q_i-q_n}_2^2 = \norm{e_i-q_n}_2^2 = 1 - 2 e_i \cdot q_n + \norm{q_n}_2^2 = 1 - 2\alpha_i + \sum_{l=1}^{n-1} \alpha_l^2
		\label{eq:sample-reg-simplex-in-Z-3}
	\end{gather}
	by virtue of~\eqref{eq:sample-reg-simplex-in-Z-1}, we have to choose the coordinates
	\begin{align} \label{eq:sample-reg-simplex-in-Z-4}
		\alpha_i = \alpha := (\sum_{l=1}^{n-1} \alpha_l^2 - 1)/2
		\qquad (i\in\{1,\dots,n-1\})
	\end{align} 
	to be independent of $i$. Inserting~\eqref{eq:sample-reg-simplex-in-Z-4} into~\eqref{eq:sample-reg-simplex-in-Z-3}, we get a quadratic equation in $\alpha$ which has the solutions 
	\begin{align} \label{eq:sample-reg-simplex-in-Z-5}
		\alpha = \alpha_{\pm} := (1\pm \sqrt{n})/(n-1).
	\end{align}
	In view of these preliminary considerations, we define the points $q_1,\dots,q_n \in \mathcal{Z}$ by
	\begin{align} \label{eq:sample-reg-simplex-in-Z-6}
		q_i := e_i \qquad (i\in\{1,\dots,n-1\})
		\qquad \text{and} \qquad 
		q_n :=  \frac{1 + \sqrt{n}}{n-1} (1,\dots,1). 
	\end{align}
	It is then clear that $\spn\{q_i-q_n: i \in \{1,\dots,n-1\}\} = \mathcal{Z}$ and thus $q_1, \dots, q_n$ are the vertices of an $(n-1)$-simplex in $\mathcal{Z}$. It is also clear by~\eqref{eq:sample-reg-simplex-in-Z-2} and~\eqref{eq:sample-reg-simplex-in-Z-3} that these vertices all have the same distance from each other, namely $\sqrt{2}$. Consequently, $q_1,\dots,q_n$ are the vertices of a regular $(n-1)$-simplex in $\mathcal{Z}$. 
	In particular, by shifting them by their barycenter $c$ and by then  normalizing, we obtain the vertices
	\begin{align} \label{eq:sample-reg-simplex-in-Z-7}
		p_i := (q_i-c)/\nu \qquad (i \in \mathcal{Y})
	\end{align}
	of a regular $(n-1)$-simplex satisfying~\eqref{eq:barycenter-0-and-unit-distance-from-0}, as desired. Specifically,
	\begin{align} \label{eq:sample-reg-simplex-in-Z-8}
		c := \frac{1}{n} \sum_{i\in\mathcal{Y}} q_i = \frac{1+1/\sqrt{n}}{n-1} \, (1,\dots,1)
		\qquad \text{and} \qquad
		\nu := \norm{q_i-c}_2 = \sqrt{1-1/n}. 
	\end{align} 
	It remains to prove that any other regular $(n-1)$-simplex satisfying~\eqref{eq:barycenter-0-and-unit-distance-from-0} is congruent to the simplex constructed above. In order to do so, we have only to notice that any such other simplex has the same edge length as the simplex constructed above and that all regular simplices with equal edge length are congruent to each other.  
\end{proof}

\begin{lm} \label{lm:central-angle-of-reg-simplex}
	Suppose $p_1,\dots,p_n$ are the vertices of a regular $(n-1)$-simplex in $\mathcal{Z}$ with~\eqref{eq:barycenter-0-and-unit-distance-from-0}. Then 
	\begin{align} \label{eq:simplex-central-angle-of-reg-simplex}
		p_i \cdot p_j = 1 \qquad (i=j) 
		\qquad \text{and} \qquad
		p_i \cdot p_j = -1/(n-1) \qquad (i\ne j). 
	\end{align}
\end{lm}

\begin{proof}
	An immediate consequence of~\citep{parks2002} (proof of Theorem~1). 
\end{proof}

\subsection{Simplex-induced segmentation of the latent space}

In this section, we introduce the segmentation of the latent space $\mathcal{Z}$ underlying our classification method and establish the core properties of this segmentation. We choose a segmentation into $n$ convex cone segments $\mathcal{C}_1, \dots, \mathcal{C}_n$ and we define these cone segments in terms of the vertices of an $(n-1)$-simlex with barycenter $0$. As usual, a convex cone in $\mathcal{Z}$ is a convex subset of $\mathcal{Z}$ that is closed under addition and under scalar multiplication with positive scalars~\citep{tuy2016} (Proposition~1.7). 

\begin{lm} \label{lm:barycentric-char-of-C_k}
	Suppose $p_1,\dots,p_n$ are the vertices of an $(n-1)$-simplex in $\mathcal{Z}$ with~(\ref{eq:barycenter-0-and-unit-distance-from-0}.a) and, for every $k \in \mathcal{Y}$, let 
	\begin{align} \label{eq:def-of-C_k}
		\mathcal{C}_k := \bigg\{z \in \mathcal{Z}: z = \sum_{i\in\mathcal{Y}\setminus\{k\}} c_i \cdot (-p_i) \text{ for some } c_i \in [0,\infty) \bigg\}
	\end{align}
	be the convex cone generated by the mirrored vertices $-p_i$ for $i \in \mathcal{Y}\setminus\{k\}$. Suppose further that $z$ is a point in $\mathcal{Z}$ having the representation
	\begin{align} \label{eq:barycentric-representation-of-z}
		z = \sum_{i\in\mathcal{Y}} r_i \cdot (-p_i)
	\end{align}
	for some $r_1, \dots, r_n \in \R$. Then, for every $k \in \mathcal{Y}$, 
	one has the following equivalence: $z \in \mathcal{C}_k$ if and only if $r_k \le r_i$ for all $i \in \mathcal{Y}$.
\end{lm}

\begin{proof}
	Choose and fix an arbitrary $k \in \mathcal{Y}$. It then follows from~\eqref{eq:barycentric-representation-of-z} with the help of~\eqref{eq:p_k-in-terms-of-other-p_i} that
	\begin{align} \label{eq:barycentric-char-of-C_k-1}
		z = r_k \sum_{i\in\mathcal{Y}\setminus\{k\}} p_i + \sum_{i\in\mathcal{Y}\setminus\{k\}} r_i \cdot (-p_i) = \sum_{i\in\mathcal{Y}\setminus\{k\}} (r_i - r_k) \cdot (-p_i).
	\end{align}
	If $r_k \le r_i$ for all $i \in \mathcal{Y}$, then it immediately follows from~\eqref{eq:barycentric-char-of-C_k-1} and the definition~\eqref{eq:def-of-C_k} of $\mathcal{C}_k$ that $z \in \mathcal{C}_k$, as desired.
	If, conversely, $z \in \mathcal{C}_k$, then 
	\begin{align} \label{eq:barycentric-char-of-C_k-2}
		z = \sum_{i\in\mathcal{Y}\setminus\{k\}} c_i \cdot (-p_i)
	\end{align}
	for some $c_i \in [0,\infty)$ by the definition~\eqref{eq:def-of-C_k} of $\mathcal{C}_k$ and, therefore, it follows by virtue of~\eqref{eq:barycentric-char-of-C_k-1} and~\eqref{eq:barycentric-char-of-C_k-2} that
	\begin{align} \label{eq:barycentric-char-of-C_k-3}
		\sum_{i\in\mathcal{Y}\setminus\{k\}} (r_i - r_k) \cdot p_i = -z = \sum_{i\in\mathcal{Y}\setminus\{k\}} c_i \cdot p_i.
	\end{align}
	Since $\{p_i: i \in \mathcal{Y}\setminus\{k\}\}$ is linearly independent by Lemma~\ref{lm:vertices-lin-independent}, we conclude from~\eqref{eq:barycentric-char-of-C_k-3} that $r_i-r_k = c_i \ge 0$ for all $i \in \mathcal{Y}\setminus\{k\}$ and therefore $r_k \le r_i$ for all $i \in \mathcal{Y}$, as desired.
\end{proof}

\begin{lm} \label{lm:closure-and-interior-of-C_k}
	Suppose $p_1,\dots,p_n$ are the vertices of an $(n-1)$-simplex in $\mathcal{Z}$ with~(\ref{eq:barycenter-0-and-unit-distance-from-0}.a) and let $\mathcal{C}_k$ be the cone defined in~\eqref{eq:def-of-C_k}. Then the closure $\ol{\mathcal{C}}_k$ and the interior $\mathcal{C}_k^{\circ}$ of $\mathcal{C}_k$ are given, respectively, by
	\begin{align} \label{eq:closure-and-interior-of-C_k}
		\ol{\mathcal{C}}_k = \mathcal{C}_k 
		\quad \text{and} \quad 
		\mathcal{C}_k^{\circ} = \bigg\{z \in \mathcal{Z}: z = \sum_{i\in\mathcal{Y}\setminus\{k\}} c_i \cdot (-p_i) \text{ for some } c_i \in (0,\infty) \bigg\}.
	\end{align}
	In particular, $\mathcal{C}_k$ is a closed set in $\mathcal{Z}$. 
\end{lm}

\begin{proof}
	Choose and fix an arbitrary $k \in \mathcal{Y}$ and let $\phi_k: \R^{n-1} \to \mathcal{Z}$ be the linear map defined by $\phi_k(e_i) = -p_i$ for $i \in \{1, \dots, k-1\}$ and $\phi_k(e_i) = -p_{i+1}$ for $i \in \{k,\dots,n-1\}$, where $e_1, \dots, e_{n-1}$ denote the canonical unit vectors in $\R^{n-1}$. It is then obvious that the cone $\mathcal{C}_k$ is the image under $\phi_k$ of the non-negative orthant $P := [0,\infty)^{n-1}$ in $\R^{n-1}$. In short,
	\begin{align} \label{eq:closure-and-interior-of-C_k-1}
		\mathcal{C}_k = \phi_k(P).
	\end{align}
	It is also clear, by the linear independence of $\{\phi_k(e_i): i \in \{1,\dots,n-1\}\}$ (Lemma~\ref{lm:vertices-lin-independent}), that $\phi_k$ is a linear isomorphism of $\mathcal{Z} = \R^{n-1}$ and thus also a homeomorphism of $\mathcal{Z}$ (Theorem~VII.1.6 in of~\citep{amann2008}). Since obviously $\ol{P} = P$ and $P^{\circ} = (0,\infty)^{n-1}$, we see by~\eqref{eq:closure-and-interior-of-C_k-1} and~\citep{dugundji1966} (Theorem~III.11.3 and Theorem~III.11.4) that 
	\begin{align*}
		\ol{\mathcal{C}}_k = \ol{\phi_k(P)} = \phi_k(\ol{P}) = \phi_k(P) = \mathcal{C}_k,
		\qquad
		\mathcal{C}_k^{\circ} = (\phi_k(P))^{\circ} = \phi_k(P^{\circ}) = \phi_k((0,\infty)^{n-1}).
	\end{align*}
	Clearly, $\phi_k((0,\infty)^{n-1})$ is equal to the set on the right-hand side of~(\ref{eq:closure-and-interior-of-C_k}.b) and thus the proof is finished. 
\end{proof}

With the help of the preceding lemmas, we can prove that the cone segments cover the whole latent space and that they are essentially non-overlapping (up to boundary points). It should be noticed that for this result, we do not have to require the underlying $(n-1)$-simplex to be regular.

\begin{prop} \label{prop:segmentation-of-Z}
	Suppose $p_1,\dots,p_n$ are the vertices of an $(n-1)$-simplex in $\mathcal{Z}$ with~(\ref{eq:barycenter-0-and-unit-distance-from-0}.a) and let $\mathcal{C}_1, \dots, \mathcal{C}_n$ be the cones defined in~\eqref{eq:def-of-C_k}. Then, for every $k \in \mathcal{Y}$, the vertex $p_k$ lies on the central ray
	\begin{align} \label{eq:central-ray-of-C_k}
		\bigg\{z \in \mathcal{Z}: z = \sum_{i\in\mathcal{Y}\setminus\{k\}} c \cdot (-p_i) \text{ for some } c \in [0,\infty)\bigg\}
	\end{align}
	of the cone $\mathcal{C}_k$. Also, the cones $\mathcal{C}_1, \dots, \mathcal{C}_n$ cover the whole of $\mathcal{Z}$, and any two cones $\mathcal{C}_k$ and $\mathcal{C}_l$ for $k\ne l$ overlap only at their boundaries. In short,
	\begin{align} \label{eq:segmentation-of-Z-appendix}
		\mathcal{Z} = \bigcup_{k \in \mathcal{Y}} \mathcal{C}_k
		\qquad \text{and} \qquad
		\mathcal{C}_k^{\circ} \cap \mathcal{C}_l = \emptyset \qquad (k\ne l). 
	\end{align}
\end{prop}

\begin{proof}
	In view of~\eqref{eq:p_k-in-terms-of-other-p_i}, it is obvious that the vertex $p_k$ lies on the central ray~\eqref{eq:central-ray-of-C_k} of $\mathcal{C}_k$ for every $k \in \mathcal{Y}$.
	\par
	
	In order to prove~(\ref{eq:segmentation-of-Z-appendix}.a), we have only to show that every $z \in \mathcal{Z}$ is contained in some $\mathcal{C}_k$. Let $z \in \mathcal{Z}$ be an arbitrary point in $\mathcal{Z}$. It then follows by~\eqref{eq:vertices-lin-independent} that $z$ has a representation of the form~\eqref{eq:barycentric-representation-of-z} for some $r_1, \dots, r_n \in \R$. So, by Lemma~\ref{lm:barycentric-char-of-C_k}, we see that $z \in \mathcal{C}_k$ for every $k$ with $r_k = \min_{i\in\mathcal{Y}} r_i$.
	\par
	
	In order to prove~(\ref{eq:segmentation-of-Z-appendix}.b), we argue by contradiction. Let $k,l \in \mathcal{Y}$ be fixed with $k\ne l$ and assume, for the sake of argument, that there exists a point $z \in \mathcal{Z}$ with
	\begin{align} \label{eq:segmentation-of-Z-1}
		z \in \mathcal{C}_k^{\circ} \cap \mathcal{C}_l. 
	\end{align}
	In view of~\eqref{eq:vertices-lin-independent}, $z$ has a representation of the form~\eqref{eq:barycentric-representation-of-z} for some $r_1, \dots, r_n \in \R$. Since $z \in \mathcal{C}_k \cap \mathcal{C}_l$ by assumption~\eqref{eq:segmentation-of-Z-1}, we therefore see by Lemma~\ref{lm:barycentric-char-of-C_k} that
	\begin{align} \label{eq:segmentation-of-Z-2}
		r_k = \min_{i\in\mathcal{Y}} r_i = r_l.
	\end{align}
	In order to arrive at a contradiction, we consider suitable perturbations $z^{\eps}$ of $z$, namely
	\begin{align} \label{eq:segmentation-of-Z-3}
		z^{\eps} := \sum_{i\in\mathcal{Y}} r_i^{\eps} \cdot (-p_i)
		\qquad (\eps \in (0,\infty)),
	\end{align}
	where $r_i^{\eps} := r_i + \eps$ for $i \in \mathcal{Y}\setminus\{l\}$ and $r_l^{\eps} := r_l$. It is then clear by~\eqref{eq:segmentation-of-Z-2} that $r_l^{\eps} = r_l = r_k < r_k + \eps = r_k^{\eps}$ for every $\eps \in (0,\infty)$ 
	and therefore we see by Lemma~\ref{lm:barycentric-char-of-C_k} that
	\begin{align} \label{eq:segmentation-of-Z-5}
		z^{\eps} \notin \mathcal{C}_k 
		\qquad (\eps \in (0,\infty)).
	\end{align}
	Since, on the other hand, $z \in \mathcal{C}_k^{\circ}$ by assumption~\eqref{eq:segmentation-of-Z-1} and the perturbations $z^{\eps}$ obviously converge to $z$, we also see  that
	\begin{align} \label{eq:segmentation-of-Z-6}
		z^{\eps} \in \mathcal{C}_k 
		\qquad (\eps \in (0,\eps_0])
	\end{align} 
	for some sufficiently small $\eps_0 > 0$. Contradiction between~\eqref{eq:segmentation-of-Z-5} and~\eqref{eq:segmentation-of-Z-6}! So, our  assumption~\eqref{eq:segmentation-of-Z-1} cannot be true, as desired. 
\end{proof}

If we additionally require the $(n-1)$-simplex from the previous proposition to be regular, we can further prove that the corresponding cone segments are mutually congruent. 

\begin{prop} \label{prop:C_k-and-C_l-congruent}
	Suppose $p_1,\dots,p_n$ are the vertices of a regular $(n-1)$-simplex in $\mathcal{Z}$ with~\eqref{eq:barycenter-0-and-unit-distance-from-0} and let $\mathcal{C}_1, \dots, \mathcal{C}_n$ be the cone segments defined in~\eqref{eq:def-of-C_k}. Then any two of these cone segments are congruent to each other. 
\end{prop}

\begin{proof}
	We proceed in three steps and, for the entire proof, we fix $k,l \in \mathcal{Y}$ with $k \ne l$. 
	As a first step, we show that the mid-perpendicular hyperplane 
	\begin{align} \label{eq:def-M_kl}
		\mathcal{M}_{kl} := \{z \in \mathcal{Z}: z \cdot (p_k-p_l) = 0 \}
	\end{align}
	between the vertices $p_k$ and $p_l$ is equal to the subspace of $\mathcal{Z}$ spanned by all other vertices, that is,
	\begin{align} \label{eq:C_k-and-C_l-congruent-1}
		\mathcal{M}_{kl} = \spn\{p_i: i \in \mathcal{Y}\setminus\{k,l\}\}. 
	\end{align}
	Indeed, in view of Lemma~\ref{lm:central-angle-of-reg-simplex}, we have $p_i \cdot (p_k-p_l) = 0$ for all $i \in \mathcal{Y}\setminus\{k,l\}$ and therefore
	\begin{align} \label{eq:C_k-and-C_l-congruent-2}
		\{p_i: i \in \mathcal{Y}\setminus\{k,l\}\} \subset \mathcal{M}_{kl}.
	\end{align}
	Since $\{p_i: i \in \mathcal{Y}\setminus\{k,l\}\}$ is a linearly independent set of $n-2$ vectors by Lemma~\ref{lm:vertices-lin-independent} and $\mathcal{M}_{kl}$ is an $(n-2)$-dimensional subspace of $\mathcal{Z}$, the asserted equality~\eqref{eq:C_k-and-C_l-congruent-1} follows from~\eqref{eq:C_k-and-C_l-congruent-2} and a trivial dimensionality argument.
	\par
	
	As a second step, we show that the segment $\mathcal{C}_l$ is the mirror image of the segment $\mathcal{C}_k$ under the reflection $\rho_{kl}$ at the mid-perpendicular hyperplane $\mathcal{M}_{kl}$, 
	that is,
	\begin{align} \label{eq:C_k-and-C_l-congruent-3}
		\rho_{kl}(\mathcal{C}_k) = \mathcal{C}_l,
	\end{align}
	Indeed, the reflection at $\mathcal{M}_{kl}$ is the linear map $\rho_{kl}$ given by
	\begin{align} \label{eq:C_k-and-C_l-congruent-4}
		\rho_{kl}(z) = \pi_{kl}(z) - (z-\pi_{kl}(z)) = 2\pi_{kl}(z) - z 
		\qquad (z\in \mathcal{Z}),
	\end{align}
	where $\pi_{kl}$ is the orthogonal projection onto $\mathcal{M}_{kl}$. 
	It then follows 
	by the first step that $\pi_{kl}(p_i) = p_i$ for all $i \in \mathcal{Y}\setminus\{k,l\}$ and that $\pi_{kl}(p_k) = (p_k+p_l)/2 = \pi_{kl}(p_l)$ and therefore
	\begin{align} \label{eq:C_k-and-C_l-congruent-5}
		\rho_{kl}(p_i) = p_i \quad (i \in \mathcal{Y}\setminus\{k,l\})
		\quad \text{and} \quad
		\rho_{kl}(p_k) = p_l 
		\quad \text{and} \quad
		\rho_{kl}(p_l) = p_k. 
	\end{align}
	And from this, in turn, the asserted equality~\eqref{eq:C_k-and-C_l-congruent-3} immediately follows by the definition~\eqref{eq:def-of-C_k} of the cone segments.  
	\par
	
	As a third step, we finally establish the asserted congruence of the segments $\mathcal{C}_k$ and $\mathcal{C}_l$. 
	Indeed, this immediately follows by the second step, because reflections obviously are congruence transformations (isometries) of $\mathcal{Z}$. 
\end{proof}

With the preceding results at hand, we can establish a distance-based characterization of our cone segments which is central for an economical computation of our classifier's class label and class label probability predictions $\widehat{y}(x)$ and  $\widehat{p}(\,\cdot\,|x)$, respectively. See~\eqref{eq:casimac-computation-appendix} and~\eqref{eq:class-probability-approximant} below.

\begin{thm} \label{thm:norm-char-of-C_k}
	Suppose $p_1,\dots,p_n$ are the vertices of a regular $(n-1)$-simplex in $\mathcal{Z}$ with~\eqref{eq:barycenter-0-and-unit-distance-from-0} and let $\mathcal{C}_1, \dots, \mathcal{C}_n$ be the cone segments defined in~\eqref{eq:def-of-C_k}. Then these cone segments can be characterized in terms of the distances from the central vectors $p_1, \dots, p_n$ of the cones $\mathcal{C}_1, \dots, \mathcal{C}_n$. Specifically,
	\begin{align} \label{eq:norm-char-of-C_k}
		\mathcal{C}_k = \big\{ z \in \mathcal{Z}: \norm{z-p_k}_2 \le \norm{z-p_l}_2 \text{ for all } l \in \mathcal{Y} \big\}.
	\end{align}
	And analogously, 
	for the interiors of the cone segments one has the following characterization:
	\begin{align} \label{eq:norm-char-of-C_k^circ}
		\mathcal{C}_k^{\circ} = \big\{ z \in \mathcal{Z}: \norm{z-p_k}_2 < \norm{z-p_l}_2 \text{ for all } l \in \mathcal{Y}\setminus\{k\} \big\}.
	\end{align}
\end{thm}

\begin{proof}
	We proceed in four steps and, for the entire proof, we fix $k \in \mathcal{Y}$.
	As a first step, we observe that the set on the right-hand side of~\eqref{eq:norm-char-of-C_k} is nothing but the intersection of the  half-spaces
	\begin{align} \label{eq:def-H_kl}
		\mathcal{H}_{kl}(p_k) := \{z \in \mathcal{Z}: z\cdot (p_k-p_l) \ge 0\}
	\end{align}
	for all $l \in \mathcal{Y} \setminus \{k\}$, that is, the half-spaces confined by the mid-perpendicular hyperplanes $\mathcal{M}_{kl}$ from~\eqref{eq:def-M_kl} and stretching to the side of $p_k$ (instead of its mirror image $p_l$). In other words, as a first step we observe that 
	\begin{align} \label{eq:norm-char-of-C_k-1}
		\big\{ z \in \mathcal{Z}: \norm{z-p_k}_2 \le \norm{z-p_l}_2 \text{ for all } l \in \mathcal{Y} \big\} = \bigcap_{l \in \mathcal{Y} \setminus \{k\}} \mathcal{H}_{kl}(p_k).
	\end{align}
	Indeed, by~(\ref{eq:barycenter-0-and-unit-distance-from-0}.b) we immediately obtain
	\begin{align} \label{eq:norm-char-of-C_k-1.1}
		\norm{z-p_k}^2_2 = \norm{z}_2^2 - 2 z\cdot p_k + 1
		\qquad \text{and} \qquad
		\norm{z-p_l}^2_2 = \norm{z}_2^2 - 2 z\cdot p_l + 1
	\end{align}
	for every $l \in \mathcal{Y}$ and every $z \in \mathcal{Z}$. And from this, in turn, the asserted equality~\eqref{eq:norm-char-of-C_k-1} readily follows. 
	\par
	
	As a second step, we show that the segment $\mathcal{C}_k$ is contained in the right-hand side of~\eqref{eq:norm-char-of-C_k} or, equivalently (by the first step), that 
	\begin{align} \label{eq:norm-char-of-C_k-2}
		\mathcal{C}_k \subset \bigcap_{l \in \mathcal{Y} \setminus \{k\}} \mathcal{H}_{kl}(p_k). 
	\end{align}
	So, let $z \in \mathcal{C}_k$. We can then represent $z$ in the form~\eqref{eq:barycentric-char-of-C_k-2} with some $c_i \in [0,\infty)$ and we can thus conclude, using Lemma~\ref{lm:central-angle-of-reg-simplex}, that
	\begin{align*}
		z \cdot (p_k-p_l) 
		= c_l  (-p_l) \cdot (p_k-p_l) + \sum_{i \in \mathcal{Y}\setminus\{k,l\}} c_i  (-p_i) \cdot (p_k-p_l)  = c_l (1+1/(n-1)) \ge 0
	\end{align*}
	for every $l \in \mathcal{Y} \setminus \{k\}$. 
	And therefore, $z$ belongs to the set on the right-hand side of~\eqref{eq:norm-char-of-C_k-2}, as desired.
	\par
	
	As a third step, we show that conversely the right-hand side of~\eqref{eq:norm-char-of-C_k} is contained in the segment $\mathcal{C}_k$ or, equivalently (by the first step), that
	\begin{align} \label{eq:norm-char-of-C_k-3}
		\bigcap_{l \in \mathcal{Y} \setminus \{k\}} \mathcal{H}_{kl}(p_k)
		\subset \mathcal{C}_k.
	\end{align}
	So, let $z \in \bigcap_{l \in \mathcal{Y} \setminus \{k\}} \mathcal{H}_{kl}(p_k)$. We then have $z \cdot (p_k-p_l) \ge 0$ for every $l \in \mathcal{Y} \setminus \{k\}$ and therefore we see, using Lemma~\ref{lm:central-angle-of-reg-simplex}, that the perturbations
	\begin{align}
		z^{\eps} := z + \eps w := z + \eps \sum_{i \in \mathcal{Y}\setminus\{k\}} (p_k-p_i)
		\qquad (\eps \in (0,\infty))
	\end{align}
	of $z$ satisfy the strict inequality
	\begin{align} \label{eq:norm-char-of-C_k-3.1}
		z^{\eps} \cdot (p_k-p_l) = z \cdot (p_k-p_l) + \eps \norm{p_k-p_l}_2^2 + \eps (n-2)(1+1/(n-1)) > 0
	\end{align}
	for every $l \in \mathcal{Y} \setminus \{k\}$ and every $\eps \in (0,\infty)$. So, by~\eqref{eq:norm-char-of-C_k-1.1} with $z$ replaced by $z^{\eps}$, this strict inequality~\eqref{eq:norm-char-of-C_k-3.1} implies
	\begin{align} \label{eq:norm-char-of-C_k-3.2}
		\norm{z^{\eps}-p_k}_2^2 < \norm{z^{\eps}-p_l}_2^2
		\qquad (l \in \mathcal{Y} \setminus \{k\} \text{ and } \eps \in (0,\infty)).
	\end{align}
	And from~\eqref{eq:norm-char-of-C_k-3.2}, in turn, we can easily conclude that
	\begin{align} \label{eq:norm-char-of-C_k-3.3}
		z^{\eps} \in \mathcal{C}_k \qquad (\eps \in (0,\infty)). 
	\end{align}
	Indeed, by~(\ref{eq:segmentation-of-Z-appendix}.a), we see that for every $\eps \in (0,\infty)$ there exists an index $k' = k'_{\eps} \in \mathcal{Y}$ such that $z^{\eps} \in \mathcal{C}_{k'}$ and therefore, by the first and the second step with $k$ replaced by $k'$, 
	\begin{align}
		\norm{z^{\eps}-p_{k'}}_2 \le \norm{z^{\eps}-p_l}_2 
		\qquad (l \in \mathcal{Y} \setminus \{k'\}).
	\end{align}
	Clearly, this is compatible with~\eqref{eq:norm-char-of-C_k-3.2} only if $k' = k$. Consequently, we must have $k' = k$ so that $z^{\eps} \in \mathcal{C}_{k'} = \mathcal{C}_k$ and~\eqref{eq:norm-char-of-C_k-3.3} is established. Since the perturbations $z^{\eps}$ obviously converge to $z$, we see from~\eqref{eq:norm-char-of-C_k-3.3} and~(\ref{eq:closure-and-interior-of-C_k}.a) that $z \in \mathcal{C}_k$, as desired.
	\par
	
	As a fourth step, we finally establish the asserted equalities~\cref{eq:norm-char-of-C_k,eq:norm-char-of-C_k^circ}. 
	Indeed, \eqref{eq:norm-char-of-C_k} is an immediate consequence of~\cref{eq:norm-char-of-C_k-1,eq:norm-char-of-C_k-2,eq:norm-char-of-C_k-3}. And \eqref{eq:norm-char-of-C_k^circ}, in turn, is readily obtained by noting that the interior of the half-space $\mathcal{H}_{kl}(p_k)$ is given by
	\begin{align}
		(\mathcal{H}_{kl}(p_k))^{\circ} = \{z \in \mathcal{Z}: z\cdot (p_k-p_l) > 0\}
	\end{align}
	and by then taking the interior on both sides of the inclusions~\eqref{eq:norm-char-of-C_k-2} and~\eqref{eq:norm-char-of-C_k-3}. 
\end{proof}

In view of the normalization condition~(\ref{eq:barycenter-0-and-unit-distance-from-0}.b) on the vertices of our simplex, the norm characterization~\eqref{eq:norm-char-of-C_k} immediately translates to a scalar-product characterization of our cone segments. We will use it to prove the invariance of the cone segments under the compression and inflation maps from~\eqref{eq:def-of-compression-map} and~\eqref{eq:def-of-inflation-map} below. 

\begin{cor} \label{cor:scalar-product-char-of-C_k}
	Suppose $p_1,\dots,p_n$ are the vertices of a regular $(n-1)$-simplex in $\mathcal{Z}$ with~\eqref{eq:barycenter-0-and-unit-distance-from-0} and let $\mathcal{C}_1, \dots, \mathcal{C}_n$ be the cone segments defined in~\eqref{eq:def-of-C_k}. Then 
	\begin{align} \label{eq:scalar-product-char-of-C_k}
		\mathcal{C}_k = \big\{ z \in \mathcal{Z}: z \cdot p_k \ge z \cdot p_l  \text{ for all } l \in \mathcal{Y} \big\}.
	\end{align}
\end{cor}

\begin{proof}
	An immediate consequence of~\eqref{eq:norm-char-of-C_k} in conjunction with~\eqref{eq:norm-char-of-C_k-1.1}.
\end{proof}

\subsection{Core properties of calibrated simplex-mapping classifiers} 

In this section, we establish the core properties of our CASIMAC. We begin by proving that the underlying training data transformation $f$ maps every training datapoint $x_i$ to the interior $\mathcal{C}_{y(x_i)}^{\circ}$ of the cone segment corresponding to the true class label $y(x_i) = y_i$. 

\begin{prop} \label{prop:generic-properties-of-training-data-trf}
	Suppose $p_1,\dots,p_n$ are the vertices of a regular $(n-1)$-simplex in $\mathcal{Z}$ with~\eqref{eq:barycenter-0-and-unit-distance-from-0} and let $\mathcal{C}_1, \dots, \mathcal{C}_n$ be the cone segments defined in~\eqref{eq:def-of-C_k}. Suppose further that $\mathcal{D} = \{(x_i,y_i): i \in \{1,\dots,D\}\}$ is a finite subset of $\mathcal{X} \times \mathcal{Y}$ and that the map $f: \mathcal{D}_{\mathcal{X}} := \{x_1,\dots,x_D\} \to \mathcal{Z}$ is defined as in~\eqref{eq:training-data-trf-defining-formula}, \eqref{eq:attraction-and-repulsion-coefficients-definition}, and \eqref{eq:nearest-neighbors-definition} with a semimetric $d$ on $\mathcal{X}$ and hyperparameters $\alpha,\beta \in [0,\infty)$ and $k_{\alpha}, k_{\beta} \in \N$ satisfying~\eqref{eq:conditions-on-hyperparameters}. Then 
	\begin{align} \label{eq:generic-properties-of-training-data-trf}
		0 < A_{k_{\alpha},d}(x), R_{k_{\beta,d}}(x,y) < \infty 
	\end{align}
	for every $x \in \mathcal{D}_{\mathcal{X}}$ and $y \in \mathcal{Y} \setminus \{y(x)\}$. In particular, $f(x) \in \mathcal{C}_{y(x)}^{\circ}$ for every $x \in \mathcal{D}_{\mathcal{X}}$. 
\end{prop}

\begin{proof}
	Choose and fix $x \in \mathcal{D}_{\mathcal{X}}$. Also, for every $y \in \mathcal{Y}$, let $\mathcal{D}_{\mathcal{X},y} := \{x' \in \mathcal{D}_{\mathcal{X}}: y(x') = y\}$ be the set of training datapoints belonging to class $y$ and let $c := \min\{ |\mathcal{D}_{\mathcal{X},y}|: y \in \mathcal{Y}\}$ be the size of the smallest class in the training data. In order to prove~\eqref{eq:generic-properties-of-training-data-trf}, we also fix $y \in \mathcal{Y} \setminus \{y(x)\}$. It is then obvious that
	\begin{align}
		|\mathcal{D}_{\mathcal{X},y(x)} \setminus \{x\}| = |\mathcal{D}_{\mathcal{X},y(x)}| - 1 \ge c-1
		\qquad \text{and} \qquad
		|\mathcal{D}_{\mathcal{X},y} \setminus \{x\}| = |\mathcal{D}_{\mathcal{X},y}| \ge c.
	\end{align}
	Combining this with~(\ref{eq:conditions-on-hyperparameters}.b) and (\ref{eq:conditions-on-hyperparameters}.c), we further see that there exists a subset $X_{\alpha}' \subset \mathcal{D}_{\mathcal{X},y(x)} \setminus \{x\}$ with $|X_{\alpha}'| = k_{\alpha}$ and a subset $X_{\beta}' \subset \mathcal{D}_{\mathcal{X},y} \setminus \{x\}$ with $|X_{\beta}'| = k_{\beta}$ and therefore, by the definition~\eqref{eq:nearest-neighbors-definition} of nearest neighbors, 
	\begin{align} \label{eq:generic-properties-of-training-data-trf-1}
		\mathrm{NN}_{k_{\alpha},d}(x,\mathcal{D}_{\mathcal{X},y(x)}) < \infty
		\qquad \text{and} \qquad 
		\mathrm{NN}_{k_{\beta},d}(x,\mathcal{D}_{\mathcal{X},y}) < \infty.
	\end{align}
	Since moreover $d$ is a semimetric, we have $\sum_{x'\in X'} d(x,x') > 0$ for every non-empty subset $X'$ of $\mathcal{X} \setminus \{x\}$ and therefore the definition~\eqref{eq:nearest-neighbors-definition} of nearest neighbors further shows that
	\begin{align} \label{eq:generic-properties-of-training-data-trf-2}
		\mathrm{NN}_{k_{\alpha},d}(x,\mathcal{D}_{\mathcal{X},y(x)}) > 0
		\qquad \text{and} \qquad 
		\mathrm{NN}_{k_{\beta},d}(x,\mathcal{D}_{\mathcal{X},y}) > 0.
	\end{align}
	In view of~\cref{eq:attraction-and-repulsion-coefficients-definition,eq:generic-properties-of-training-data-trf-1,eq:generic-properties-of-training-data-trf-2}, the assertion~\eqref{eq:generic-properties-of-training-data-trf} is clear. And since  
	\begin{align*}
		f(x) = \sum_{y\in \mathcal{Y}\setminus\{y(x)\}} \Big( \alpha A_{k_{\alpha},d}(x) + \beta R_{k_{\beta},d}(x,y) \Big) \cdot (-p_y)
	\end{align*}
	by~(\ref{eq:barycenter-0-and-unit-distance-from-0}.a), we finally conclude from~(\ref{eq:conditions-on-hyperparameters}.a) and \eqref{eq:generic-properties-of-training-data-trf} in conjunction with~(\ref{eq:closure-and-interior-of-C_k}.b) that $f(x) \in \mathcal{C}_{y(x)}^{\circ}$, as desired.
\end{proof}

As an immediate consequence of the norm characterization~\eqref{eq:norm-char-of-C_k} of our cone segments, we obtain the  alternative representation~\eqref{eq:casimac-computation-appendix} of our classifier's class label predictions $\widehat{y}(x)$, which is much simpler computationally than the geometrically inspired definition~\eqref{eq:casimac-definition-appendix}. Additionally, we prove that for the training datapoints $x_i$, our classifier correctly predicts the true class label $y(x_i) = y_i$ provided that the regression model perfectly predicts its training datapoints $(x_i,f(x_i)) \in \mathcal{D}^f$. 

\begin{cor} \label{cor:casimac}
	Suppose the assumptions of the previous proposition are satisfied. Suppose further that $\widehat{f}: \mathcal{X} \to \mathcal{Z}$ is an arbitrary map and let $\widehat{y}: \mathcal{X} \to \mathcal{Y}$ be the map defined by
	\begin{align} \label{eq:casimac-definition-appendix}
		\widehat{y}(x) := \widehat{g}(\widehat{f}(x)) := \min\{y \in \mathcal{Y}: \widehat{f}(x) \in \mathcal{C}_y\} \qquad (x \in \mathcal{X}),
	\end{align}
	where $\widehat{g}(z) := \min\{y \in \mathcal{Y}: z \in \mathcal{C}_y\}$. Then
	\begin{align} \label{eq:casimac-computation-appendix}
		\widehat{y}(x) = \min\Big\{ y \in \mathcal{Y}: \big\|\widehat{f}(x) - p_y\big\|_2 = \min_{l\in\mathcal{Y}} \big\|\widehat{f}(x) - p_l\big\|_2 \Big\}
		\qquad (x \in \mathcal{X}).
	\end{align} 
	If, in addition, $\widehat{f}(x_i) = f(x_i)$ for all $i \in \{1,\dots,D\}$, then
	\begin{align} \label{eq:casimac-perfect-class-reconstruction-appendix}
		\widehat{y}(x_i) = y(x_i) = y_i \qquad (i \in \{1,\dots,D\}). 
	\end{align}
\end{cor}

\begin{proof}
	In view of the norm characterization~\eqref{eq:norm-char-of-C_k} of our cone segments, the assertion~\eqref{eq:casimac-computation-appendix} is obvious. 
	If $\widehat{f}(x_i) = f(x_i)$ for all $i$, then $\widehat{f}(x_i) = f(x_i) \in \mathcal{C}_{y(x_i)}^{\circ}$ 
	by Proposition~\ref{prop:generic-properties-of-training-data-trf}. Since $\mathcal{C}_{y(x_i)}^{\circ}$ by~(\ref{eq:segmentation-of-Z-appendix}.b) does not overlap with any other cone segment except for $\mathcal{C}_{y(x_i)}$ itself, we also obtain the assertion~\eqref{eq:casimac-perfect-class-reconstruction-appendix}.
\end{proof}

As a consequence of the essential disjointness~(\ref{eq:segmentation-of-Z-appendix}.b) of our cone segments, we obtain the convenient alternative representation~\eqref{eq:alternative-representation-of-class-probabilities} of our classifier's class label probability predictions $\widehat{p}(\,\cdot\,|x)$. 

\begin{cor} \label{cor:alternative-representation-of-class-probabilities}
	Suppose $p_1,\dots,p_n$ are the vertices of a regular $(n-1)$-simplex in $\mathcal{Z}$ with~\eqref{eq:barycenter-0-and-unit-distance-from-0} and let $\mathcal{C}_1, \dots, \mathcal{C}_n$ be the cone segments defined in~\eqref{eq:def-of-C_k}. Suppose further that $\widehat{q}(\,\cdot\,|x)$ for every $x \in \mathcal{X}$ is a probability density on $\mathcal{Z}$ and let
	\begin{align} \label{eq:def-of-class-probabilities}
		\widehat{p}(y|x) := \int_{\{\widehat{g} = y\}} \widehat{q}(z|x) \d z
		\qquad (x \in \mathcal{X} \text{ and } y \in \mathcal{Y}), 
	\end{align}
	where $\widehat{g}(z) := \min\{y \in \mathcal{Y}: z \in \mathcal{C}_y\}$. Then
	\begin{align} \label{eq:alternative-representation-of-class-probabilities}
		\widehat{p}(y|x) = \int_{\mathcal{C}_y} \widehat{q}(z|x) \d z = \int_{\mathcal{C}_y^{\circ}} \widehat{q}(z|x) \d z
		\qquad (x \in \mathcal{X} \text{ and } y \in \mathcal{Y}). 
	\end{align}
\end{cor}

\begin{proof}
	As a first step, we observe that 
	\begin{align} \label{eq:alternative-representation-of-class-probabilities-1}
		\mathcal{C}_y^{\circ} \subset \{z \in \mathcal{Z}: \widehat{g}(z) = y\} \subset \mathcal{C}_y
	\end{align}
	for every $y \in \mathcal{Y}$. Indeed, the second inclusion is just a trivial consequence of the definition of $\widehat{g}$ and the first inclusion directly follows from the fact that $\mathcal{C}_y^{\circ}$ by~(\ref{eq:segmentation-of-Z-appendix}.b) does not overlap with any cone segment except $\mathcal{C}_y$ itself.  
	As a second step, we show that  
	\begin{align} \label{eq:alternative-representation-of-class-probabilities-2}
		\partial \mathcal{C}_y = \ol{\mathcal{C}}_y \setminus \mathcal{C}_y^{\circ} = \mathcal{C}_y \setminus \mathcal{C}_y^{\circ}
		\subset \bigcup_{k \in \mathcal{Y}\setminus \{y\}} \mathcal{M}_{y k}
	\end{align}
	for every $y \in \mathcal{Y}$, where $\mathcal{M}_{y k}$ as before is the mid-perpendicular hyperplane between the vertices $p_y$ and $p_k$. In order to see this, let $y \in \mathcal{Y}$ be fixed and let $z \in \partial \mathcal{C}_y = \mathcal{C}_y \setminus \mathcal{C}_y^{\circ}$. We then have, on the one hand, that
	\begin{align} \label{eq:alternative-representation-of-class-probabilities-3}
		\norm{z-p_y}_2 = \min_{l\in \mathcal{Y}} \norm{z-p_l}_2 
	\end{align}
	by~\eqref{eq:norm-char-of-C_k} and, on the other hand, this minimum must be attained also at another index $k \ne y$, that is, there must be a $k \in \mathcal{Y} \setminus \{y\}$ with
	\begin{align} \label{eq:alternative-representation-of-class-probabilities-4}
		\norm{z-p_k}_2 = \min_{l\in \mathcal{Y}} \norm{z-p_l}_2.
	\end{align}
	(If the minimum was attained only for the index $y$, this would mean that $\norm{z-p_y}_2 < \norm{z-p_l}_2$ for all $l \ne y$ and therefore we would have  $z \in \mathcal{C}_y^{\circ}$ by~\eqref{eq:norm-char-of-C_k^circ}. Contradiction to our choice of $z$!) Combining~\eqref{eq:alternative-representation-of-class-probabilities-3} and~\eqref{eq:alternative-representation-of-class-probabilities-4}, we see that 
	\begin{align}
		\norm{z-p_y}_2 = \norm{z-p_k}_2
	\end{align}
	for some $k \in \mathcal{Y} \setminus \{y\}$. So, by~\eqref{eq:def-M_kl} and~\eqref{eq:norm-char-of-C_k-1.1}, we see that $z$ must lie on the mid-perpendicular hyperplane $\mathcal{M}_{yk}$. And thus~\eqref{eq:alternative-representation-of-class-probabilities-2} is proven. 
	Since hyperplanes are null sets \wrt~Lebesgue measure on $\mathcal{Z}$, we further conclude from~\eqref{eq:alternative-representation-of-class-probabilities-2} that the boundary $\partial \mathcal{C}_y$ of every cone segment is a null set as well. Combining this, in turn, with~\eqref{eq:alternative-representation-of-class-probabilities-1}, we immediately obtain the alternative representations of $\widehat{p}(y|x)$ from~\eqref{eq:alternative-representation-of-class-probabilities}.
\end{proof}

In the following, we turn to the computation of our class label probability predictions~\eqref{eq:def-of-class-probabilities}, using standard Monte Carlo approximation techniques. In principle, the following approximation result is true for completely arbitrary probability density functions $\widehat{q}(\,\cdot\,|x)$ on $\mathcal{Z}$, but it does require to draw samples from $\widehat{q}(\,\cdot\,|x)$. In the special case of normal-distribution densities $\widehat{q}(\,\cdot\,|x)$, drawing samples is standard, of course (Section~3.24 of~\citep{fishman1996}, for instance). 

\begin{prop} \label{prop:computation-of-class-probabilities}
	Suppose $p_1,\dots,p_n$ are the vertices of a regular $(n-1)$-simplex in $\mathcal{Z}$ with~\eqref{eq:barycenter-0-and-unit-distance-from-0} and let $\mathcal{C}_1, \dots, \mathcal{C}_n$ be the cone segments defined in~\eqref{eq:def-of-C_k}. Suppose further that $\widehat{q}(\,\cdot\,|x)$ for every $x \in \mathcal{X}$ is a probability density on $\mathcal{Z}$ and let
	\begin{align} \label{eq:class-probabilities}
		\widehat{p}(y|x) := \int_{\mathcal{C}_y} \widehat{q}(z|x) \d z = \int_{\mathcal{Z}} I_{\mathcal{C}_y}(z) \widehat{q}(z|x) \d z
		\qquad (x \in \mathcal{X} \text{ and } y \in \mathcal{Y}), 
	\end{align}
	where $I_{\mathcal{C}_y}$ is the indicator function of the cone segment $\mathcal{C}_y$ in analogy to \eqref{eqn:indicator}.
	In the special case where $n=2$ 
	and where $\widehat{q}(\,\cdot\,|x)$ is normally distributed with mean $\widehat{\mu}(x)$ and variance $\widehat{\sigma}(x)^2$, the class probability predictions $\widehat{p}(y|x)$ have the closed-form representation
	\begin{align} \label{eq:class-probability-closed-form}
		\widehat{p}(1|x) = \frac{1}{2}\Big(1 - \operatorname{erf}\big(\widehat{\mu}(x)/(\sqrt{2}\, \widehat{\sigma}(x))\big)\Big)
		= 1-\widehat{p}(2|x),
	\end{align}  
	where the convention~\eqref{eq:vertex-indexing-in-binary-case} was used. 
	In the general case, the class probability prediction $\widehat{p}(y|x)$, for every fixed $x \in \mathcal{X}$ and $y \in \mathcal{Y}$, can be approximated with probability $1$ by the sample means
	\begin{align} \label{eq:class-probability-approximant}
		\widehat{p}_N(y|x) := \frac{1}{N} \sum_{i=1}^N I_{\mathcal{C}_y}(z_i),
	\end{align}
	where $z_1, \dots, z_N$ are sampled independently from the probability density $\widehat{q}(\,\cdot\,|x)$ and where the binary values $I_{\mathcal{C}_y}(z_i)$ are determined by means of the norm characterization~\eqref{eq:norm-char-of-C_k} of the cone segment $\mathcal{C}_y$. Additionally, for every fixed sample size $N \in \N$ and every $\eps > 0$, the probability for $\widehat{p}_N(y|x)$ to miss $\widehat{p}(y|x)$ by more than $\eps$ is bounded above by $1/(4 \eps^2 N)$. 
\end{prop}

\begin{proof}
	In the mentioned special case, we have $\mathcal{C}_1 = (-\infty,0]$ and $\mathcal{C}_2 = [0,\infty)$ by our convention~\eqref{eq:vertex-indexing-in-binary-case} and, with this, the asserted closed-form expression immediately follows. 
	We therefore move on to discuss the general case now. We point out that all the arguments to come are fairly standard and we give them just for the reader's convenience. So, let $x \in \mathcal{X}$ and $y \in \mathcal{Y}$ be fixed, write $Q = Q_x$ for the probability measure on $\mathcal{Z}$ with density $\widehat{q}(\,\cdot\,|x)$, and let $z_1, \dots, z_N$ be independent random variables with distribution $Q$. Writing
	\begin{align}
		\widehat{p} := \widehat{p}(y|x)
		\qquad \text{and} \qquad
		h := h_y := I_{\mathcal{C}_y}
		\qquad \text{and} \qquad
		\widehat{p}_N := \widehat{p}_N(y|x) := \frac{1}{N} \sum_{i=1} h\circ z_i,
	\end{align}
	we can express the expectation value of $h \circ z_i$ \wrt~$Q$ as 
	\begin{align} \label{eq:computation-of-class-probabilities-1}
		\E(h\circ z_i) = \int_{\mathcal{Z}}h(z) \widehat{q}(z|x) \d z = \widehat{p}
	\end{align} 
	for every $i$ and, since $h^2 = h$, we see that the variance of $h \circ z_i$ \wrt~$Q$ is given by 
	\begin{align} \label{eq:computation-of-class-probabilities-2}
		\Var(h\circ z_i) = \E((h\circ z_i)^2) - (\E(h\circ z_i))^2 = \E(h\circ z_i) - (\E(h\circ z_i))^2 = \widehat{p}(1-\widehat{p})
	\end{align}
	for every $i$. Since the $z_i$ are independent and $h$ is bounded, the random variables $h\circ z_i$ are independent and (square) integrable and therefore the strong law of large numbers in conjunction with~\eqref{eq:computation-of-class-probabilities-1} shows that, as $N \to \infty$, the sample means $\widehat{p}_N$ converge to $\widehat{p}$ with $Q$-probability $1$ (or, put differently, $Q$-almost surely). 
	Also, Bienaym\'{e}'s identity in conjunction with~\eqref{eq:computation-of-class-probabilities-2} shows that 
	\begin{align}
		\Var(\widehat{p}_N) = \frac{1}{N^2} \sum_{i=1}^N \Var(h\circ z_i) = \frac{\widehat{p}(1-\widehat{p})}{N}
	\end{align}  
	for every fixed $N \in \N$. So, by Chebyshev's inequality, we get the upper bound
	\begin{align} \label{eq:computation-of-class-probabilities-3}
		Q(\{|\widehat{p}_N-\widehat{p}| > \eps\}) \le \frac{1}{\eps^2} \int |\widehat{p}_N - \widehat{p}|^2 \d Q = \frac{\Var(\widehat{p}_N)}{\eps^2} = \frac{\widehat{p}(1-\widehat{p})}{\eps^2 N}
	\end{align}
	on the $Q$-probability for $\widehat{p}_N(y|x)$ to miss $\widehat{p}(y|x)$ by more than $\eps$. 
	Since $\widehat{p}$ by definition lies between $0$ and $1$, the numerator on the right-hand side of~\eqref{eq:computation-of-class-probabilities-3} can be trivially estimated as $\widehat{p}(1-\widehat{p}) \le 1/4$. Inserting this estimate into~\eqref{eq:computation-of-class-probabilities-3}, we obtain 
	\begin{align} \label{eq:computation-of-class-probabilities-4}
		Q(\{|\widehat{p}_N-\widehat{p}| > \eps\}) \le \frac{1}{4 \eps^2 N}
	\end{align}
	for every $N \in \N$ and $\eps > 0$, which is precisely the asserted probabilistic bound on the approximation error. 
\end{proof}

In addition to the bound~\eqref{eq:computation-of-class-probabilities-4}, one could also apply the central limit theorem to get approximate $95 \%$-confidence intervals in terms of the sample means~\eqref{eq:class-probability-approximant} and the (biased) sample variances
\begin{align*}
	\widehat{\sigma}_N(y|x)^2 := \frac{1}{N} \sum_{i=1}^N \big( h\circ z_i - \widehat{p}_N(y|x) \big)^2
	= \widehat{p}_N(y|x) \big(1-\widehat{p}_N(y|x) \big)
\end{align*}   
for sufficiently large sample sizes $N$. See~\citep{madras2002} (Section~5.1), for instance. It should be noticed, however, that it is not so straightforward to determine what actually is a sufficiently large sample size $N$. 
A detailed discussion of this topic can be found~\citep{fishman1996} (Chapter~2).

\subsection{Compressing the latent space to a reference simplex}

In this section, we show how the latent space can be compressed to a reference simplex in a diffeomorphic and cone-segment-preserving manner. In order to do so, we need a lemma on barycentric coordinates.

\begin{lm} \label{lm:barycentric-coordinates}
	Suppose $p_1,\dots,p_n$ are the vertices of an $(n-1)$-simplex $\mathcal{S}$ in $\mathcal{Z}$. Then for every point $z \in \mathcal{S}$ there exist unique numbers $\lambda_i = \lambda_i(z) \in [0,\infty)$ for $i \in \mathcal{Y}$, the so-called barycentric coordinates of $z$ \wrt~$\mathcal{S}$, such that
	\begin{align} \label{eq:barycentric-coordinates}
		z = \sum_{i\in\mathcal{Y}} \lambda_i \cdot p_i 
		\qquad \text{and} \qquad
		\sum_{i\in\mathcal{Y}} \lambda_i = 1
	\end{align}
	Also, a point $z \in \mathcal{S}$ belongs to the interior $\mathcal{S}^{\circ}$ of $\mathcal{S}$ if and only if all its barycentric coordinates $\lambda_1(z), \allowbreak \dots, \allowbreak \lambda_n(z)$ \wrt~$\mathcal{S}$ are strictly positive. And finally, the barycentric-coordinate map $\mathcal{S} \ni z \mapsto (\lambda_1(z),\dots, \lambda_n(z))$ is infinitely differentiable. 
\end{lm}

\begin{proof}
	Consider the map $\phi: \R^{n-1} \to \mathcal{Z}$ defined by $\phi(\lambda) := \ul{\phi}(\lambda) + p_n$ for all $\lambda \in \R^{n-1}$, where $\ul{\phi}$ is the linear map with $\ul{\phi}(e_i) = p_i-p_n$ for all $i \in \{1,\dots,n-1\}$ and $e_1, \dots, e_{n-1}$ denote the canonical unit vectors in $\R^{n-1}$. It is then clear that 
	\begin{align} \label{eq:barycentric-coordinates-1}
		\phi(\lambda) = \sum_{i=1}^{n-1} \lambda_i \ul{\phi}(e_i) + p_n
		= \sum_{i=1}^{n-1} \lambda_i \cdot p_i + \bigg(1 - \sum_{i=1}^{n-1} \lambda_i\bigg) \cdot p_n
	\end{align}
	for every $\lambda = \sum_{i=1}^{n-1} \lambda_i e_i \in \R^{n-1}$ and from this, in turn, it immediately follows that $\mathcal{S} = \conv\{p_1, \dots, p_n\}$ is the image under $\phi$ of the lower-left corner $\Delta := \{\lambda \in [0,\infty)^{n-1}: \sum_{i=1}^{n-1} \lambda_i \le 1\}$ of the unit cube in $\R^{n-1}$. In short, 
	\begin{align} \label{eq:barycentric-coordinates-2}
		\mathcal{S} = \phi(\Delta). 
	\end{align} 
	It also immediately follows from the linear independence of $\{\ul{\phi}(e_i): i \in \{1,\dots,n-1\}\}$ that $\ul{\phi}$ is a linear isomorphism of $\mathcal{Z} = \R^{n-1}$ and thus also a diffeomorphism of $\mathcal{Z}$ (Theorem~VII.1.6 in conjunction with Example~VII.2.3 (a) and Exercise~VII.5.1 of~\citep{amann2008}). 
	Consequently, the translate $\phi$ of $\ul{\phi}$ is a diffeomorphism of $\mathcal{Z}$ as well. 
	With these preliminary observations, the assertions of the lemma immediately follow. Indeed, by~\eqref{eq:barycentric-coordinates-1} and the bijectivity of $\phi$, the barycentric coordinates of any given $z \in \mathcal{S}$ are uniquely determined by $z$, namely 
	\begin{align} \label{eq:barycentric-coordinates-3}
		(\lambda_1(z),\dots, \lambda_{n-1}(z)) = \phi^{-1}(z)
		\qquad \text{and} \qquad
		\lambda_n(z) = 1 - \sum_{i=1}^{n-1} \lambda_i(z)
	\end{align} 
	for every $z \in \mathcal{S}$. In view of~\eqref{eq:barycentric-coordinates-3} and the infinite differentiability of $\phi^{-1}$, in turn, the asserted infinite differentiability of $\mathcal{S} \ni z \mapsto (\lambda_1(z),\dots, \lambda_n(z))$ follows. And finally, since the interior of $\Delta$ is obviously given by $\Delta^{\circ} = \{\lambda \in (0,\infty)^{n-1}: \sum_{i=1}^{n-1} \lambda_i < 1\}$, we see by~\eqref{eq:barycentric-coordinates-2} and~\citep{dugundji1966} (Theorem~III.11.3) and~\eqref{eq:barycentric-coordinates-1} that
	\begin{align*}
		\mathcal{S}^{\circ} = (\phi(\Delta))^{\circ} = \phi(\Delta^{\circ}) 
		= \big\{z \in \mathcal{Z}: \eqref{eq:barycentric-coordinates} \text{ holds true with } \lambda_1, \dots, \lambda_n \in (0,\infty)\big\}
	\end{align*} 
	which, in turn, yields the asserted characterization of the interior points of $\mathcal{S}$. 
\end{proof}

\begin{prop} \label{prop:compression-of-Z}
	Suppose $p_1,\dots,p_n$ are the vertices of a regular $(n-1)$-simplex in $\mathcal{Z}$ with~\eqref{eq:barycenter-0-and-unit-distance-from-0} and let $\mathcal{C}_1, \dots, \mathcal{C}_n$ be the cone segments defined in~\eqref{eq:def-of-C_k}. Then the compression map $C: \mathcal{Z} \to \mathcal{S}^{\circ}$, defined by
	\begin{align} \label{eq:def-of-compression-map}
		C(z) := \sum_{i\in\mathcal{Y}} \mu_i(z) \cdot p_i
		\quad \text{with} \quad
		\mu_i(z) := \exp(\tau p_i \cdot z)/\Big(\sum_{j\in\mathcal{Y}}\exp(\tau p_j \cdot z)\Big)
	\end{align}
	and a fixed $\tau \in (0,\infty)$, is a diffeomorphism of $\mathcal{Z}$ onto $\mathcal{S}^{\circ}$. Its inverse is given by the inflation map $I: \mathcal{S}^{\circ} \to \mathcal{Z}$, defined by
	\begin{align} \label{eq:def-of-inflation-map}
		I(w) := \frac{n-1}{\tau n}\sum_{i\in\mathcal{Y}} \ln(\lambda_i(w)) \cdot p_i 
	\end{align}
	with $\lambda_1(w), \dots, \lambda_n(w)$ being the barycentric coordinates of $w$ \wrt~$\mathcal{S}$ and $\tau$ is the same positive number as in~\eqref{eq:def-of-compression-map}. Additionally, $C$ and $I$ leave the segments $\mathcal{C}_1, \dots, \mathcal{C}_n$ invariant. 
\end{prop}

\begin{proof}
	It is clear by the definitions~\eqref{eq:def-of-compression-map} and~\eqref{eq:def-of-inflation-map} and by Lemma~\ref{lm:barycentric-coordinates} that the compression $C$ is an  infinitely differentiable map from $\mathcal{Z}$ into $\mathcal{S}^{\circ}$ and that, conversely, the inflation $I$ is an infinitely differentiable map from $\mathcal{S}^{\circ}$ into $\mathcal{Z}$.
	We now show that $I$ is the inverse of $C$ or, equivalently, that
	\begin{align} \label{eq:compression-of-Z-1}
		C(I(w)) = w \qquad (w \in \mathcal{S}^{\circ})
		\qquad \text{and} \qquad
		I(C(z)) = z \qquad (z \in \mathcal{Z}). 
	\end{align}
	In order to verify~(\ref{eq:compression-of-Z-1}.a), we confirm with the help of Lemma~\ref{lm:central-angle-of-reg-simplex} that
	\begin{align*}
		\tau p_i \cdot I(w) = \ln \lambda_i(w) - \frac{1}{n} \sum_{j\in\mathcal{Y}} \ln \lambda_j(w)
		\qquad (w \in \mathcal{S}^{\circ})
	\end{align*}
	and from this, in turn, we arrive at~(\ref{eq:compression-of-Z-1}.a) in a straightforward manner.
	In order to verify~(\ref{eq:compression-of-Z-1}.b), we first notice  that the positive numbers $\mu_1(z), \dots, \mu_n(z)$ from~\eqref{eq:def-of-compression-map} are nothing but the barycentric coordinates $\lambda_1(C(z)), \dots, \lambda_n(C(z))$ of the point $C(z) \in \mathcal{S}^{\circ}$ \wrt~$\mathcal{S}$ and, thus, 
	\begin{align} \label{eq:compression-of-Z-2}
		I(C(z)) = \frac{n-1}{\tau n}\sum_{i\in\mathcal{Y}} \ln(\mu_i(z)) \cdot p_i = \frac{n-1}{n}\sum_{i\in\mathcal{Y}} (p_i \cdot z) \, p_i
		\qquad (z\in\mathcal{Z}),
	\end{align}
	where for the second equality we used (\ref{eq:barycenter-0-and-unit-distance-from-0}.a).
	With the help of Lemma~\ref{lm:central-angle-of-reg-simplex} and again~(\ref{eq:barycenter-0-and-unit-distance-from-0}.a), we further see that $\frac{n-1}{n}\sum_{i\in\mathcal{Y}} (p_i \cdot p_j) \, p_i = p_j$ for every $j \in \mathcal{Y}$ and, by linearly extending this identity to arbitrary $z = \sum_{j\in\mathcal{Y}} r_j p_j$ (Lemma~\ref{lm:vertices-lin-independent}), we conclude that the right-hand side of~\eqref{eq:compression-of-Z-2} equals $z$, as desired. So, \eqref{eq:compression-of-Z-1} is proved, that is, $I = C^{-1}$ is the inverse of $C$. In particular, by the infinite differentiability of $C$ and $I$ pointed out above, $C$ is a diffeomorphism from $\mathcal{Z}$ onto $\mathcal{S}^{\circ}$.
	It thus only remains to prove that $C$ and $I$ leave the segments $\mathcal{C}_1, \dots, \mathcal{C}_n$ invariant or, in other words, that
	\begin{align} \label{eq:compression-of-Z-3}
		C(\mathcal{C}_k) = \mathcal{C}_k \cap \mathcal{S}^{\circ} 
		\qquad \text{and} \qquad
		I(\mathcal{C}_k \cap \mathcal{S}^{\circ}) = \mathcal{C}_k
	\end{align}
	for every $k \in \mathcal{Y}$. In order to see this, we notice that 
	\begin{align*}
		C(z) = \sum_{i\in\mathcal{Y}} \mu_i(z) \cdot p_i = \sum_{i\in\mathcal{Y}} (-\mu_i(z)) \cdot (-p_i)
	\end{align*}
	and thus, by Lemma~\ref{lm:barycentric-char-of-C_k} and~\eqref{eq:def-of-compression-map} and Corollary~\ref{cor:scalar-product-char-of-C_k}, the following chain of equivalences holds true for every $k \in \mathcal{Y}$ and $z \in \mathcal{Z}$:
	\begin{align*} 
		C(z) \in \mathcal{C}_k 
		\quad \text{iff} \quad 
		-\mu_k(z) = \min_{l\in\mathcal{Y}} (-\mu_l(z)) 
		\quad \text{iff} \quad 
		p_k\cdot z = \max_{l\in\mathcal{Y}} p_l \cdot z 
		\quad \text{iff} \quad 
		z \in \mathcal{C}_k.
	\end{align*}
	In short, $C(z) \in \mathcal{C}_k$ if and only if $z \in \mathcal{C}_k$. And this, in turn, immediately implies the asserted invariance statements~\eqref{eq:compression-of-Z-3}. 
\end{proof}

\section{Implementation}
\label{sect:implementation}

\begin{figure}
	\begin{center}
		\includegraphics{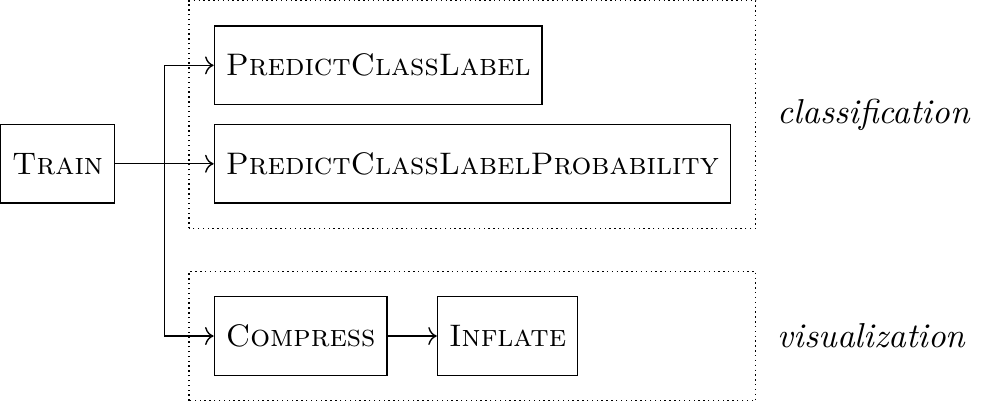}
		\caption{An overview of the functionality of the provided implementation~\citep{casimac-github} as outlined in \cref{alg:casimac}. After the initial training stage, there are two possible use cases, namely classification and visualization, as described in detail in \cref{sect:method}. The first use case refers to the prediction of class labels and of class probabilities for unseen feature space points, whereas the second use case allows to perform a transformation from the latent space to a reference simplex and back for visualization purposes.} \label{fig:implementation}
	\end{center}
\end{figure}

A Python implementation of our proposed method is available online~\citep{casimac-github}. The implementation includes the following main functionality, which is outlined in \cref{alg:casimac}. A sketch of the functionality is presented in \cref{fig:implementation}.
\begin{itemize}
	\item \textproc{Train} (\cref{alg:casimac:train}): Calculate the simplex vertices $p_1, \dots, p_n$ and train the regression model based on a training data set $\mathcal{D}$, the user-defined parameters $\alpha$, $\beta$, $k_{\alpha}$, $k_{\beta}$, and $d$ for the training data transformation $f$ and the hyperparameters $h$ of the regression model. In this manner, obtain a point prediction $\widehat{f}(x)$ -- possibly along with a probabilistic prediction $\widehat{q}(\,\cdot\,|x)$ -- for every $x\in \mathcal{X}$. For the calculation of the simplex vertices $p_1, \dots, p_n$, we provide two methods. First, the explicit method via \cref{eq:sample-reg-simplex-in-Z-7} and second, an iterative method. For the latter, we choose $p_1$ to be the first canonical basis vector and determine the remaining elements of $p_2, \dots, p_n$ one after another based on the already specified elements such that (\ref{eq:barycenter-0-and-unit-distance-from-0}.b) and (\ref{eq:simplex-central-angle-of-reg-simplex}.b) are fulfilled. The resulting simplex vectors from the two methods differ only by a joint rotation so that there is no substancial effect on our method apart from minor numerical differences. All results in this paper are based on simplex vertices from the iterative method.
	\item \textproc{PredictClassLabel} (\cref{alg:casimac:predictlabel}): Predict the class label of (possibly unseen) feature space points $x \in \mathcal{X}$ based on the point prediction $\widehat{f}(x)$ of the trained regression model and on the simplex vertices $p_1, \dots, p_n$.
	\item \textproc{PredictClassLabelProbability} (\cref{alg:casimac:predictproba}): Predict the probability of class label $y \in \mathcal{Y}$ for unseen feature space points $x \in \mathcal{X}$ based on the probabilistic prediction $\widehat{q}(\,\cdot\,|x)$ of the trained regression model, 
	on the simplex vertices $p_1, \dots, p_n$, and on a fixed sample size $N$. We presume that $\widehat{q}(\,\cdot\,|x)$ is normally distributed with mean $\widehat{\mu}(x)$ and variance $\widehat{\sigma}(x)^2$.
	\item \textproc{Compress} (\cref{alg:casimac:transform}): Transform the latent space point $z \in \mathcal{Z}$ to reference-simplex counterpart $w = C(z) \in -\mathcal{S}^{\circ}$ based on the user-defined parameter $\tau$ and the simplex vertices $p_1, \dots, p_n$.
	\item \textproc{Inflate} (\cref{alg:casimac:invtransform}): Transform the reference-simplex point $w \in -\mathcal{S}^{\circ}$ to latent space counterpart $z = I(w) \in \mathcal{Z}$ based on the user-defined parameter $\tau$ and the simplex vertices $p_1, \dots, p_n$.
\end{itemize}

In particular, our implementation is intended to be used with scikit-sklearn~\citep{sklearn2011}. We therefore do not prescribe the specific form of the regression model, but instead allow the user to plug in an arbitrary scikit-sklearn estimator. 
\begin{algorithm}[!htb]
	\caption{Summary of the main functionality of the provided implementation~\citep{casimac-github}.} \label{alg:casimac}
	\begin{algorithmic}[1]
		\Function{Train}{$\mathcal{D}$, $\alpha$, $\beta$, $k_{\alpha}$, $k_{\beta}$, $d$, $h$} \label{alg:casimac:train}
		\State Calculate simplex vertices $p_1, \dots, p_n$ either explicitly according to~\cref{eq:sample-reg-simplex-in-Z-7} using~\cref{eq:sample-reg-simplex-in-Z-6,eq:sample-reg-simplex-in-Z-8} or iteratively as described in \cref{sect:implementation}.
		\State Calculate $f(x_i) = f_{\alpha,\beta,k_{\alpha},k_{\beta},d}(x_i|\mathcal{D})$ for $i \in \{1,\dots, D\}$ using~\cref{eq:training-data-trf-defining-formula,eq:attraction-and-repulsion-coefficients-definition,eq:nearest-neighbors-definition,eq:metric-d,eq:conditions-on-hyperparameters} 
		\State Calculate the transformed training data set $\mathcal{D}^f$ from \eqref{eq:transformed-training-data} 
		\State Train a regression model with hyperparameters $h$, by fitting it to $\mathcal{D}^f$ 
		\EndFunction
		
		\Function{PredictClassLabel}{$x$, $\widehat{f}$, $p_1, \dots, p_n$} \label{alg:casimac:predictlabel}
		\State Calculate the regression model's point prediction $\widehat{z} = \widehat{f}(x)$ for $x$
		\State Calculate the distances $\norm{\widehat{z}-p_1}_2, \dots, \norm{\widehat{z}-p_n}_2$ 
		\State Calculate the label $\widehat{y} = \widehat{g}(\widehat{z})$ based on the above distances and~\eqref{eq:casimac-alternative-definition}
		\State \Return $\widehat{y}$ 
		\EndFunction
		
		\Function{PredictClassLabelProbability}{$x$, $y$, $\widehat{\mu}$, $\widehat{\sigma}$, $p_1, \dots, p_n$, $N$} \label{alg:casimac:predictproba}
		\If {$n=2$}
		\State Calculate $\widehat{p} = \widehat{p}(y|x)$ according to~\eqref{eq:class-probability-closed-form}
		\Else
		\State Sample $z_1, \dots, z_N$ independently from $\widehat{q}(\,\cdot\,|x)$
		\State Calculate the distances $\norm{z_i-p_1}_2, \dots, \norm{z_i-p_n}_2$ for $i \in \{1,\dots,N\}$
		\State Calculate $I_{\mathcal{C}_y}(z_1),\dots, I_{\mathcal{C}_y}(z_N)$ based on the above distances and on~\eqref{eq:norm-char-of-C_k}
		\State Calculate $\widehat{p} = \widehat{p}_N(y|x)$ according to~\eqref{eq:class-probability-approximant} 
		\EndIf
		\State \Return $\widehat{p}$
		\EndFunction
		
		\Function{Compress}{$z$, $\tau$, $p_1, \dots, p_n$} \label{alg:casimac:transform}
		\State Calculate $w = C(z)$ according to~\eqref{eq:def-of-compression-map}
		\State \Return $w$	
		\EndFunction
		
		\Function{Inflate}{$w$, $\tau$, $p_1, \dots, p_n$} \label{alg:casimac:invtransform}
		\State Calculate $z = I(w)$ according to~\eqref{eq:def-of-inflation-map}
		\State \Return $z$	
		\EndFunction
	\end{algorithmic}
\end{algorithm}

\section{Hyperparameters}
\label{sect:hyperparameters}

In \cref{sect:validation} we use cross-validation to select the best hyperparameters for the classifiers out of a pre-defined set of possible choices. Specifically, we vary the number of nearest neighbors $k \in \{ \num{5},\allowbreak \num{10},\allowbreak \num{15},\allowbreak \num{20} \}$ for kNN, the hidden layer sizes $L \in \{ (\num{5}),\allowbreak (\num{10}),\allowbreak (\num{5},\num{5}), (\num{5},\num{10}),\allowbreak (\num{10},\num{10}),\allowbreak (\num{5},\num{5},\num{5}),\allowbreak (\num{5},\num{5},\num{10}),\allowbreak (\num{5},\num{10},\num{10}),\allowbreak (\num{10},\num{10},\num{10}) \}$ (all with a rectified linear unit as their activation function) for MLP and the kernel parameters for both GPC and the GPR model used for CASIMAC. We choose a sum of a Mat\'{e}rn kernel and a white-noise kernel~\citep{rasmussen2006} for these kernels and tune the Mat\'{e}rn coefficient $\nu \in \{ \frac{3}{2}, \frac{5}{2}, \infty \}$. The remaining kernel parameters are optimized for each cross-validation setup as described in \cite{rasmussen2006}. Additionally, for CASIMAC, we tune both $\gamma$ from~\eqref{eq:gamma} as well as $k_{\alpha}$ and $k_{\beta}$ according to~\eqref{eq:conditions-on-hyperparameters}.
In \cref{tab:app:hyper-params}, we list the cross-validated hyperparameters which result in the best accuracy over all classification tasks for each data set from \cref{sect:real-world-data}. Analogously, for the synthetic data set from \cref{sect:synthetic-data} we get the best hyperparameters $\gamma=1 $, $k_\alpha = 10$, $k_\beta = 10$, and $\nu=\infty$ for CASIMAC and $\nu=\frac{3}{2}$ for GPC. And finally, for the \texttt{alcohol-3} data set from \cref{sect:visualization-example} $\gamma=\frac{1}{2}$, $k_\alpha = 1$, and $k_\beta = 5$ are fixed and we find $\nu=\infty$ for CASIMAC from the cross-validation.

\begin{table}
	\centering
	\caption{The best cross-validated hyperparameters for the real-world data sets from \cref{tab:datasets}.}\label{tab:app:hyper-params}
	\begin{tabular}{cccccc}\toprule
		& \texttt{alcohol} & \texttt{climate} & \texttt{hiv} & \texttt{pine} & \texttt{wifi} \\ \midrule
		\STAB{\rotatebox[origin=c]{90}{\hspace{.1cm}\textbf{CASIMAC}\hspace{.1cm}}}%
		& \makecell{$\gamma=0 $ \\ $k_\alpha = 5$ \\ $k_\beta = 5$ \\ $\nu=\infty$} &
		\makecell{$\gamma=0 $ \\ $k_\alpha = 5$ \\ $k_\beta = 5$ \\ $\nu=\infty$} &
		\makecell{$\gamma=0 $ \\ $k_\alpha = 20$ \\ $k_\beta = 20$ \\ $\nu=\frac{3}{2}$} &
		\makecell{$\gamma=1 $ \\ $k_\alpha = 1$ \\ $k_\beta =5$ \\ $\nu=\frac{3}{2}$} &
		\makecell{$\gamma=1/3$ \\ $k_\alpha = 20$ \\ $k_\beta = 20$ \\ $\nu=\infty$} \\ 
		\STAB{\rotatebox[origin=c]{90}{\hspace{.1cm}\textbf{GPC}\hspace{.1cm}}}%
		& $\nu=\frac{3}{2}$ &
		$\nu=\frac{3}{2}$ &
		$\nu=\infty$ &
		$\nu=\frac{3}{2}$ &
		$\nu=\frac{3}{2}$ \\ 
		\STAB{\rotatebox[origin=c]{90}{\hspace{.1cm}\textbf{kNN}\hspace{.1cm}}}%
		& $k=\num{5}$ &
		$k=\num{5}$ &
		$k=\num{5}$ &
		$k=\num{5}$ &
		$k=\num{5}$ \\ 
		\STAB{\rotatebox[origin=c]{90}{\hspace{.1cm}\textbf{MLP}\hspace{.1cm}}}%
		& \makecell{$L=$ \\ $(\num{10},\num{10},\num{10})$} &
		\makecell{$L=$ \\ $(\num{5}, \num{5})$} &
		\makecell{$L=$ \\ $(\num{5}, \num{10})$} &
		\makecell{$L=$ \\ $(\num{10},\num{10},\num{10})$} &
		\makecell{$L=$ \\ $(\num{5}, \num{5})$} \\ \bottomrule
	\end{tabular}
\end{table}
\end{appendix}
\FloatBarrier
\bibliography{literature}

\begin{thebibliography}{10}

\bibitem{challis2015}
Edward Challis, Peter Hurley, Laura Serra, Marco Bozzali, Seb Oliver, and Mara
  Cercignani.
\newblock {Gaussian} process classification of {A}lzheimer's disease and mild
  cognitive impairment from resting-state {fMRI}.
\newblock {\em NeuroImage}, 112:232--243, 2015.

\bibitem{niculescu2005}
Alexandru Niculescu-Mizil and Rich Caruana.
\newblock Predicting good probabilities with supervised learning.
\newblock In {\em International Conference on Machine Learning}, ICML '05,
  pages 625--632, New York, NY, USA, 2005. ACM.

\bibitem{platt2000}
John~C. Platt.
\newblock Probabilistic outputs for support vector machines and comparisons to
  regularized likelihood methods.
\newblock In {\em Advances in Large Margin Classifiers}, pages 61--74. MIT
  Press, 1999.

\bibitem{zadrozny2002}
Bianca Zadrozny and Charles Elkan.
\newblock Transforming classifier scores into accurate multiclass probability
  estimates.
\newblock In {\em International Conference on Knowledge Discovery and Data
  Mining}, KDD '02, pages 694--699, New York, NY, USA, 2002. ACM.

\bibitem{gebel2009}
Martin Gebel.
\newblock {\em Multivariate calibration of classification scores into the
  probability space}.
\newblock PhD thesis, University of Dortmund, 2009.

\bibitem{calandra2016}
Roberto Calandra, Jan Peters, Carl~Edward Rasmussen, and Marc~Peter Deisenroth.
\newblock Manifold {Gaussian} processes for regression.
\newblock In {\em International Joint Conference on Neural Networks}, IJCNN
  '16, pages 3338--3345. IEEE, 2016.

\bibitem{wilson2016}
Andrew~G. Wilson, Zhiting Hu, Russ~R Salakhutdinov, and Eric~P. Xing.
\newblock Stochastic variational deep kernel learning.
\newblock In {\em Advances in Neural Information Processing Systems}, pages
  2586--2594, 2016.

\bibitem{alshedivat2017}
Maruan Al-Shedivat, Andrew~Gordon Wilson, Yunus Saatchi, Zhiting Hu, and
  Eric~P. Xing.
\newblock Learning scalable deep kernels with recurrent structure.
\newblock {\em {The Journal of Machine Learning Research}}, 18(1):2850--2886,
  2017.

\bibitem{bradshaw2017}
John Bradshaw, Alexander~{G. de G.} Matthews, and Zoubin Ghahramani.
\newblock Adversarial examples, uncertainty, and transfer testing robustness in
  {Gaussian} process hybrid deep networks.
\newblock {\em preprint arXiv:1707.02476}, 2017.

\bibitem{daskalakis2020}
Constantinos Daskalakis, Petros Dellaportas, and Aristeidis Panos.
\newblock Scalable {Gaussian} processes, with guarantees: Kernel approximations
  and deep feature extraction.
\newblock {\em preprint arXiv:2004.01584}, 2020.

\bibitem{iwata2017}
Tomoharu Iwata and Zoubin Ghahramani.
\newblock Improving output uncertainty estimation and generalization in deep
  learning via neural network {Gaussian} processes.
\newblock {\em preprint arXiv:1707.05922}, 2017.

\bibitem{cremanns2017}
Kevin Cremanns and Dirk Roos.
\newblock Deep {Gaussian} covariance network.
\newblock {\em preprint arXiv:1710.06202}, 2017.

\bibitem{liu2018}
Haitao Liu, Yew-Soon Ong, Xiaobo Shen, and Jianfei Cai.
\newblock When {Gaussian} process meets big data: {A} review of scalable {GPs}.
\newblock {\em preprint arXiv:1807.01065}, 2018.

\bibitem{lifecycle2021}
Patrick~Otto Ludl, Raoul Heese, Johannes H{\"o}ller, Norbert Asprion, and
  Michael Bortz.
\newblock Using machine learning models to explore the solution space of large
  nonlinear systems underlying flowsheet simulations with constraints.
\newblock {\em Frontiers of Chemical Science and Engineering}, Aug 2021.

\bibitem{rasmussen2006}
Carl~E. Rasmussen and Christopher K.~I. Williams.
\newblock {\em {Gaussian} Processes for Machine Learning}.
\newblock Adaptative computation and machine learning series. University Press
  Group Limited, 2006.

\bibitem{murphy2012}
Kevin~P. Murphy.
\newblock {\em Machine learning: a probabilistic perspective}.
\newblock Adaptive computation and machine learning series. MIT Press, 2012.

\bibitem{tran2015}
Dustin Tran, Rajesh Ranganath, and David~M. Blei.
\newblock The variational {Gaussian} process.
\newblock {\em preprint arXiv:1511.06499}, 2015.

\bibitem{hensman2015}
James Hensman, Alexander~G. Matthews, Maurizio Filippone, and Zoubin
  Ghahramani.
\newblock {MCMC} for variationally sparse {Gaussian} processes.
\newblock In C.~Cortes, N.~D. Lawrence, D.~D. Lee, M.~Sugiyama, and R.~Garnett,
  editors, {\em Advances in Neural Information Processing Systems 28}, pages
  1648--1656. Curran Associates, Inc., 2015.

\bibitem{gpy2014}
{GPy}.
\newblock {GPy}: A {Gaussian} process framework in {Python}.
\newblock Available online: \url{http://github.com/SheffieldML/GPy}, 2012.

\bibitem{dugundji1966}
James Dugundji.
\newblock {\em Topology}.
\newblock Allyn and Bacon, 1966.

\bibitem{wilson1931}
Wallace~Alvin Wilson.
\newblock On semi-metric spaces.
\newblock {\em American Journal of Mathematics}, 53:361--373, 1931.

\bibitem{deza2016}
Michel~Marie Deza and Elena Deza.
\newblock {\em Encyclopedia of Distances}.
\newblock Springer, Berlin, 2016.

\bibitem{casimac-github}
Raoul Heese.
\newblock {CASIMAC}: Calibrated simplex mapping classifier in {Python}.
\newblock Available online: \url{https://github.com/raoulheese/casimac}, 2022.

\bibitem{sklearn2011}
F.~Pedregosa, G.~Varoquaux, A.~Gramfort, V.~Michel, B.~Thirion, O.~Grisel,
  M.~Blondel, P.~Prettenhofer, R.~Weiss, V.~Dubourg, J.~Vanderplas, A.~Passos,
  D.~Cournapeau, M.~Brucher, M.~Perrot, and E.~Duchesnay.
\newblock {Scikit-learn}: Machine learning in {Python}.
\newblock {\em {Journal of Machine Learning Research}}, 12:2825--2830, 2011.

\bibitem{uci}
Dheeru Dua and Casey Graff.
\newblock {UCI} machine learning repository.
\newblock Available online: \url{http://archive.ics.uci.edu/ml}, 2017.

\bibitem{adak2019}
M.~Fatih Adak, Peter Lieberzeit, Purim Jarujamrus, and Nejat Yumusak.
\newblock Classification of alcohols obtained by {QCM} sensors with different
  characteristics using {ABC} based neural network.
\newblock {\em {Engineering Science and Technology, an International Journal}},
  2019.

\bibitem{lucas2013}
D.~D. Lucas, R.~Klein, J.~Tannahill, D.~Ivanova, S.~Brandon, D.~Domyancic, and
  Y.~Zhang.
\newblock Failure analysis of parameter-induced simulation crashes in climate
  models.
\newblock {\em {Geoscientific Model Development}}, 6(4):1157--1171, 2013.

\bibitem{thorsteinn2014}
Thorsteinn R{\"o}gnvaldsson, Liwen You, and Daniel Garwicz.
\newblock State of the art prediction of {HIV-1} protease cleavage sites.
\newblock {\em {Bioinformatics}}, 31(8):1204--1210, 12 2014.

\bibitem{baumgardner2015}
Marion~F. Baumgardner, Larry~L. Biehl, and David~A. Landgrebe.
\newblock 220 band {AVIRIS} hyperspectral image data set: {June} 12, 1992
  {Indian Pine Test Site 3}.
\newblock Available online: \url{https://purr.purdue.edu/publications/1947/1},
  9 2015.

\bibitem{data-pine}
M.~Gra{\~{n}}a and B.~Ayerdi M.~A.~Veganzons.
\newblock Hyperspectral remote sensing scenes.
\newblock Available online:
  \url{http://www.ehu.eus/ccwintco/index.php/Hyperspectral_Remote_Sensing_Scenes},
  2014.

\bibitem{rohra2916}
Jayant~G. Rohra, Boominathan Perumal, Swathi~Jamjala Narayanan, Priya Thakur,
  and Rajen~B. Bhatt.
\newblock User localization in an indoor environment using fuzzy hybrid of
  particle swarm optimization \& gravitational search algorithm with neural
  networks.
\newblock In {\em International Conference on Soft Computing for Problem
  Solving}, SocProS '16, pages 286--295, 2016.

\bibitem{opitz2019macro}
Juri Opitz and Sebastian Burst.
\newblock {Macro F1 and Macro F1}.
\newblock {\em preprint arXiv:1911.03347}, 2019.

\bibitem{degroot1983}
Morris~H. {DeGroot} and Stephen~E. Fienberg.
\newblock The comparison and evaluation of forecasters.
\newblock {\em {Journal of the Royal Statistical Society. Series D (The
  Statistician)}}, 32(1/2):12--22, 1983.

\bibitem{xiao2017}
Han Xiao, Kashif Rasul, and Roland Vollgraf.
\newblock {Fashion-MNIST}: a novel image dataset for benchmarking machine
  learning algorithms.
\newblock {\em preprint arXiv:1708.07747}, 2017.

\bibitem{wang2004}
{Zhou Wang}, A.~C. {Bovik}, H.~R. {Sheikh}, and E.~P. {Simoncelli}.
\newblock Image quality assessment: from error visibility to structural
  similarity.
\newblock {\em {IEEE Transactions on Image Processing}}, 13(4):600--612, April
  2004.

\bibitem{wang2009}
Zhou Wang and Alan~C. Bovik.
\newblock Mean squared error: {Love} it or leave it? {A} new look at signal
  fidelity measures.
\newblock {\em {IEEE Signal Processing Magazine}}, 26(1):98--117, Jan 2009.

\bibitem{scikit-image}
Stefan {Van der Walt}, Johannes~L. Sch{\"o}nberger, Juan Nunez-Iglesias,
  Fran{\c{c}}ois Boulogne, Joshua~D. Warner, Neil Yager, Emmanuelle Gouillart,
  and Tony Yu.
\newblock {scikit-image}: image processing in {Python}.
\newblock {\em {PeerJ}}, 2:e453, 2014.

\bibitem{informed2019}
Laura~von Rueden, Sebastian Mayer, Katharina Beckh, Bogdan Georgiev, Sven
  Giesselbach, Raoul Heese, Birgit Kirsch, Michal Walczak, Julius Pfrommer,
  Annika Pick, Rajkumar Ramamurthy, Jochen Garcke, Christian Bauckhage, and
  Jannis Schuecker.
\newblock Informed machine learning - a taxonomy and survey of integrating
  prior knowledge into learning systems.
\newblock {\em IEEE Transactions on Knowledge and Data Engineering}, page~1,
  2021.

\bibitem{dillon2017}
Joshua~V. Dillon, Ian Langmore, Dustin Tran, Eugene Brevdo, Srinivas Vasudevan,
  Dave Moore, Brian Patton, Alex Alemi, Matthew~D. Hoffman, and Rif~A. Saurous.
\newblock {TensorFlow} distributions.
\newblock {\em preprint arXiv:1711.10604}, abs/1711.10604, 2017.

\bibitem{rupp2012}
Matthias Rupp, Alexandre Tkatchenko, Klaus-Robert M{\"u}ller, and O.~Anatole
  {Von Lilienfeld}.
\newblock Fast and accurate modeling of molecular atomization energies with
  machine learning.
\newblock {\em Physical review letters}, 108(5):058301, 2012.

\bibitem{bellet2013}
Aur\'{e}lien Bellet, Amaury Habrard, and Marc Sebban.
\newblock A survey on metric learning for feature vectors and structured data.
\newblock {\em preprint arXiv:1306.6709}, 2013.

\bibitem{kurnatowski2021}
Martin~von Kurnatowski, Jochen Schmid, Patrick Link, Rebekka Zache, Lukas
  Morand, Torsten Kraft, Ingo Schmidt, Jan Schwientek, and Anke Stoll.
\newblock Compensating data shortages in manufacturing with monotonicity
  knowledge.
\newblock {\em Algorithms}, 14(12), 2021.

\bibitem{schmid2021}
Jochen Schmid.
\newblock Approximation, characterization, and continuity of multivariate
  monotonic regression functions.
\newblock {\em Analysis and Applications},
  https://doi.org/10.1142/S0219530521500299, 2021.

\bibitem{link2022}
Patrick Link, Miltiadis Poursanidis, Jochen Schmid, Rebekka Zache, Martin von
  Kurnatowski, Uwe Teicher, and Steffen Ihlenfeldt.
\newblock Capturing and incorporating expert knowledge into machine learning
  models for quality prediction in manufacturing.
\newblock {\em Journal of Intelligent Manufacturing}, 33(7):2129--2142, 2022.

\bibitem{dhanabal2011}
S.~Dhanabal and Dr.~S. Chandramathi.
\newblock A review of various k-nearest neighbor query processing techniques.
\newblock {\em {International Journal of Computer Applications}}, 31(7):14--22,
  10 2011.

\bibitem{rockafellar1970}
R.~Tyrrell Rockafellar.
\newblock {\em Convex analysis}.
\newblock Princeton Mathematical Series. Princeton University Press, Princeton,
  N. J., 1970.

\bibitem{coxeter1973}
Harold S.~M. Coxeter.
\newblock {\em Regular polytopes}.
\newblock Dover books on advanced mathematics. Dover, New York, 1973.

\bibitem{tuy2016}
Hoang Tuy.
\newblock {\em Convex analysis and global optimization}.
\newblock Springer, 2nd edition, 2016.

\bibitem{parks2002}
Harold~R. Parks and Dean~C. Wills.
\newblock An elementary calculation of the dihedral angle of the regular
  $n$-simplex.
\newblock {\em {The American Mathematical Monthly}}, 109(8):756--758, 2002.

\bibitem{amann2008}
Herbert Amann and Joachim Escher.
\newblock {\em Analysis II}.
\newblock Birkh\"auser, 2008.

\bibitem{fishman1996}
George~S. Fishman.
\newblock {\em {Monte} {Carlo}: concepts, algorithms, and applications.}
\newblock Springer, 1996.

\bibitem{madras2002}
Neal Madras.
\newblock {\em Lectures on {Monte} {Carlo} methods.}
\newblock Fields Institute Monographs. American Mathematical Society, 2002.

\end{thebibliography}

\end{document}